\documentclass[runningheads]{llncs}
\usepackage{accv}

\usepackage{graphicx}
\usepackage{amsmath,amssymb}
\usepackage{color}
\usepackage{colortbl}
\usepackage{xcolor}
\usepackage{array}
\usepackage{tabularx}
\usepackage{multirow}
\usepackage{todonotes}
\usepackage{cite}
\usepackage{booktabs}
\usepackage{caption}
\usepackage{subcaption}
\usepackage{nicefrac}
\usepackage{tikz}
\usepackage{censor}
\usepackage{algorithm}
\usepackage{algpseudocode}
\usepackage{wasysym}
\usepackage{pifont} 
\usetikzlibrary{automata, positioning, patterns, patterns.meta}
\usepackage[accsupp]{axessibility}  

\usepackage{hyperref}
\usepackage{orcidlink}
\newcommand{\cmark}{\color{green!40!black}{\ding{51}}}
\newcommand{\xmark}{\color{red}{\ding{55}}}
\newcommand{\mcrot}[4]{\multicolumn{#1}{#2}{\rlap{\rotatebox{#3}{#4}~}}} 
\newcommand*{\twoelementtable}[3][l]%
{%
    \begin{tabular}[t]{@{}#1@{}}%
        #2\tabularnewline
        #3%
    \end{tabular}%
}

\DeclareMathOperator*{\argmin}{arg\,min}
\newcommand{\Sym}{\mathcal{S}}
\newcommand{\eclass}{\mathbb{F}}
\newcommand{\ecweights}{\theta}
\newcommand{\avec}{\alpha}  
\newcommand{\svec}{\beta}   
\newcommand{\ivec}{\svec}   
\newcommand{\pvec}{\vartheta}  
\newcommand{\evec}{\varphi} 
\newcommand{\emo}{e}        
\newcommand{\Iphi}{I_{\varphi^{(\emo)}}} 
\newcommand{\siglevel}{\delta} 

\newcommand{\paratitle}[1]{\textbf{#1}}

\colorlet{angry}{orange}
\colorlet{disgust}{green}
\colorlet{fear}{red}
\colorlet{happy}{violet}
\colorlet{sad}{Brown}
\colorlet{surprise}{Magenta}
\colorlet{neutral}{gray}

\definecolor{lowacc}{HTML}{440154}
\definecolor{highacc}{HTML}{FDE725}

\newcommand{\cwidth}{1.1cm}

\newcolumntype{a}{>{\columncolor{angry!20}\centering}m{\cwidth}}
\newcolumntype{d}{>{\columncolor{disgust!20}\centering}m{\cwidth}}
\newcolumntype{e}{>{\columncolor{fear!20}\centering}m{\cwidth}}
\newcolumntype{f}{>{\columncolor{happy!20}\centering}m{\cwidth}}
\newcolumntype{g}{>{\columncolor{sad!20}\centering}m{\cwidth}}
\newcolumntype{h}{>{\columncolor{surprise!20}\centering\arraybackslash}m{\cwidth}}

\newcommand{\neutral}{\emph{\color{neutral}neutral }} 
\newcommand{\angry}{\emph{\color{angry}angry }} 
\newcommand{\disgust}{\emph{\color{disgust}disgust }} 
\newcommand{\fear}{\emph{\color{fear}fear }} 
\newcommand{\happy}{\emph{\color{happy}happy }} 
\newcommand{\sad}{\emph{\color{sad}sad }} 
\newcommand{\surprise}{\emph{\color{surprise}surprise }} 

\title{
    Facing Asymmetry - Uncovering the Causal Link between Facial Symmetry and Expression Classifiers using Synthetic Interventions
}
\titlerunning{Facing Asymmetry via Synthetic Interventions}
\authorrunning{Büchner et al.}

\author{
    Tim Büchner$^*$\inst{1}\orcidID{0000-0002-6879-552X} \and
    Niklas Penzel$^*$\inst{1}\orcidID{0000-0001-8002-4130} \and
    Orlando Guntinas-Lichius\inst{2}\orcidID{0000-0001-9671-0784} \and
    Joachim Denzler\inst{1}\orcidID{0000-0002-3193-3300}
}
\institute{
    Computer Vision Group, Friedrich Schiller University Jena, 07743 Jena, Germany
    \and
    Dept. of Otorhinolaryngology, Jena University Hospital, 07747 Jena, Germany
    \\
    \email{tim.buechner@uni-jena.de}
}

\begin{document}
\pagestyle{headings}
\mainmatter
\maketitle

\begin{NoHyper}
\def\thefootnote{*}\footnotetext{These authors contributed equally to this work.}\def\thefootnote{\arabic{footnote}}
\end{NoHyper}

\Crefname{figure}{Fig.}{Figs.}
\crefname{figure}{fig.}{figs.}

\Crefname{section}{Sec.}{Secs.}
\crefname{section}{sec.}{secs.}

\Crefname{equation}{Eq.}{Eqs.}
\crefname{equation}{eq.}{eqs.}

\begin{abstract}
Understanding expressions is vital for deciphering human behavior, and nowadays, end-to-end trained black box models achieve high performance.
Due to the black-box nature of these models, it is unclear how they behave when applied out-of-distribution.
Specifically, these models show decreased performance for unilateral facial palsy patients.
We hypothesize that one crucial factor guiding the internal decision rules is facial symmetry.
In this work, we use insights from causal reasoning to investigate the hypothesis.
After deriving a structural causal model, we develop a synthetic interventional framework.
This approach allows us to analyze how facial symmetry impacts a network's output behavior while keeping other factors fixed.
\underline{All} 17 investigated expression classifiers significantly lower their output activations for reduced symmetry.
This result is congruent with observed behavior on real-world data from healthy subjects and facial palsy patients.
As such, our investigation serves as a case study for identifying causal factors that influence the behavior of black-box models.


\end{abstract}

\section{Introduction}
Emotional expressiveness is a crucial topic in our daily life for communicating our internal state and for understanding other people~\cite{thompson1987empathy,roberts1996empathy}.
The state-of-the-art for automatically classifying the six base emotions~\cite{ekman1992argument} is achieved by end-to-end trained black box neural networks~\cite{pham2021facial,savchenko2023facial,zhangDualDirectionAttentionMixed2023,wenDistractYourAttention2023,boundori2023EmoNext,daneVcVekEMOCAEmotionDriven2022}.
However, it remains unclear how the internal decision-making processes of these models respond to out-of-distribution inputs due to likely unbalanced training data.
Specifically, we observe a performance degradation when classifying facial expressions in individuals with unilateral facial palsy, a condition impairing the ability to produce symmetrical facial expressions due to underlying nerve damage.
Although intuition suggests that facial asymmetry could influence model behavior, we lack a quantifiable way to test and validate this hypothesis.

We leverage causal reasoning principles to address this limitation and move beyond empirical analytics to uncover a contributing factor in the underlying mechanisms driving model decision-making.
Specifically, our work answers the interventional question~\cite{bareinboim2022onpearl}: ``If we only change the facial symmetry for an input, then how does the output of an expression classifier behave?''
First, we provide evidence that a symmetry bias exists for real-world data inside all models using associational methods~\cite{buechner2024power,penzel2023analyzing,reimers2020determining,reimers2021debiasing}.
Second, moving up on the causal hierarchy~\cite{bareinboim2022onpearl}, we build an interventional framework derived from a structural causal model that allows us to generate synthetic faces and connect symmetry with classifier outputs.
To accurately quantify this relationship, we develop an interpretable score and an accompanying hypothesis test.
As a case study, we analyze 17 expression classifiers and find significant changes in their predictions for \underline{all} of them.
Specifically, we find that decreases in facial symmetry result in lower logit activations.
Our study highlights the importance of symmetry influencing expression classifiers, emphasizing the general need for investigations beyond predictive performance.

\section{Related Work}
Synthetic data has become a widely accepted tool for evaluating and training computer vision models in diverse applications, such as object detection~\cite{wu2022synthetic,nowruzi2019much,thalhammer2019sydpose,vanherle2022analysis}, pose estimation~\cite{thalhammer2019sydpose,choithwaniPoseBiasDatasetBias2023,josifovski2018object,tremblay2018falling}, segmentation~\cite{sankaranarayanan2018learning,saleh2018effective,chen2019learning,ros2016synthia,takmaz20233d}, 3D reconstruction~\cite{richardson20163d,hu2021sail,fengLearningAnimatableDetailed2021c,qiuSCULPTORSkeletonConsistentFace2022,daneVcVekEMOCAEmotionDriven2022}, and also for facial tasks~\cite{choithwaniPoseBiasDatasetBias2023,kortylewskiEmpiricallyAnalyzingEffect2018,buechner2023lets,buechner2023improved,richardson20163d}.
We develop a generative interventional framework that fixes possible other confounding factors to isolate the impact of facial asymmetry on expression classification.

\paratitle{Facial Expression Classification.}
Since the standardization of facial expression into six base \emph{emotions} by Ekman~\cite{ekman1992argument}, state-of-the-art performance for automated classification is achieved by end-to-end trained black-box models~\cite{baltrusaitisOpenFaceOpenSource2016,baltrusaitisOpenFaceOpenSource2018,pham2021facial,savchenko2023facial,chenStaticDynamicAdapting2023,wasiARBExAttentiveFeature2023,maoPOSTERSimplerStronger2023,zhouExploringEmotionFeatures2019,daneVcVekEMOCAEmotionDriven2022,Savchenko_2022_CVPRW,savchenko2022classifying}.
While such models reach high performance, their inner workings remain opaque.
Hence, the relationship between facial symmetry and predictions remains unclear, especially in medical contexts like facial palsy~\cite{Buechner20232D3DFace,buchnerFacesVolumesMeasuring2023,banksClinicianGradedElectronicFacial2015,demecoQuantitativeAnalysisMovements2021,katsumiQuantitativeAnalysisFacial2015,knoedlerReliableRapidAutomated2022,ozsoyThreedimensionalObjectiveEvaluation2021,patelFacialAsymmetryAssessment2015}.
To address this uncertainty, we study the effects of facial asymmetry on expression classifiers in a controlled setting: the expression space of 3D Morphable Models~\cite{blanzMorphableModelSynthesis1999,egger3DMorphableFace2020,gerigMorphableFaceModels2018,zhu2023facescape}, more specifically FLAME~\cite{liLearningModelFacial2017}.
While EMOCA~\cite{deng2020disentangled}, an extension of DECA, also relies on the FLAME expression space for classification, our approach takes a different route.
We maximize the logit activation output for each model and emotion combination by leveraging the expression space, ensuring optimal performance. 
We then rely on methods from explainability to perform an in-depth investigation into the model behavior.

\paratitle{Explaining Model Decisions Behavior.}
Local explainability methods, e.g.,~\cite{Springenberg2014StrivingFS, Ribeiro2016WhySI, Selvaraju2016GradCAMVE, Smilkov2017SmoothGradRN, Sundararajan2017AxiomaticAF}, are used to investigate the behavior of machine learning models for singular inputs, e.g., highlight important image regions.
Further, in \cite{lapuschkin2019unmasking}, such local explanations are summarized to form a more conclusive general understanding of a classifier (global explanation).
The focus is put on shortcut biases leading to so-called ``Clever Hans predictors''~\cite{lapuschkin2019unmasking}.
However, such local attribution methods necessitate a semantic interpretation, for example, by a domain expert.
Especially for medical applications, more abstract but relevant features increase the complexity~\cite{Nachbar1994TheAR,rossDevelopmentSensitiveClinical1996,neumannValidierungDeutschenVersion2016}.
We are interested in analyzing facial symmetry, a complex feature not directly part of the input.
Global approaches, for example, \cite{kim2018interpretability,reimers2020determining}, can determine the usage of such abstract features.
Especially, \cite{reimers2020determining} is based on causal principles \cite{reichenbach1956direction,pearl2009causality,peters2017elements} and tests for conditional dependence between the feature and the network predictions given the labels.
In \cite{reimers2021conditional,penzel2022investigating,penzel2023analyzing,buechner2024power}, this approach is applied to various application domains such as skin lesion analysis, digital agriculture, or emotion classification.
However, here we go one step beyond and extend their approach by an interventional framework to generate more in-depth answers according to Pearl's causal hierarchy \cite{bareinboim2022onpearl}.

\paratitle{Synthetic Face Generation.}
Generative models have long been the go-to approach for modeling human faces.
Ranging from parametric 3D Morphable Models (3DMMs)~\cite{blanzMorphableModelSynthesis1999,paysan3DFaceModel2009,gerigMorphableFaceModels2018,egger3DMorphableFace2020}, Active Appearance Models~\cite{cootes2001active,matthews2004active,haase2014instance,gao2010review}, or learned in an entirely data-driven manner~\cite{liLearningModelFacial2017,yangLearningGeneralizedPhysical2024,zhu2023facescape,yang2020facescape,gerigMorphableFaceModels2018,qiuSCULPTORSkeletonConsistentFace2022,wagnerSoftDECAComputationallyEfficient2023}.
The disentanglement of identity, expression, pose, and appearance is a powerful tool for bias identification~\cite{choithwaniPoseBiasDatasetBias2023,kortylewskiEmpiricallyAnalyzingEffect2018} or image manipulation~\cite{tewari2020stylerig,deng2020disentangled,medinMOSTGAN3DMorphable2022,piao2021inverting}.
In contrast, Generative Adversarial Networks prioritize photorealism over control, embedding multiple facial properties into a single latent representation, making it challenging to have specific control over the generation~\cite{buechner2023lets,buechner2023improved,karrasStyleBasedGeneratorArchitecture2019,pumarola2018unsupervised,pumarola2018GANimation,choi2018stargan,perarnau2016invertible}.
Many approaches utilize neural networks to compute the 3DMM parameters from 2D images, either by reconstructing faces~\cite{fengLearningAnimatableDetailed2021c,qiuSCULPTORSkeletonConsistentFace2022,yang2020facescape,yangLearningGeneralizedPhysical2024,wagnerSoftDECAComputationallyEfficient2023,linSingleShotImplicitMorphable2023,guo2020towards} or training in an adversarial manner~\cite{medinMOSTGAN3DMorphable2022,piao2021inverting,tewari2020stylerig,deng2020disentangled}.
We aim to quantify the impact of facial asymmetry on the predictive behavior of expression classifiers.
To achieve this, we prioritize control over photorealism in our generative pipeline using DECA~\cite{fengLearningAnimatableDetailed2021c}.
Hence, we can fix other confounding factors, like appearance, lighting, and pose, at the same time. 
Building upon FLAME~\cite{liLearningModelFacial2017}, we alter the geometric face model to induce subtle variations, thereby creating realistic facial asymmetry.

\section{Evaluating Models by Intervening on Facial Symmetry}
\label{sec:method}
Studying mimicry is crucial when analyzing facial palsy, which impacts the mobility of the facial muscles.
An objective evaluation of the nerve damage is commonly done via data-driven methods~\cite{Buechner20232D3DFace,guntinas2023high,knoedlerReliableRapidAutomated2022,katsumiQuantitativeAnalysisFacial2015}.
In this work, we focus on one facial feature likely impacting the downstream performance of expression classifiers: facial symmetry.
We start by detailing our investigation's causal model and framing the question we are trying to answer in Pearl's causal hierarchy~\cite{bareinboim2022onpearl}.
Afterward, we describe the adapted 3D Morphable Model to perform interventions by changing one face side's geometry.
Lastly, we derive an interpretable score and corresponding significance test to quantify the impact of facial symmetry on a model's prediction.

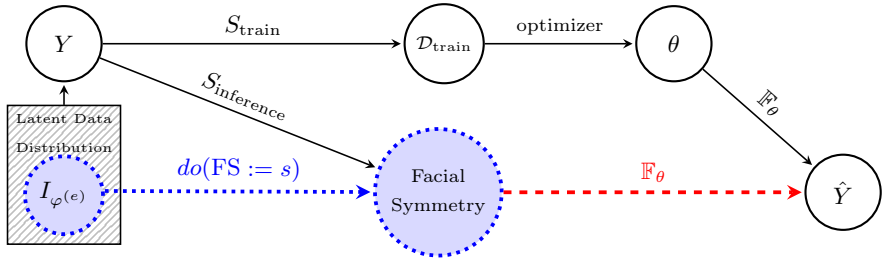
\begin{figure}[t]
    \centering
    \begin{tikzpicture}[
    > = stealth,
    shorten > = 1pt,
    auto,
    node distance = 3cm,
    semithick
]
\tikzstyle{every state}=[
    draw = black,
    thick,
    fill = white,
    minimum size = 10mm
]

\tikzset{FS/.style={
    circle, draw=blue, fill=blue!15, thick,
    }
}
\tikzdeclarepattern{
  name=mylines,
  parameters={
      \pgfkeysvalueof{/pgf/pattern keys/size},
      \pgfkeysvalueof{/pgf/pattern keys/angle},
      \pgfkeysvalueof{/pgf/pattern keys/line width},
  },
  bounding box={
    (0,-0.5*\pgfkeysvalueof{/pgf/pattern keys/line width}) and
    (\pgfkeysvalueof{/pgf/pattern keys/size},
0.5*\pgfkeysvalueof{/pgf/pattern keys/line width})},
  tile size={(0.5\pgfkeysvalueof{/pgf/pattern keys/size},
\pgfkeysvalueof{/pgf/pattern keys/size})},
  tile transformation={rotate=\pgfkeysvalueof{/pgf/pattern keys/angle}},
  defaults={
    size/.initial=5pt,
    angle/.initial=45,
    line width/.initial=.4pt,
  },
  code={
      \draw [line width=\pgfkeysvalueof{/pgf/pattern keys/line width}]
        (0,0) -- (\pgfkeysvalueof{/pgf/pattern keys/size},0);
  },
}

\node
    [state]
    (gt)
    {$Y$};
\node
    [FS,align=center, line width=0.5mm, dotted]
    (x)
    [below right = 1cm and 4cm of gt]
    {\scriptsize Facial\\\scriptsize Symmetry};

\node
    [rectangle,pattern={mylines[size=2pt,line width=.6pt,angle=45]}, pattern color=black!25, draw=black, align=center, anchor=south west]
    (lat)
    [below = 0.3cm of gt]
    {\tiny Latent Data\\\tiny Distribution\\
    \tikz 
        \node
            [FS,align=center,  line width=0.5mm, dotted]
            (int)
            [below = 0.9cm of gt]
            {$\Iphi$};
    };
\node
    [state]
    (ts)
    [right = 4cm of gt]
    {\scriptsize$\mathcal{D}_{\text{train}}$};
    
\node
    [state]
    (p)
    [right = 4cm of x]
    {$\hat{Y}$};

\node
    [state]
    (w)
    [right = 2cm of ts]
    {$\theta$};

\path[black, ->, draw, above, sloped] (lat) -- node {} (gt);
\path[black, ->, draw, above, sloped] (gt) -- node {$S_{\text{train}}$} (ts);
\path[black, ->, draw, above, sloped] (gt) edge node {$S_{\text{inference}}$} (x);
\path[black, ->, draw, above, sloped] (ts) edge node {\scriptsize optimizer} (w);
\path[black, ->, draw, above, sloped] (w) -- node {$\eclass_\ecweights$} (p);
\path[red, dashed, line width=0.5mm, ->, draw, above, sloped] (x) -- node {$\eclass_\ecweights$} (p);
\path[blue, dotted, line width=0.5mm, ->, draw, above, sloped] (int) -- node {$do(\text{FS}:= s)$} (x);
\end{tikzpicture}
    \caption{
    Expression classifier structural causal model:
    $Y$ is the expression influenced by the latent distribution of all facial images (hatched box), $S_*$ samples from this latent distribution~\cite{reimers2020determining}.
    $\mathcal{D}_{\text{train}}$ is the training data distribution.
    The model architecture is an exogenous variable, and weights $\ecweights$ are learned using an optimizer, i.e., a training algorithm.
    The model's predictions $\hat{Y}$ result from the trained model $\eclass_\ecweights$.
    We investigate whether $\eclass_\ecweights$ is independent of the model predictions $\hat{Y}$ ({\color{red} dashed red arrow}).
    Additionally, we analyze the changes in behavior for varying facial symmetries.
    Toward this goal, we perform synthetic interventions ({\color{blue}$do(\text{Facial Symmetry}:= s)$}) on {\color{blue}facial symmetry} variable using 3d morphable models {\color{blue} $\Iphi$}.
    Note that these {\color{blue} $\Iphi$} are a part (subpopulation) of the latent distribution of all facial images.
    Adapted from Figure 2 in \cite{reimers2020determining}.}
    \label{fig:scm}
\end{figure}

\subsection{Preliminaries \& Causal Point of View}
\label{sec:ci_framework}

Causal inference tries to answer causal questions from data~\cite{peters2017elements}.
This includes interventions, i.e., additional experiments and purely observational data.
Importantly, causal questions can be categorized into a hierarchy.
This so-called Pearl's causal hierarchy (PCH)~\cite{bareinboim2022onpearl} consists of the three levels ordered by increasing difficulty: associational, interventional, and counterfactual questions.
The latter two are analyzed using the \emph{do}-operator~\cite{pearl2009causality}, which changes a variable to a constant value, e.g., we write $do(\text{Facial Symmetry}:= s)$ for the variable facial symmetry.

Furthermore, framing data-generating processes and complex interactions of our physical reality as directed graphs enables us to precisely define and investigate the underlying causal mechanisms \cite{pearl2009causality,peters2017elements}.
The resulting models are called structural causal models (SCMs), and we include a formal definition in the supplementary material.
Nevertheless, to understand this framework, it is important to interpret the dependencies, i.e., connections in the graph, as assignments and not as algebraic mappings~\cite{peters2017elements}.
Specifically, the connections between variables in such an SCM function like physical mechanisms and not like instantaneous equations.
This work extends a specific SCM to model supervised learning~\cite{reimers2020determining}.

We visualize our SCM for expression classification in \Cref{fig:scm}, enabling us to study different questions about the decision process.
For example, Reimers et al. \cite{reimers2020determining} answer associational questions of whether a feature, such as facial symmetry, is used during the prediction, i.e., does the red dashed arrow exist in \Cref{fig:scm}.
Intuitively, they measure if there is a statistically significant shift in classifier outputs for inputs of the same class but with different feature manifestations.

Other works visualize such significant changes for feature variations \cite{penzel2023analyzing,buechner2024power}.
However, they lack actionable descriptions of how the model would behave if a particular feature, e.g., facial symmetry, changes for a specific individual.
In this work, we go one step up on the PCH.
We employ a synthetic rendering pipeline to alter the facial symmetry while controlling other factors.
Hence, we answer the interventional question: ``If we change the facial symmetry for an input, then how does the output of the expression classifier behave?''
Please note that the levels of the PCH are disjunct and increasing in difficulty.
In~\cite {bareinboim2022onpearl}, the authors prove the Causal Hierarchy Theorem (CHT), which states that one needs data of at least the corresponding level to answer causal questions of that level.

In the following, we describe how we generate synthetic data ($\Iphi$ in \Cref{fig:scm}), where we have fine-grained control over facial symmetry and realized emotional expressions. 
Using this framework, we generate new interventional data.
Hence, we do not violate the CHT~\cite{bareinboim2022onpearl}.
Further, while our approach of synthetic generation necessarily introduces a domain shift (see \Cref{fig:scm}), we argue that it enables us to go beyond simple interventions.
Specifically, our framework allows us to vary the facial symmetry for a specific individual and measure changes in the classifier outputs while fixing other confounding factors.
Finally, we discuss how we quantify systematic output changes and determine significance.

\subsection{Facial Symmetry Intervention Framework}
\label{sec:eval_frame_work}
We require a controllable face generation method to answer interventional questions of the form: ``If we change the facial symmetry for an input, then how does the output of the emotion classifier behave?''
Additionally, the generation process has to ensure that only facial expressions contribute to the changes measured by the expression classifier.
Therefore, we select a 3D Morphable Model (3DMM)~\cite{blanzMorphableModelSynthesis1999,gerigMorphableFaceModels2018,paysan3DFaceModel2009,liLearningModelFacial2017,egger3DMorphableFace2020}, to be precise FLAME~\cite{liLearningModelFacial2017}, used in the DECA architecture to create synthetic facial images~\cite{fengLearningAnimatableDetailed2021c}.
Although the generated faces introduce a domain shift, the underlying representation of identity, expression, and appearance gives us complete control over individual changes.
Therefore, this disentanglement ensures we can causally link facial changes to the model's predictive behavior.
In the following, we detail our face generation framework to (a) find the expression parameters for optimal classifier activation and (b) introduce a controllable symmetry value $s$ for interventional reasoning.

For all synthetic face images $I$ in this work, we utilize the DECA pipeline \cite{fengLearningAnimatableDetailed2021c}: $I = \mathcal{R}(\mathcal{M}, \mathcal{B}, c)$, composed of the face model $\mathcal{M}$, camera position $c \in \mathbb{R}^{3}$ (fixed to $[0,0,0]^T$ in this work) and illumination process $\mathcal{B}$ used in the differential renderer $\mathcal{R}(\cdot)$~\cite{fengLearningAnimatableDetailed2021c}.
To study facial asymmetry, we alter the face model geometry$\mathcal{M}$ formally defined as $\mathcal{M}(\svec,\pvec,\evec,\avec) = \{\mathcal{G}(\svec,\pvec,\evec), \mathcal{A}(\avec)\}$.
DECA employs the geometric components of FLAME $\mathcal{G}$ using the identity $\svec \in \mathbb{R}^{100}$, expression $\evec \in \mathbb{R}^{50}$, and pose $\pvec \in \mathbb{R}^{6}$ blendshape parameters~\cite{liLearningModelFacial2017}.
The texture is computed from the appearance model $\mathcal{A}$ from the Basel Face Model using the parameter $\avec$~\cite{paysan3DFaceModel2009,gerigMorphableFaceModels2018,fengLearningAnimatableDetailed2021c}.
In FLAME, the face geometry is modeled as 
\begin{equation}
    \mathcal{G}(\svec,\pvec,\evec) = W(T + B_{I}(\svec, \mathcal{I}) + B_{P}(\pvec, \mathcal{P}) + B_{E}(\evec, \mathcal{E}), J(\pvec), \pvec, \mathcal{W}),
    \label{eq:face_model}
\end{equation}
with $W$ being a standard skinning function to rotate the modified $N$ face vertices of the template model $T \in \mathbb{R}^{N \times 3}$ around predefined FLAME joints $J\in \mathbb{R}^{3K}$.
$B$ denotes a linear blend skinning (LBS)~\cite{lewis2000posespace} function of the according blend shapes with identity $\mathcal{I} \in \mathbb{R}^{100 \times N \times 3}$, expression $\mathcal{E} \in \mathbb{R}^{50 \times N \times 3}$, and pose $\mathcal{P} \in \mathbb{R}^{6 \times N \times 3}$.
We use the blending weights $\mathcal{W}\in \mathbb{R}^{K \times N}$ of the original FLAME model~\cite{liLearningModelFacial2017}.

Our changes must ensure that (a) under full facial symmetry, the original geometry holds, and (b) a symmetry scalar $s$ specifies facial symmetry and enables interventional queries.
Furthermore, we formalize a time parameter $t$ to control the interpolation between \neutral and a \emph{target} facial expressions~\cite{linSingleShotImplicitMorphable2023,egger3DMorphableFace2020}.

We extend a recent approach by freezing geometry parts to simulate facial asymmetry~\cite{yangLearningGeneralizedPhysical2024}.
Using a scaling parameter $s$, we can simulate different \emph{freeze} states ranging from $0.0$ defining complete asymmetry to $1.0$ defining complete symmetry.
We artificially induce facial asymmetry by changing only the left side of the face (person's point of view).
Therefore, we recompose the FLAME expression space such that $B_{E}(\evec, \mathcal{E}) = B_{E}(\evec, \mathcal{E}^{L}) + B_{E}(\evec, \mathcal{E}^{R})$.
Thus, we define
\begin{equation}
    \mathcal{E}^{L}_{i} =
    \begin{cases}
        \mathcal{E}_{i},    & \text{if the vertex $i$ is on the left side of the face} \\
        0,                  & \text{otherwise}
    \end{cases}
\end{equation}
such that the linear blend skinning function $B_E(\evec, \mathcal{E}^{L})$ changes only vertices on the left side of the face~\cite{lewis2000posespace,yangLearningGeneralizedPhysical2024}.
The same applies to $\mathcal{E}^{R}$.
Scaling the blendshape vectors in $\mathcal{E}^{L}$ with $s$ induces a symmetry difference between the faces' sides.
Lastly, we multiply the expression parameters $\evec$ with $t$ to create dynamic expressions.
Our geometric face model $\mathcal{G}_{s,t}$ with symmetry parameter is
\begin{align}
\begin{split}
    \mathcal{G}_{s,t}(\svec,\pvec,\evec,s,t) = &W(T + B_{S}(\beta, \mathcal{S}) + B_{P}(\pvec, \mathcal{P}) + \\
    &B_{E}(t\cdot\evec, \mathcal{E}^{R}) + B_{E}(t\cdot\evec, s\cdot\mathcal{E}^{L}), J(\pvec), \pvec, \mathcal{W}).
\end{split}
    \label{eq:face_model_asym_time}
\end{align}

\begin{figure}[t]
    \centering
    \begin{subfigure}[b]{0.13\textwidth}
        \centering
        \includegraphics[width=\textwidth]{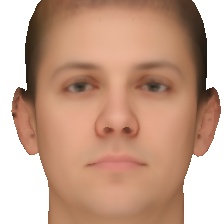}
        \caption{\scriptsize{\neutral}}
        \label{fig:rmn_neutral}
    \end{subfigure}
    \hfill
    \begin{subfigure}[b]{0.13\textwidth}
        \centering
        \includegraphics[width=\textwidth]{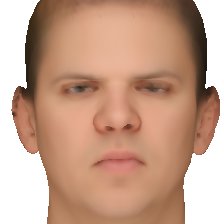}
        \caption{\scriptsize{\angry}}
        \label{fig:rmn_angry}
    \end{subfigure}
    \hfill
    \begin{subfigure}[b]{0.13\textwidth}
        \centering
        \includegraphics[width=\textwidth]{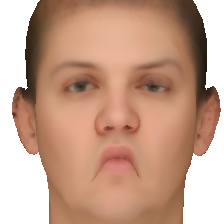}
        \caption{\scriptsize{\disgust}}
        \label{fig:rmn_disgust}
    \end{subfigure}
    \hfill
    \begin{subfigure}[b]{0.13\textwidth}
        \centering
        \includegraphics[width=\textwidth]{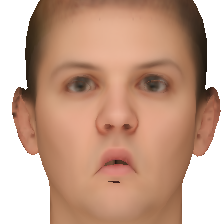}
        \caption{\scriptsize{\fear}}
        \label{fig:rmn_fear}
    \end{subfigure}
    \hfill
    \begin{subfigure}[b]{0.13\textwidth}
        \centering
        \includegraphics[width=\textwidth]{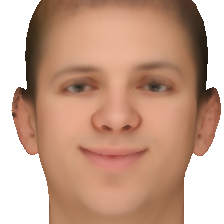}
        \caption{\scriptsize{\happy}}
        \label{fig:rmn_happy}
    \end{subfigure}
    \hfill
    \begin{subfigure}[b]{0.13\textwidth}
        \centering
        \includegraphics[width=\textwidth]{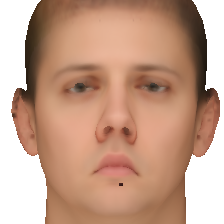}
        \caption{\scriptsize{\sad}}
        \label{fig:rmn_sad}
    \end{subfigure}
    \hfill
    \begin{subfigure}[b]{0.13\textwidth}
        \centering
        \includegraphics[width=\textwidth]{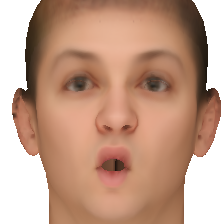}
        \caption{\scriptsize{\surprise}}
        \label{fig:rmn_surprise}
    \end{subfigure}
    \begin{subfigure}[b]{0.13\textwidth}
        \centering
        \includegraphics[width=\textwidth]{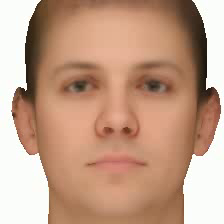}
        \caption{$t=\nicefrac{0}{6}$}
        \label{fig:hse_happy_s0_t00}
    \end{subfigure}
    \hfill
    \begin{subfigure}[b]{0.13\textwidth}
        \centering
        \includegraphics[width=\textwidth]{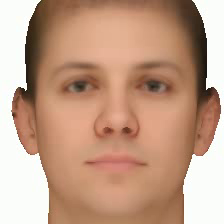}
         \caption{$t=\nicefrac{1}{6}$}
        \label{fig:hse_happy_s0_t15}
    \end{subfigure}
    \hfill
    \begin{subfigure}[b]{0.13\textwidth}
        \centering
        \includegraphics[width=\textwidth]{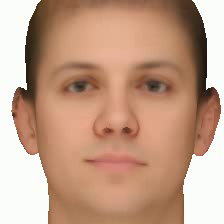}
        \caption{$t=\nicefrac{2}{6}$}
        \label{fig:hse_happy_s0_t30}
    \end{subfigure}
    \hfill
    \begin{subfigure}[b]{0.13\textwidth}
        \centering
        \includegraphics[width=\textwidth]{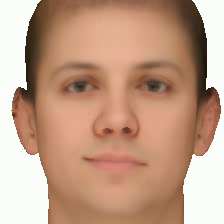}
        \caption{$t=\nicefrac{3}{6}$}
        \label{fig:hse_happy_s0_t45}
    \end{subfigure}
    \hfill
    \begin{subfigure}[b]{0.13\textwidth}
        \centering
        \includegraphics[width=\textwidth]{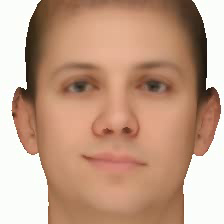}
        \caption{$t=\nicefrac{4}{6}$}
        \label{fig:hse_happy_s0_t60}
    \end{subfigure}
    \hfill
    \begin{subfigure}[b]{0.13\textwidth}
        \centering
        \includegraphics[width=\textwidth]{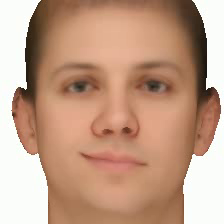}
        \caption{$t=\nicefrac{5}{6}$}
        \label{fig:hse_happy_s0_t75}
    \end{subfigure}
    \hfill
    \begin{subfigure}[b]{0.13\textwidth}
        \centering
         \includegraphics[width=\textwidth]{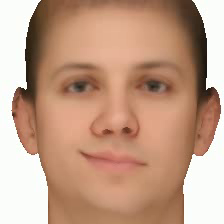}
        \caption{$t=\nicefrac{6}{6}$}
        \label{fig:hse_happy_s0_t89}
    \end{subfigure}
    \caption{
        We display the optimized synthetic face images $\Iphi$ for the \neutral expression (\subref{fig:rmn_neutral}) and for the six base emotions (\subref{fig:rmn_angry}) - (\subref{fig:rmn_surprise}) based on the ResidualMaskingNet classifier~\cite{pham2021facial}.
        Furthermore, we simulate with our geometric face model $\mathcal{G}_{s,t}(\cdot)$ different interpolations $t$ for a symmetry of $s=0.0$.
        At $t=0.0$ (\subref{fig:hse_happy_s0_t00}) we have a \neutral expression morphing into an asymmetric \happy expression at $t=1.0$ (\subref{fig:hse_happy_s0_t89}).
    }
    \label{fig:simulated_flame_emotions}
\end{figure}

Thus, the synthetic face image $\Iphi$ updates for a single individual, i.e., $\ivec$ and $\avec$ are fixed (omitted for clarity), and target expression ($\emo$) vector $\evec^{(\emo)}$ with symmetry $s$ and temporal dynamic $t$ to $\Iphi(s,t) = \mathcal{R}(\mathcal{G}_{s,t}(\svec,\pvec,\evec^{(\emo)},s,t), \mathcal{A}(\avec), \mathcal{B}, c)$.

Within this framework, for an individual, we can (a) modify expressions, (b) simulate facial asymmetry, and (c) simulate movements, ensuring that only changes in facial expression result in changes in the classifier's behavior.
While our synthetic faces are a domain shift for most classifiers, the comparisons are all relative and contained within this new domain.
Hence, representing out-of-domain scenarios in which they are applied~\cite{savchenko2022classifying,Savchenko_2022_CVPRW,savchenko2023facial,pham2021facial,buechner2023lets,buechner2023improved}. 
Furthermore, we optimize the facial expression parameters such that a classifier output $\eclass^{(\emo)}_{\ecweights}$ correctly identifies the given image $\Iphi$ as the target emotion via
\begin{equation}
    \evec^{(\emo)} = \argmin_{\hat{\evec}^{(\emo)} \in [-3, 3]^{|\evec^{(\emo)}|}} 1 - \eclass_\ecweights^{(\emo)}(I_{\hat{\evec}^{(\emo)}}).
    \label{eq:optim_facial_expressions}
\end{equation}
For this estimation problem, all parameters apart from $\evec$ are fixed during the optimization, minimizing other confounding factors~\cite{buechner2024power,eggerIdentityExpressionAmbiguity3D2021a,weihererApproximatingIntersectionsDifferences2024,choithwaniPoseBiasDatasetBias2023}, enforcing that only changes in facial expression influence the classifier output.
Given that we cannot use a gradient-based optimizer as changes in $\evec$ result in no changes in the parameters of $\eclass_\ecweights$, we use the \emph{differential evolution algorithm}~\cite{stornDifferentialEvolutionSimple1997} for optimization using a search range of $[-3,3]$~\cite{liLearningModelFacial2017,egger3DMorphableFace2020}.
In~\Cref{fig:simulated_flame_emotions}, we visualize renderings for the six base emotions given the ResidualMaskingNet as classifier~\cite{pham2021facial}.
The supplementary material provides more examples and expressions parameters $\evec^{(\emo)}$.

\subsection{Measuring Systematic Change}
\label{sec:score}

Given our rendering pipeline $\Iphi$, specified in the previous section, we need a score function to measure systematic changes in expression classifier behavior concerning facial symmetry.
Hence, we define a facial symmetry impact score for a specific trained model $\eclass_\ecweights$.
To be precise, we measure one score for each possible expression $\emo$ predicted by the selected classifier, henceforth, $\eclass_\ecweights^{(\emo)}$.

\begin{figure}[t]
    \centering
    \begin{subfigure}[t]{0.32\textwidth}
        \centering
        \includegraphics[width=\textwidth,trim={0 1.0cm 0 2.5cm},clip]{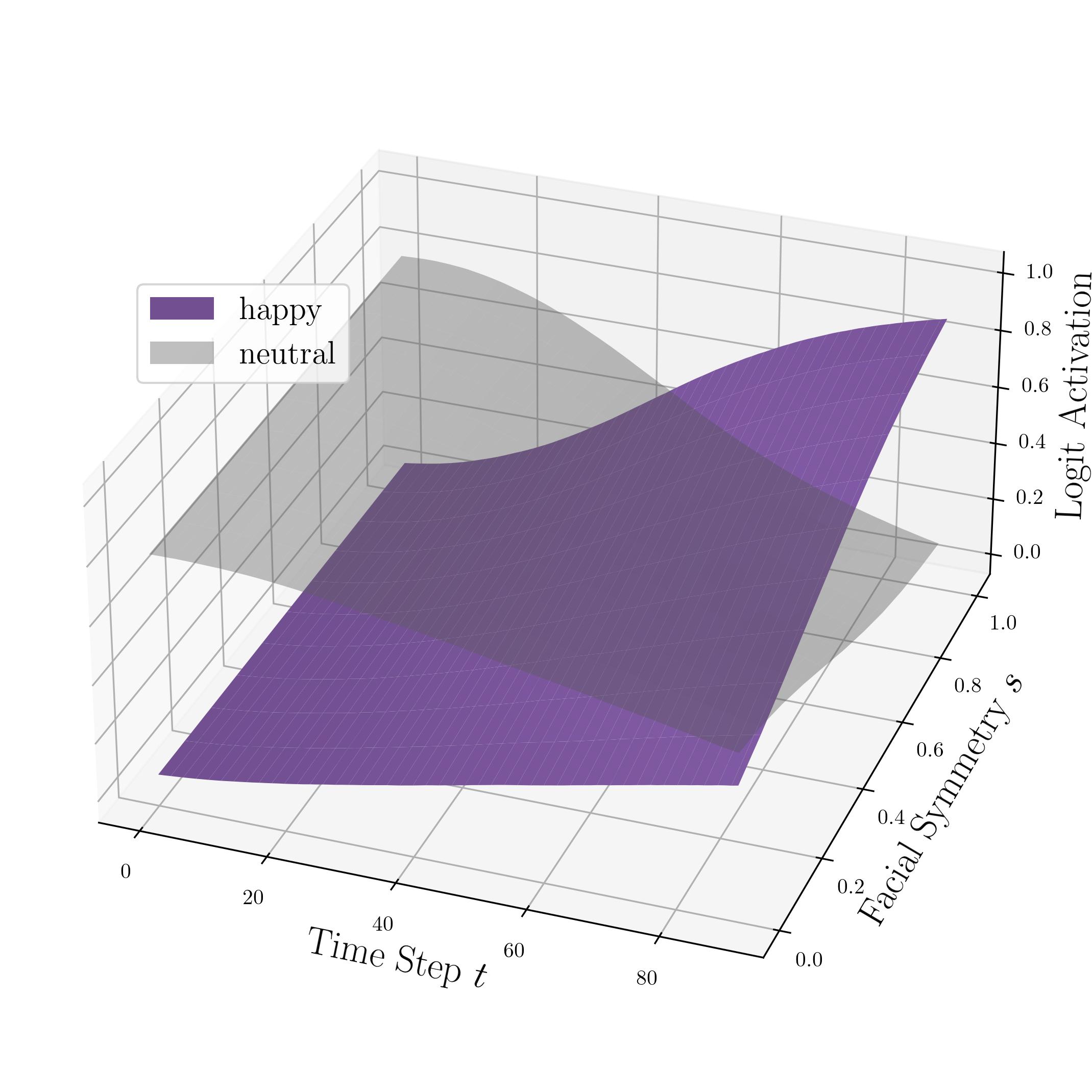}
        \caption{Logit activations from \neutral towards \happy}
        \label{fig:fais_vis_a}
    \end{subfigure}
    \hfill
    \begin{subfigure}[t]{0.32\textwidth}
        \centering
        \includegraphics[width=\textwidth,trim={0 1.0cm 0 2.5cm},clip]{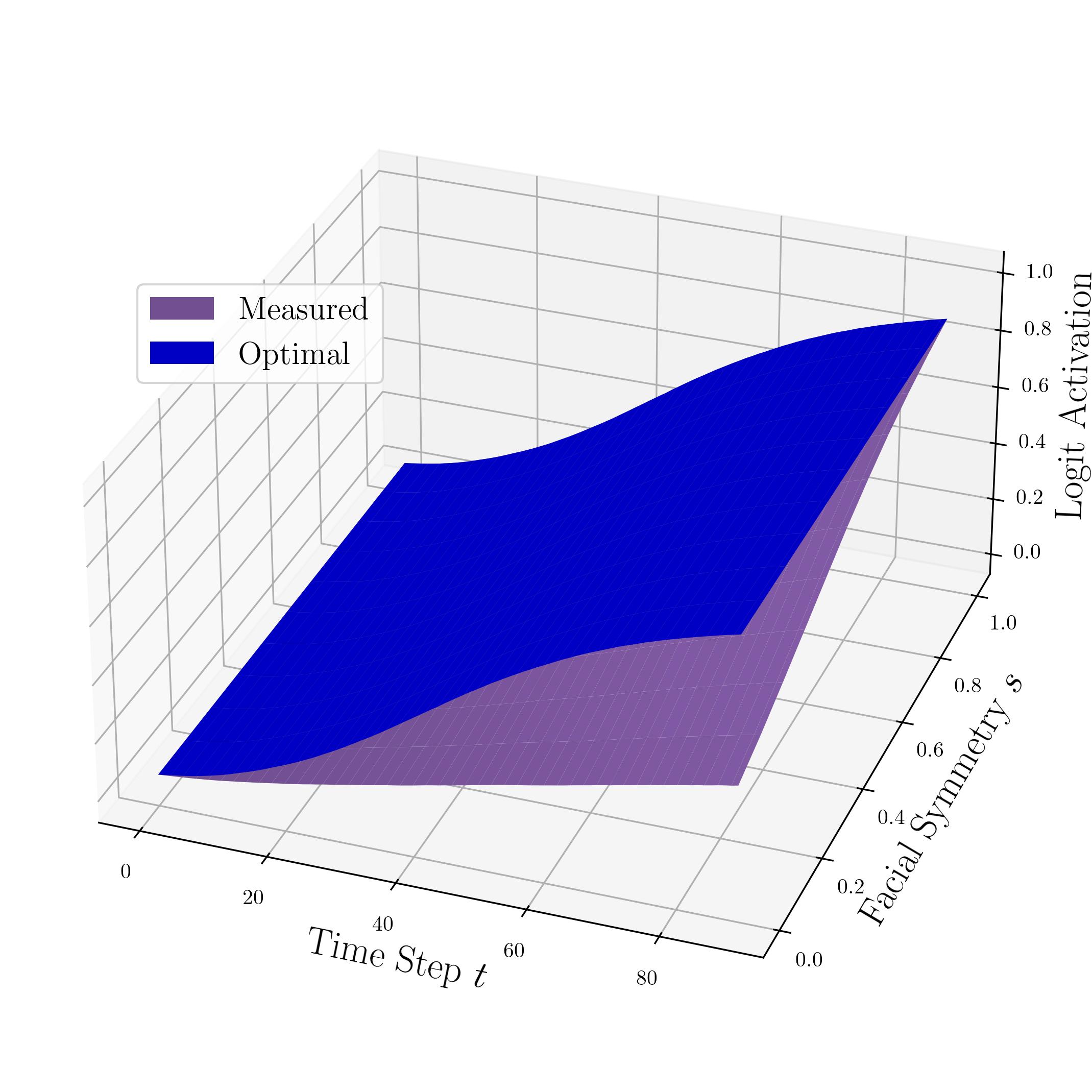}
        \caption{{\color{blue} Optimal} compared to the {\color{happy} measured} activation}
        \label{fig:fais_vis_b}
    \end{subfigure}
    \hfill
    \begin{subfigure}[t]{0.32\textwidth}
        \centering
        \includegraphics[width=\textwidth,trim={0 1.0cm 0 2.5cm},clip]{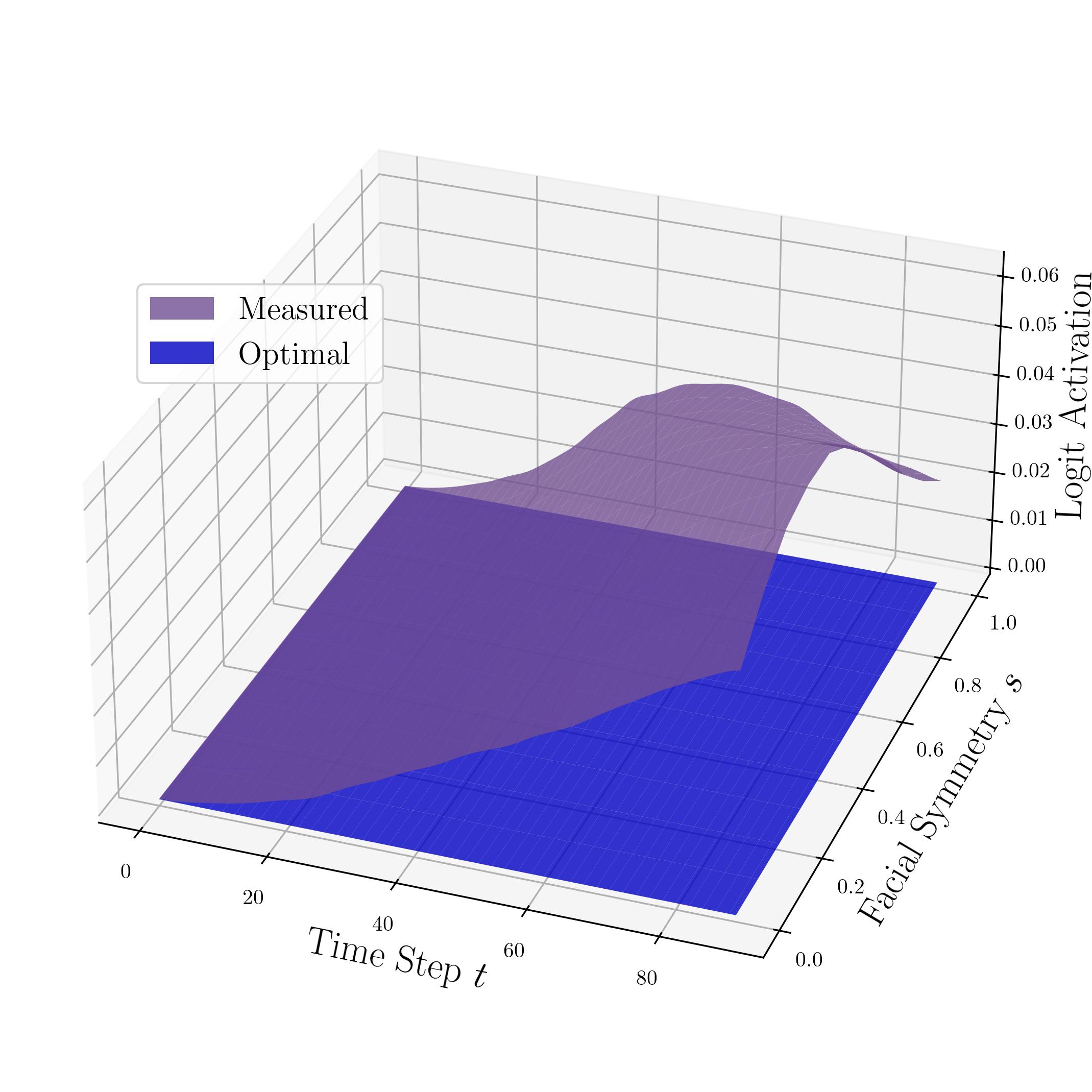}
        \caption{Computed $\nabla_{s}\eclass$ of the {\color{blue} optimal} and {\color{happy} measured} surfaces}
        \label{fig:fais_vis_c}
    \end{subfigure}
    \caption{
        Visualization of our impact score for a classifier's \happy logit activation:
        In a synthetic setting, a model was shown a face transition from \neutral to a \happy expression~(\subref{fig:fais_vis_a}).
        A model would be invariant toward changes along the symmetry axis if $\nabla_{s}\eclass = 0$. 
        However, the actual activation logits ({\color{happy}\emph{happy}}) show a lower activation~(\subref{fig:fais_vis_b}).
        This is more evident in the visualization of the estimated $\nabla_s$ in (\subref{fig:fais_vis_c}).
    }
    \label{fig:fais_vis}
\end{figure}

Using $\Iphi$ with a sampled identity, i.e., fixed $\avec$ and $\ivec$, we generate synthetic images for timesteps $t$ and facial symmetries $s$.
Now, $\eclass_\ecweights^{(\emo)}(\Iphi (s, t))$ defines a surface, where for each $s$ and $t$, we have an output activation of $\eclass_\ecweights$ for emotion~$\emo$.
\Cref{fig:fais_vis_a} visualizes two of these surfaces for the \neutral and \happy emotion.
Ideally, we would want to see no changes along the $s$ axis in these surfaces, i.e., the model is unbiased concerning symmetry.
We can measure these changes by investigating the partial derivatives $\nabla_s \eclass_\ecweights^{(\emo)}(\Iphi(s,t))$. 
A positive $\nabla_s$ indicates higher model outputs for increased symmetry, which is reversed for negative $\nabla_s$.
An unbiased model activation surface ({\color{blue}blue}) is visualized in \Cref{fig:fais_vis_b}.
This optimal surface is characterized by $\nabla_s \eclass_\ecweights^{(\emo)}$ being zero for any valid $s$ and $t$.

Of course, in reality, we do not expect the outputs of any model to stay constant for changing symmetry values.
Many factors can impact the model outputs, even for small visual changes.
Nevertheless, we expect an unbiased model to show no systematic behavioral changes, e.g., categorically lower outputs for smaller symmetry values $s$.
Hence, a more realistic ideal surface would be a noisy version of the visualization in \Cref{fig:fais_vis_b}.
In other words, for an unbiased model that does not change behavior for different facial symmetry, we expect that $\mathbb{E}_{s,t}[\nabla_s \eclass_\ecweights^{(\emo)}(\Iphi(s,t))]$ is approximately zero for some joint distribution of symmetry values $s$ and timesteps $t$.
Without loss of generality, let $[0,1]$ be a valid domain for $s$ and $t$ respectively, then we define our facial symmetry impact score $\Sym$ for a specific model $\eclass_\ecweights^{(\emo)}$ and for a fixed individual $\Iphi$ as
\begin{align}\label{eq:localS}
\begin{split}
    \Sym(\eclass_\ecweights^{(\emo)} | \Iphi)  &= \mathbb{E}_{s,t}[\nabla_s \eclass_\ecweights^{(\emo)}(\Iphi(s,t))]\\
            &= \iint_0^1 \nabla_{s} \eclass_\ecweights^{(\emo)}(\Iphi(s,t)) \cdot p(s,t) \,dt\,ds,
\end{split}
\end{align}
where $p$ is the density function describing the joint distribution of $s$ and $t$.

Calculating $\Sym(\eclass_\ecweights^{(\emo)} | \Iphi)$ directly is intractable.
Hence, we assume that $s$ and $t$ are independent and uniformly distributed.
Although this is a strong assumption, we can utilize our rendering pipeline (see \Cref{sec:eval_frame_work}) to ensure these conditions in our synthetic data.
By doing so, we can approximate $\Sym(\eclass_\ecweights^{(\emo)} | \Iphi)$ by evaluating $\eclass_\ecweights^{(\emo)}$ at a grid of finitely many equidistant samples of $\Iphi(s,t)$.

Let $\mathfrak{T}$ and $\mathfrak{S}$ be a set of equidistant time and symmetry steps in $[0,1]$, then
\begin{equation}\label{eq:localShat}
    \hat{\Sym}(\eclass_\ecweights^{(\emo)} | \Iphi) = \frac{1}{|\mathfrak{S}| \cdot |\mathfrak{T}|} \sum_{s\in \mathfrak{S}} \sum_{t\in \mathfrak{T}} \nabla_s \eclass_\ecweights^{(\emo)}(\Iphi(s,t)),
\end{equation}
approximates $\Sym(\eclass_\ecweights^{(\emo)} | \Iphi)$.
To estimate the gradient on our finite grid of $\mathfrak{T}$ and $\mathfrak{S}$, we use the implementation of \cite{fornberg1988generation} by the library \texttt{NumPy} \cite{harris2020array}. 
This algorithm minimizes the error between the actual gradient and the estimate at a grid position by solving a system of linear equations of the neighboring grid points.

While $\hat{\Sym}(\eclass_\ecweights^{(\emo)} | \Iphi)$ enables us to investigate changes for a single individual $\Iphi$ with fixed $\avec$ and $\ivec$, we are additionally interested in explaining models $\eclass_\theta$ more globally concerning specific emotions.
Hence, we define a global score for an emotion $\emo$ as $\Sym(\eclass_\ecweights^{(\emo)}) = \mathbb{E}_{\avec,\ivec}[\Sym(\eclass_\ecweights^{(\emo)} | \Iphi)]$.
For a set of $N$ individuals $\mathfrak{I}$, by using the same assumptions as in \Cref{eq:localShat}, we approximate $\Sym(\eclass_\ecweights^{(\emo)})$ with
\begin{equation}\label{eq:globalShat}
    \hat{\Sym}(\eclass_\ecweights^{(\emo)}) = \frac{1}{N} \sum_{\Iphi\in \mathfrak{I}} \hat{\Sym}(\eclass_\ecweights^{(\emo)} | \Iphi).
\end{equation}

\subsubsection{Testing for Statistical Significance:}
Values for $\hat{\Sym}(\eclass_\ecweights^{(\emo)})$ and $\hat{\Sym}(\eclass_\ecweights^{(\emo)}|\Iphi)$ close to zero indicate less change for varying facial symmetry values $s$.
Further, the sign of our scores can be interpreted as over- (positive) or under-predicting (negative) an emotion for increasing symmetry.
However, we also need to specify at which point values of $\hat{\Sym}(\eclass_\ecweights^{(\emo)})$ or $\hat{\Sym}(\eclass_\ecweights^{(\emo)}|\Iphi)$ are statistically significant.

We utilize permutation hypothesis tests~\cite{good2013permutation} for this goal in which the values of $\eclass_\ecweights^{(\emo)}$ in our grid of synthetic inputs are shuffled.
To control for the influence of $t$, i.e., the onset expression's strength, we only shuffle values of $\eclass_\ecweights^{(\emo)}$ while fixing $t$.
In other words, we permute along the symmetry axis in \Cref{fig:fais_vis_a}.
Afterward, the corresponding $\hat{\Sym}(\eclass_\ecweights^{(\emo)}|\Iphi^{(perm.)})$ is recalculated (\Cref{eq:localShat}).
The process repeats $K$-times to generate our distribution under the null hypothesis $H_0$, which is that the observed value $\hat{\Sym}(\eclass_\ecweights^{(\emo)}|\Iphi)$ is zero.
By counting how often we observe permutations where $|\hat{\Sym}(\eclass_\ecweights^{(\emo)}|\Iphi^{(perm.)})| > |\hat{\Sym}(\eclass_\ecweights^{(\emo)}|\Iphi)|$, we can determine a $p$-value.
Note that the absolutes are needed because negative scores are valid.
If the $p$-value is smaller than a significance level $\siglevel$, we discard $H_0$, i.e., the observed score $\hat{\Sym}(\eclass_\ecweights^{(\emo)}|\Iphi)$ is statistically significant.
We provide the hypothesis test pseudo-code in the supplementary material (Alg. 1).

While our approach tests for significance concerning a certain $\Iphi$ with fixed $\avec$ and $\ivec$, regarding $\hat{\Sym}(\eclass_\ecweights^{(\emo)})$, we additionally apply the Holm-Bonferroni correction on our results~\cite{holm1979simple}.
This is a sequential correction method to control the family-wise error rate for repeated hypothesis tests.
In other words, ensuring we do not overestimate significance, i.e., increase type-I errors, for a pre-specified $\siglevel$.
Finally, we report the ratio of significant results of the corrected tests, which intuitively captures how often we observe significant changes in behavior.
Note that, in contrast, $\hat{\Sym}(\eclass_\ecweights^{(\emo)})$ measures how strong (and in which direction) these changes are on average.
This means statistical significance is possible even if the effect size, i.e., the actual systematic change, is low.
With the scores derived above combined with the corresponding statistical hypothesis tests, we can investigate the interventional question posed in \Cref{sec:ci_framework}:  ``If the facial symmetry for an input changes, then how does the output of the expression classifier behave?''

\section{Experiments and Results}
Our investigation focuses on the impact of facial symmetry on data-driven expression classifiers.
Before we detail our experiments, we want to state our hypothesis clearly:
We expect that most classifiers show systematic differences in their behavior when intervening on facial symmetry.
Given that one face half exhibits less movement for unilateral facial palsy and our synthetic symmetry data, we assume a reduction in logit-activations for reduced symmetry based on observations in related studies~\cite{buechner2024power,yang2023change}.
Because most facial expression datasets, e.g.,~\cite{dumitru2013fer,li2017reliable,li2019reliable,Mollahosseini2019affectnet}, contain mostly healthy symmetric faces.
Our study is limited to the six base emotions~\cite{ekman1992argument}, omitting \emph{neutral} and \emph{contempt} for comparability.

We can assess existing expression classifiers, whereas other research requires training~\cite{kortylewskiEmpiricallyAnalyzingEffect2018,choithwaniPoseBiasDatasetBias2023}.
Please note that our selection of models is not intended to be comprehensive or representative of all possible classifiers but as a first case study.
We focus on a subset of models that provide code and model weights, which are likely to be applied in out-of-domain scenarios like medicine and psychology.
To test our hypothesis, we perform two groups of experiments.
First, we investigate if a symmetry bias exists for real-world data inside classifiers using associational methods~\cite{reimers2020determining,reimers2021debiasing,penzel2023analyzing,buechner2023lets}.
Second, moving up on the causal hierarchy~\cite{bareinboim2022onpearl}, we link facial symmetry and model output using our synthetic intervention framework.

\subsection{Experiment 1: Observations on Real-World Data}
\label{sec:exp2}
\begin{figure}[tb]
    \centering
    \begin{subfigure}[b]{0.13\textwidth}
        \centering
        \captionsetup{justification=centering}
        \includegraphics[width=\textwidth]{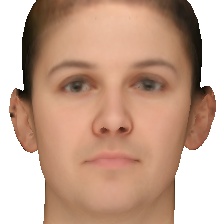}
        \caption{\scriptsize{Synth. \neutral}}
        \label{fig:fake1}
    \end{subfigure}
    \hfill
    \begin{subfigure}[b]{0.13\textwidth}
        \centering
        \captionsetup{justification=centering}
        \includegraphics[width=\textwidth]{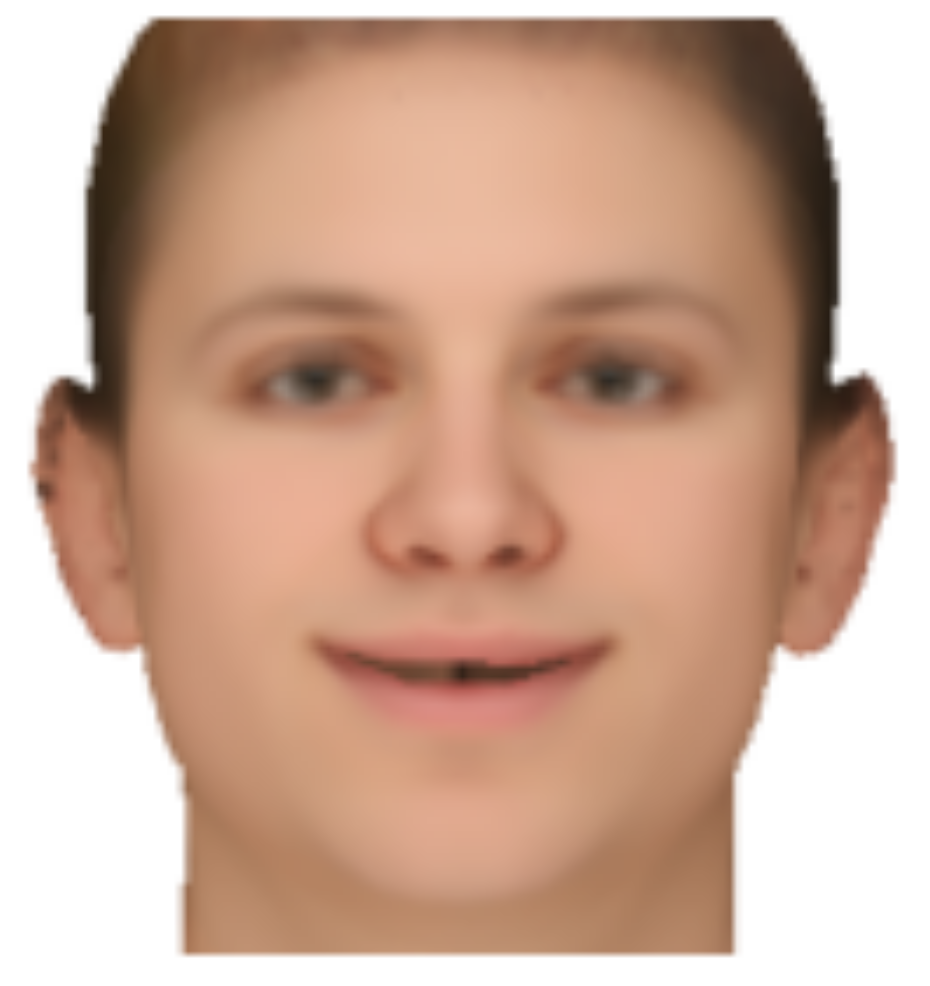}
        \caption{\scriptsize{Synth. \happy}}
        \label{fig:fake2}
    \end{subfigure}
    \hfill
    \begin{subfigure}[b]{0.13\textwidth}
        \centering
        \captionsetup{justification=centering}
        \includegraphics[width=\textwidth,clip,trim={0.5cm 0.5cm 0.5cm 0.5cm}]{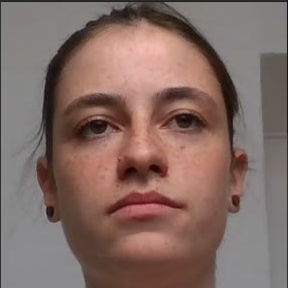}
        \caption{\scriptsize{Proband \neutral}}
        \label{fig:prob1}
    \end{subfigure}
    \hfill
    \begin{subfigure}[b]{0.13\textwidth}
        \centering
        \captionsetup{justification=centering}
        \includegraphics[width=\textwidth,clip,trim={0.5cm 0.5cm 0.5cm 0.5cm}]{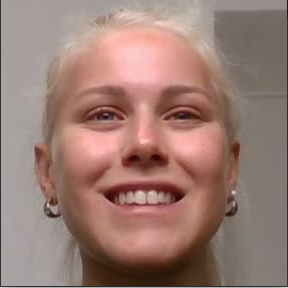}
        \caption{\scriptsize{Proband \happy}}
        \label{fig:prob2}
    \end{subfigure}
    \hfill
    \begin{subfigure}[b]{0.13\textwidth}
        \centering
        \captionsetup{justification=centering}
        \includegraphics[width=\textwidth,clip,trim={3cm 3cm 3cm 3cm}]{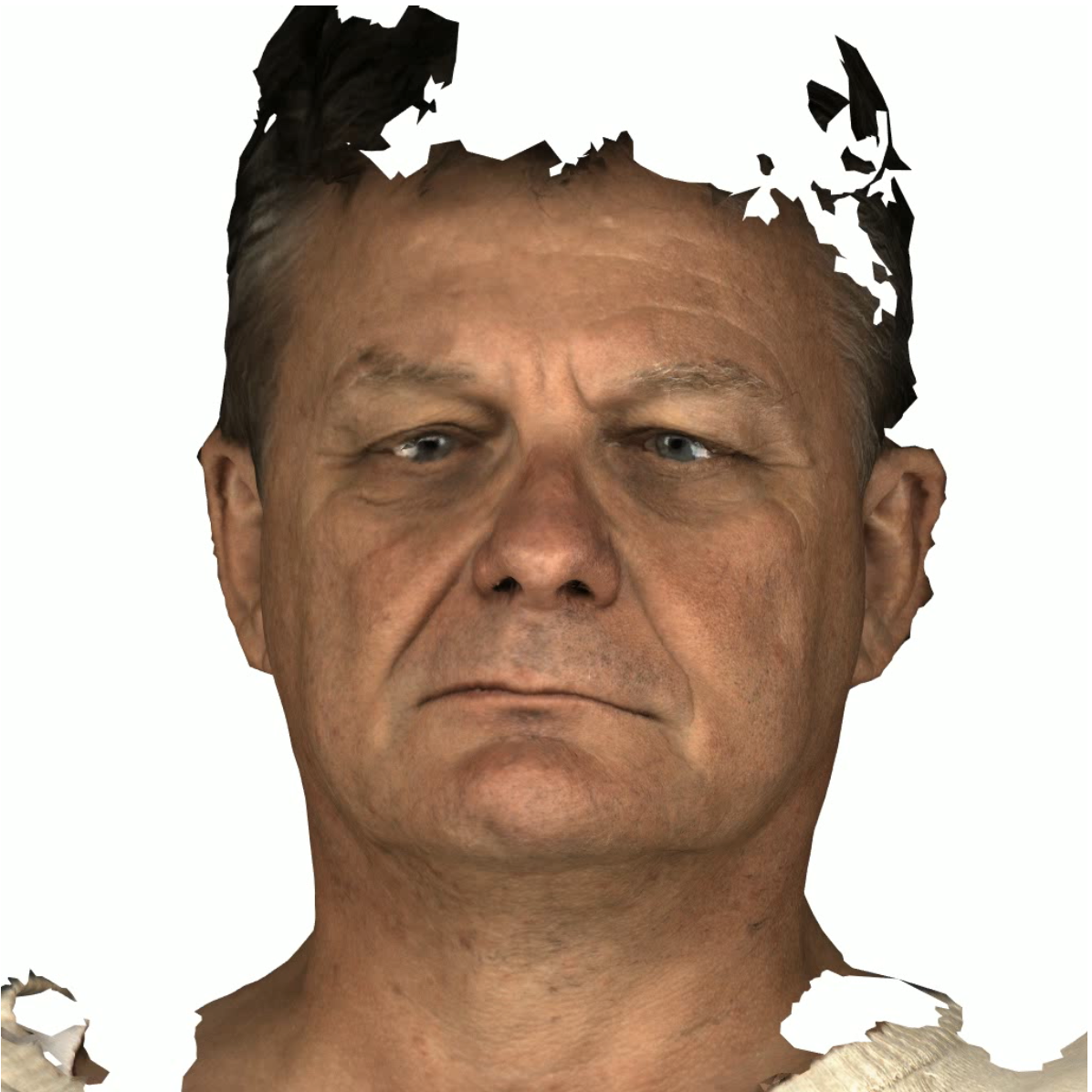}
        \caption{\scriptsize{Patient \neutral}}
        \label{fig:pat1}
    \end{subfigure}
    \hfill
    \begin{subfigure}[b]{0.13\textwidth}
        \centering
        \captionsetup{justification=centering}
        \includegraphics[width=\textwidth,clip,trim={3cm 3cm 3cm 3cm}]{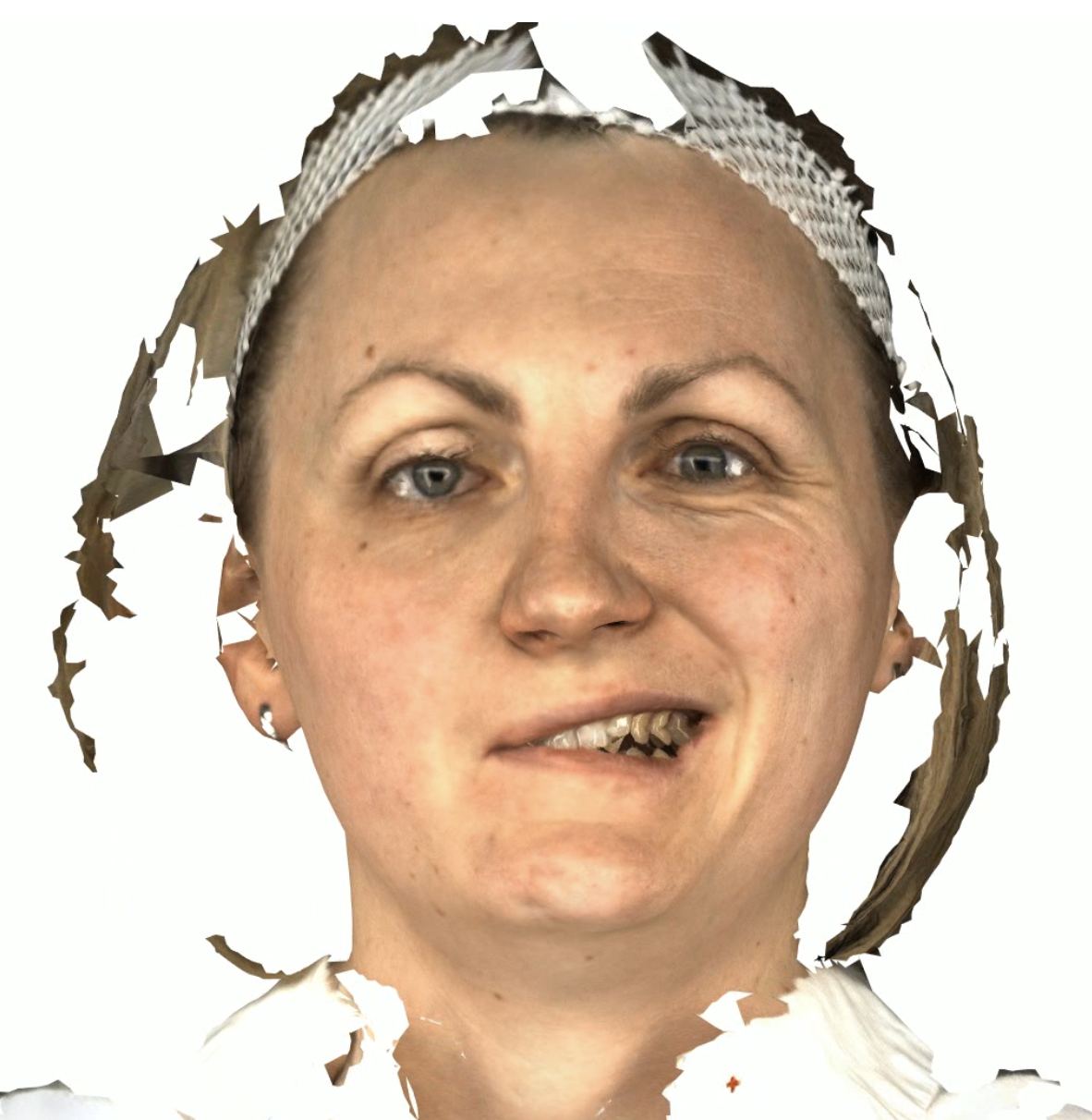}
        \caption{\scriptsize{Patient \happy}}
        \label{fig:pat2}
    \end{subfigure}
    \caption{
        We display two example faces per analyzed group (synthetic, probands, and patients).
        Both healthy probands and patients with unilateral facial palsy were instructed to mimic the shown emotions, similar to the FER2013 benchmark~\cite{dumitru2013fer}.
    }
    \label{fig:experiments_facial_palsy}
\end{figure}

Although intuition suggests that facial asymmetry influences model behavior, we must first check if this bias is present in expression classifiers.
We utilize associational methods~\cite{reimers2020determining,reimers2021debiasing,penzel2023analyzing,buechner2023lets} to show that classifiers significantly change their behavior, i.e., the red arrow exists in \Cref{fig:scm}.
Specifically, we attempt to validate our hypothesis on the first level of Pearl's hierarchy \cite{bareinboim2022onpearl}, using real-world data recorded on 36 healthy probands (18-67 years, 17 \mars, 19 \venus) and 36 patients (25-72 years, 8 \mars, 28 \venus) with unilateral chronic synkinetic facial palsy.
Probands were recorded using a RealSense camera (Intel Corporation, Santa Clara, USA) and patients with the 3dMD face system (3dMD LLC, Georgia, USA), see \Cref{fig:experiments_facial_palsy}.
We model two symmetry properties: the presence of facial palsy (binary) and LPIPS~\cite{zhangUnreasonableEffectivenessDeep2018} similarity (continuous) among face sides.

The participants' expressions are captured while mimicking the six basic emotions four times in a random order~\cite{guntinas2023high,buechner2023improved,buechner2024power,buechner2023lets} following FER2013~\cite{dumitru2013fer}.
We focus on the \happy expression as it most impacts emotional expressiveness~\cite{roberts1996empathy,thompson1987empathy,buchnerFacesVolumesMeasuring2023}.
Examples of each face group are shown in \Cref{fig:experiments_facial_palsy}.
We measure an average classification accuracy of $97.40\%$ for probands and $58.25\%$ for patients among the FER2013 models, indicating the decreased performance under facial palsy.
A more detailed analysis can be found in the supplementary material.
\begin{table}[t]
    \setlength{\tabcolsep}{2.5pt}
\begin{tabular}{lrccl@{\hspace{6\tabcolsep}}rccl@{\hspace{6\tabcolsep}}rccccl@{\hspace{6\tabcolsep}}rcl}
& \mcrot{1}{l}{60}{\scriptsize DAN~\cite{wenDistractYourAttention2023}} 
& \mcrot{1}{l}{60}{\scriptsize DDAMFN++~\cite{zhangDualDirectionAttentionMixed2023}}
& \mcrot{1}{l}{60}{\scriptsize HSEmotion~\cite{savchenko2023facial}}
& \mcrot{1}{l}{60}{\scriptsize PosterV2~\cite{maoPOSTERSimplerStronger2023}}

& \mcrot{1}{l}{60}{\scriptsize DAN~\cite{wenDistractYourAttention2023}} 
& \mcrot{1}{l}{60}{\scriptsize DDAMFN++~\cite{zhangDualDirectionAttentionMixed2023}}
& \mcrot{1}{l}{60}{\scriptsize HSEmotion~\cite{savchenko2023facial}}
& \mcrot{1}{l}{60}{\scriptsize PosterV2~\cite{maoPOSTERSimplerStronger2023}}

& \mcrot{1}{l}{60}{\scriptsize EmoNeXt-Base$^{\dagger}$~\cite{boundori2023EmoNext}}
& \mcrot{1}{l}{60}{\scriptsize EmoNeXt-Large$^{\dagger}$~\cite{boundori2023EmoNext}}
& \mcrot{1}{l}{60}{\scriptsize EmoNeXt-Small$^{\dagger}$~\cite{boundori2023EmoNext}}
& \mcrot{1}{l}{60}{\scriptsize EmoNeXt-Tiny$^{\dagger}$~\cite{boundori2023EmoNext}}
& \mcrot{1}{l}{60}{\scriptsize ResidualMaskingNet~\cite{pham2021facial}}
& \mcrot{1}{l}{60}{\scriptsize Segmentation-VGG19$^{\dagger}$~\cite{vigneshNovelFacialEmotion2023}}

& \mcrot{1}{l}{60}{\scriptsize DAN~\cite{wenDistractYourAttention2023}} 
& \mcrot{1}{l}{60}{\scriptsize DDAMFN++~\cite{zhangDualDirectionAttentionMixed2023}}
& \mcrot{1}{l}{60}{\scriptsize PosterV2~\cite{maoPOSTERSimplerStronger2023}}\\
\toprule
&  \multicolumn{4}{c}{\hspace{-0.5cm}AffectNet7} & \multicolumn{4}{c}{\hspace{-0.5cm}AffectNet8} & \multicolumn{6}{c}{\hspace{-0.5cm}FER2013} & \multicolumn{3}{c}{\hspace{-0.1cm}RAFDB} \\ 
\midrule
Facial Palsy & \cmark & \cmark & \cmark & \cmark & \cmark & \cmark & \cmark & \cmark & \cmark & \cmark & \cmark & \cmark & \cmark & \cmark & \xmark & \cmark & \cmark \\
LPIPS symmetry \cite{zhangUnreasonableEffectivenessDeep2018} & \cmark & \cmark & \cmark & \cmark & \cmark & \cmark & \cmark & \cmark & \cmark & \cmark & \cmark & \cmark & \cmark & \cmark & \cmark & \cmark & \cmark \\
\bottomrule
\end{tabular}
    \caption{
        Significance results ($p < 0.01 \rightarrow${~\cmark}) of \cite{reimers2020determining} on our data for three symmetry features. 
        We analyze \happy logits regarding the binary facial palsy state and LPIPS~\cite{zhangUnreasonableEffectivenessDeep2018}  similarity between the face sides for facial palsy patients and healthy probands images.
    }
    \label{tab:ci}
\end{table}

Following related work~\cite{buechner2024power,reimers2020determining,penzel2023analyzing}, we denote all significant behavior changes in \Cref{tab:ci} using the majority decision of three conditional independence tests~\cite{fukumizu2007kernel,runge2018conditional,chalupka2018fast} (for detailed hyperparameters see supplementary material).
We find all 17 models show a statistically significant shift in their \happy activations for varying facial symmetries in the real-world data.
Please note that DAN~\cite{wenDistractYourAttention2023}, trained on RAFDB~\cite{li2017reliable,li2019reliable}, is the only classifier where we find no significance regarding binary facial palsy, which is a highly discretized form of symmetry.
However, the continuous symmetry measure LPIPS~\cite{zhangUnreasonableEffectivenessDeep2018} indicates the same behavior changes.
These results provide evidence for our hypothesis that expression classifiers are biased toward facial symmetry, especially concerning downstream applications.

We provide more qualitative investigations of the output changes in the supplementary material.
For most classifiers, we observe, on average, a decrease in activations.
Given this decrease, this observation is expected and indicates uncertainty in predicting the \happy class.
However, these are associational investigations~\cite{bareinboim2022onpearl}, i.e., we cannot isolate changes due to only facial symmetry.
Hence, while we observe changes in classifier behavior on real-world data, our interventional investigation in \Cref{sec:exp1} provides more reliable, actionable insights.

\subsection{Experiment 2: Synthetic Facial Symmetry Interventions}
\label{sec:exp1}

\begin{figure}[t]
    \centering
    \captionsetup[subfigure]{justification=centering}
    \begin{subfigure}[b]{0.32\textwidth}
        \centering
        \includegraphics[width=\textwidth, trim=0cm 0cm 0cm 0.6cm, clip]{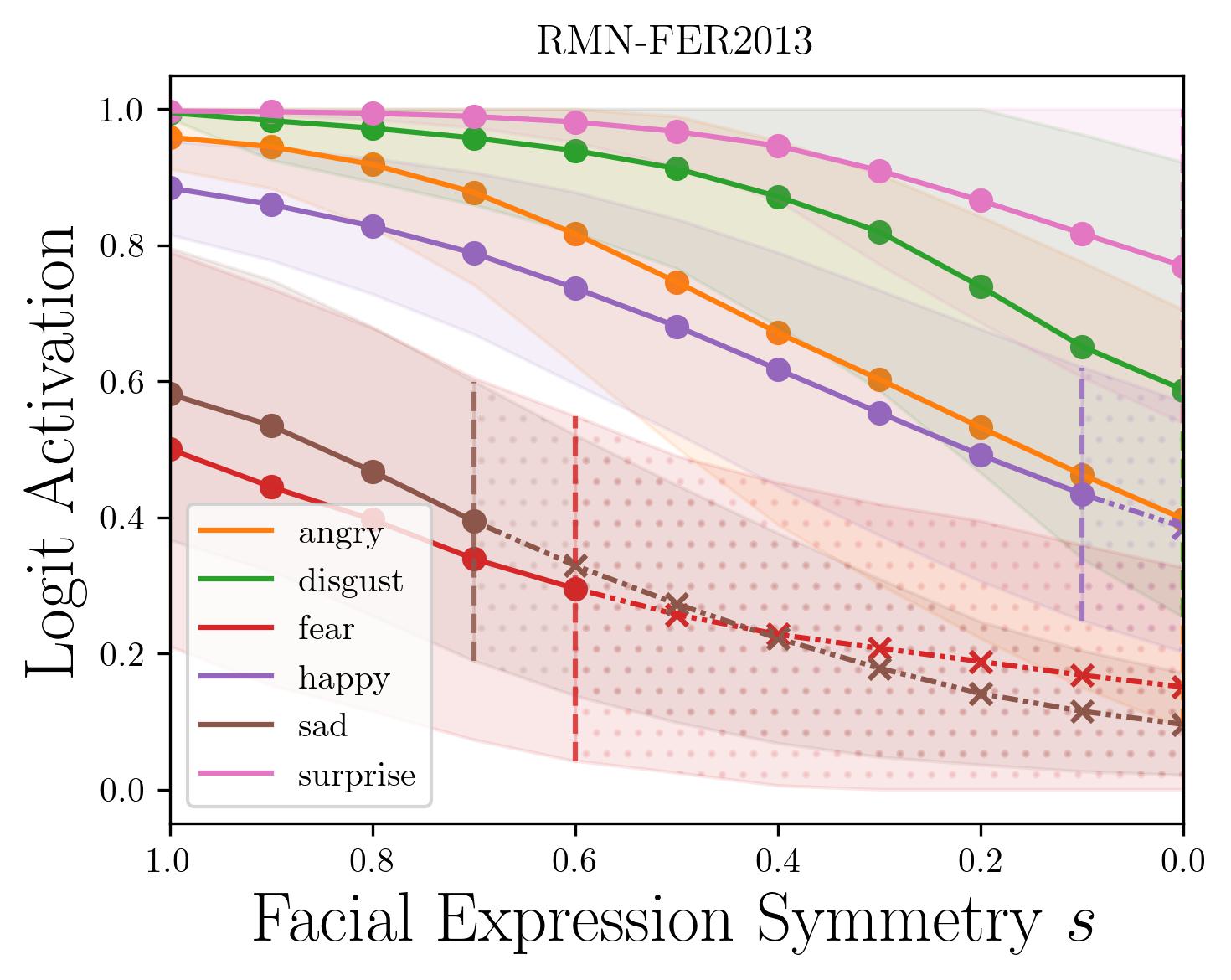}
        \caption{FER2013~\cite{dumitru2013fer}\\ResidualMaskingNet~\cite{pham2021facial}}
        \label{fig:rmn_symmetry_at_time}
    \end{subfigure}
    \hfill
    \begin{subfigure}[b]{0.32\textwidth}
        \centering
        \includegraphics[width=\textwidth, trim=0cm 0cm 0cm 0.6cm, clip]{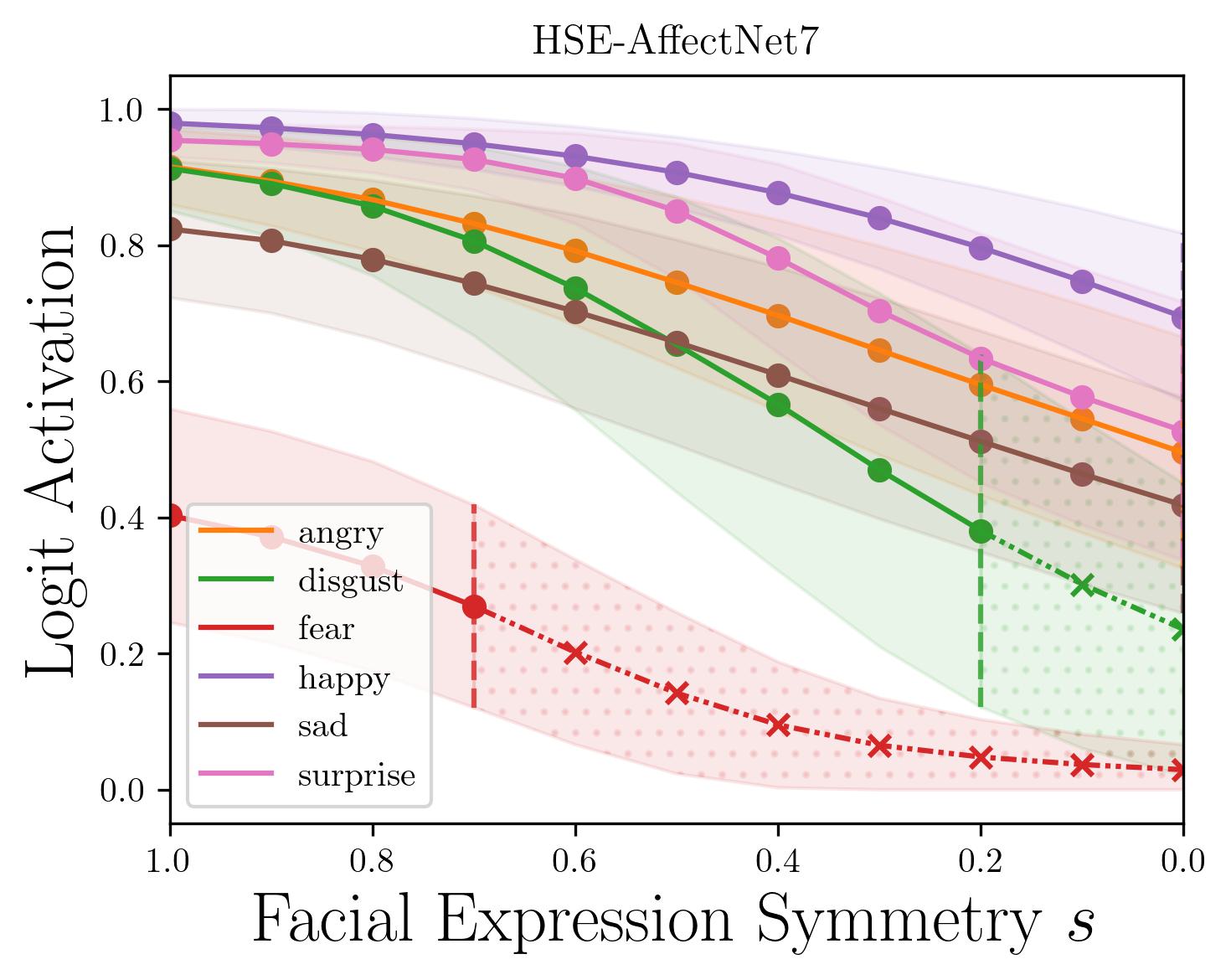}
        \caption{AffectNet~\cite{Mollahosseini2019affectnet}\\HSEmotion-7~\cite{savchenko2023facial,Savchenko_2022_CVPRW}}
        \label{fig:hse_symmetry_at_time}
    \end{subfigure}
    \hfill
    \begin{subfigure}[b]{0.32\textwidth}
        \centering
        \includegraphics[width=\textwidth, trim=0cm 0cm 0cm 0.6cm, clip]{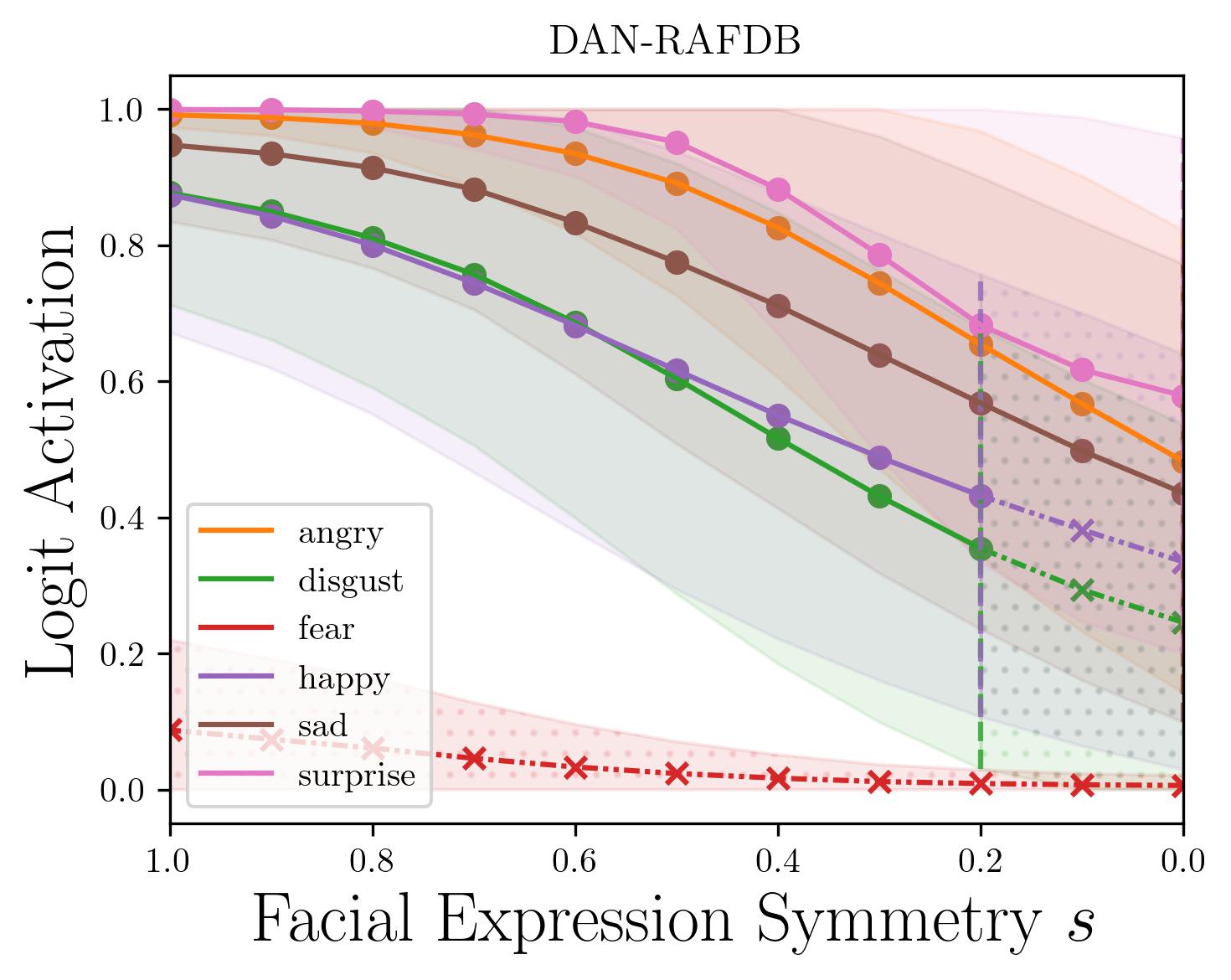}
        \caption{RAFDB~\cite{li2017reliable,li2019reliable}\\DAN~\cite{wenDistractYourAttention2023}}
        \label{fig:next_symmetry_at_time}
    \end{subfigure}
    \caption{
        We display a model's activation (mean and std.) curve at $t=1.0$ for each expression recognition dataset.
        Note that the x-axis is inverted, so we start with high symmetry.
        Lower symmetry generally results in lower logit activations across all expressions, with hatched lines indicating misclassification.
        We show the surface variants, such as \Cref{fig:fais_vis_a}, in the supplementary material.
    }
    \label{fig:models_t1_s}
\end{figure}

To confirm our hypothesis on how facial symmetry affects expression classification beyond the associational level, we perform synthetic interventions using the framework described in \Cref{sec:method}.
These enable us to measure the impact of facial asymmetry and model output.
Therefore, we create a population $\mathfrak{I}$ of $200$ identities sampled from a standard normal distribution (different $\avec$ and $\ivec$).
Following \Cref{eq:optim_facial_expressions}, we optimize the facial expression $\evec^{(\emo)}$ for each model and identity at $t=1.0$ and $s=1.0$.
To apply interventions, our finite grid spans ten equidistant symmetry steps ($s \in [0,1]$) and 90-time steps ($t \in [0,1]$), simulating three-second expression onset.
We then derive the mean logit surfaces (compare \Cref{fig:fais_vis}) of all classifiers over $\mathfrak{I}$ and display selected classifiers at $t=1.0$ in \Cref{fig:models_t1_s}.

We observe that facial symmetry impacts each expression's logit activation for the models irrespective of the dataset.
The ResidualMaskinNet \cite{pham2021facial} trained on FER2013~\cite{dumitru2013fer} does not reach high activations for \fear and {\color{sad}\emph{sad}}.
Further, they decrease even more for lower symmetry values.
Expressions such as {\color{angry}\emph{angry}}, {\color{disgust}\emph{disgust}}, or \surprise have higher activations and seem to be affected only by more pronounced asymmetry.
Especially \fear also seems problematic for the other classifiers highlighted in \Cref{fig:models_t1_s}.
Notably, the same behavior holds for HSEmotion~\cite{savchenko2023facial} and DAN~\cite{wenDistractYourAttention2023}:
Lower symmetry leads to lower output activations for all expressions, providing further evidence to confirm our hypothesis.

\begin{table}[t]
    \centering
    \caption{
        We compute $\hat{\mathcal{S}}(\eclass_{\ecweights}^{(\emo)})$, defined in \Cref{eq:globalShat}, among our population $\mathfrak{I}$.
        The models are grouped by the training dataset, and the $\dagger$ annotates models trained by us (see the supplementary material for details); otherwise, the provided models' weights were used respectively.
        All $\hat{\mathcal{S}}(\eclass_{\ecweights}^{(\emo)})$ are significant for the majority of individuals.
    }
    \begin{tabular}{llrrrrrr}
\toprule
 Dataset& $\eclass_\ecweights$ Model & \multicolumn{1}{c}{Angry} & \multicolumn{1}{c}{Disgust} & \multicolumn{1}{c}{Fear} & \multicolumn{1}{c}{Happy} & \multicolumn{1}{c}{Sad} & \multicolumn{1}{c}{Surprise} \\
\midrule

\multirow[c]{4}{*}{AffectNet7}
& DAN~\cite{wenDistractYourAttention2023} & {\cellcolor[HTML]{807DBA}} \color[HTML]{F1F1F1} 0.0249 & {\cellcolor[HTML]{8A86BF}} \color[HTML]{F1F1F1} 0.0234 & {\cellcolor[HTML]{DBDBEC}} \color[HTML]{000000} 0.0098 & {\cellcolor[HTML]{8582BC}} \color[HTML]{F1F1F1} 0.0242 & {\cellcolor[HTML]{A29FCB}} \color[HTML]{F1F1F1} 0.0192 & {\cellcolor[HTML]{7363AD}} \color[HTML]{F1F1F1} 0.0279 \\
& DDAMFN++~\cite{zhangDualDirectionAttentionMixed2023} & {\cellcolor[HTML]{EFEDF5}} \color[HTML]{000000} 0.0050 & {\cellcolor[HTML]{F3F1F7}} \color[HTML]{000000} 0.0035 & {\cellcolor[HTML]{F7F6FA}} \color[HTML]{000000} 0.0019 & {\cellcolor[HTML]{F2F0F7}} \color[HTML]{000000} 0.0040 & {\cellcolor[HTML]{F5F4F9}} \color[HTML]{000000} 0.0027 & {\cellcolor[HTML]{F3F1F7}} \color[HTML]{000000} 0.0037 \\
& HSEmotion~\cite{savchenko2023facial} & {\cellcolor[HTML]{8D89C0}} \color[HTML]{F1F1F1} 0.0229 & {\cellcolor[HTML]{66499F}} \color[HTML]{F1F1F1} 0.0308 & {\cellcolor[HTML]{D6D6E9}} \color[HTML]{000000} 0.0107 & {\cellcolor[HTML]{8683BD}} \color[HTML]{F1F1F1} 0.0239 & {\cellcolor[HTML]{9793C5}} \color[HTML]{F1F1F1} 0.0211 & {\cellcolor[HTML]{786DB2}} \color[HTML]{F1F1F1} 0.0268 \\
& PosterV2~\cite{maoPOSTERSimplerStronger2023} & {\cellcolor[HTML]{9591C4}} \color[HTML]{F1F1F1} 0.0214 & {\cellcolor[HTML]{A3A0CB}} \color[HTML]{F1F1F1} 0.0192 & {\cellcolor[HTML]{C6C7E1}} \color[HTML]{000000} 0.0133 & {\cellcolor[HTML]{8A86BF}} \color[HTML]{F1F1F1} 0.0234 & {\cellcolor[HTML]{B1B1D5}} \color[HTML]{000000} 0.0168 & {\cellcolor[HTML]{6E5AA8}} \color[HTML]{F1F1F1} 0.0290 \\

\midrule
\multirow[c]{4}{*}{AffectNet8} & DAN~\cite{wenDistractYourAttention2023} & {\cellcolor[HTML]{9995C6}} \color[HTML]{F1F1F1} 0.0209 & {\cellcolor[HTML]{908DC2}} \color[HTML]{F1F1F1} 0.0222 & {\cellcolor[HTML]{DEDDED}} \color[HTML]{000000} 0.0091 & {\cellcolor[HTML]{D3D3E8}} \color[HTML]{000000} 0.0112 & {\cellcolor[HTML]{9A96C6}} \color[HTML]{F1F1F1} 0.0207 & {\cellcolor[HTML]{7262AC}} \color[HTML]{F1F1F1} 0.0281 \\
& DDAMFN++~\cite{zhangDualDirectionAttentionMixed2023} & {\cellcolor[HTML]{F4F3F8}} \color[HTML]{000000} 0.0030 & {\cellcolor[HTML]{F4F2F8}} \color[HTML]{000000} 0.0032 & {\cellcolor[HTML]{F5F4F9}} \color[HTML]{000000} 0.0026 & {\cellcolor[HTML]{F5F4F9}} \color[HTML]{000000} 0.0028 & {\cellcolor[HTML]{F3F2F8}} \color[HTML]{000000} 0.0034 & {\cellcolor[HTML]{F3F1F7}} \color[HTML]{000000} 0.0036 \\
& HSEmotion~\cite{savchenko2023facial} & {\cellcolor[HTML]{C5C6E1}} \color[HTML]{000000} 0.0136 & {\cellcolor[HTML]{B9B9DA}} \color[HTML]{000000} 0.0156 & {\cellcolor[HTML]{E4E3F0}} \color[HTML]{000000} 0.0076 & {\cellcolor[HTML]{D3D3E8}} \color[HTML]{000000} 0.0111 & {\cellcolor[HTML]{BCBDDC}} \color[HTML]{000000} 0.0150 & {\cellcolor[HTML]{807DBA}} \color[HTML]{F1F1F1} 0.0250 \\
& PosterV2~\cite{maoPOSTERSimplerStronger2023} & {\cellcolor[HTML]{9B97C6}} \color[HTML]{F1F1F1} 0.0205 & {\cellcolor[HTML]{8E8AC0}} \color[HTML]{F1F1F1} 0.0228 & {\cellcolor[HTML]{C9C9E2}} \color[HTML]{000000} 0.0129 & {\cellcolor[HTML]{9692C4}} \color[HTML]{F1F1F1} 0.0214 & {\cellcolor[HTML]{AFAED4}} \color[HTML]{000000} 0.0171 & {\cellcolor[HTML]{786DB2}} \color[HTML]{F1F1F1} 0.0269 \\

\midrule
\multirow[c]{6}{*}{FER2013}
& EmoNeXt-Small$^{\dagger}$~\cite{boundori2023EmoNext} & {\cellcolor[HTML]{B8B8D9}} \color[HTML]{000000} 0.0157 & {\cellcolor[HTML]{F8F7FA}} \color[HTML]{000000} 0.0017 & {\cellcolor[HTML]{E3E2EF}} \color[HTML]{000000} 0.0078 & {\cellcolor[HTML]{A09DCA}} \color[HTML]{F1F1F1} 0.0195 & {\cellcolor[HTML]{E5E4F0}} \color[HTML]{000000} 0.0074 & {\cellcolor[HTML]{8784BD}} \color[HTML]{F1F1F1} 0.0238 \\
& EmoNeXt-Tiny$^{\dagger}$~\cite{boundori2023EmoNext} & {\cellcolor[HTML]{DEDDED}} \color[HTML]{000000} 0.0092 & {\cellcolor[HTML]{F8F7FA}} \color[HTML]{000000} 0.0017 & {\cellcolor[HTML]{EFEDF5}} \color[HTML]{000000} 0.0050 & {\cellcolor[HTML]{CCCCE4}} \color[HTML]{000000} 0.0124 & {\cellcolor[HTML]{EFEDF5}} \color[HTML]{000000} 0.0051 & {\cellcolor[HTML]{928FC3}} \color[HTML]{F1F1F1} 0.0219 \\
& EmoNeXt-Base$^{\dagger}$~\cite{boundori2023EmoNext}  & {\cellcolor[HTML]{DEDEED}} \color[HTML]{000000} 0.0089 & {\cellcolor[HTML]{FAF9FC}} \color[HTML]{000000} 0.0009 & {\cellcolor[HTML]{DCDCEC}} \color[HTML]{000000} 0.0096 & {\cellcolor[HTML]{A8A6CF}} \color[HTML]{F1F1F1} 0.0184 & {\cellcolor[HTML]{DCDCEC}} \color[HTML]{000000} 0.0096 & {\cellcolor[HTML]{8E8AC0}} \color[HTML]{F1F1F1} 0.0227 \\
& EmoNeXt-Large$^{\dagger}$~\cite{boundori2023EmoNext}  & {\cellcolor[HTML]{BDBEDC}} \color[HTML]{000000} 0.0149 & {\cellcolor[HTML]{E9E8F2}} \color[HTML]{000000} 0.0065 & {\cellcolor[HTML]{B4B4D7}} \color[HTML]{000000} 0.0163 & {\cellcolor[HTML]{9A96C6}} \color[HTML]{F1F1F1} 0.0207 & {\cellcolor[HTML]{8885BE}} \color[HTML]{F1F1F1} 0.0236 & {\cellcolor[HTML]{8D89C0}} \color[HTML]{F1F1F1} 0.0228 \\
& ResidualMaskingNet~\cite{pham2021facial} & {\cellcolor[HTML]{6B53A4}} \color[HTML]{F1F1F1} 0.0298 & {\cellcolor[HTML]{674BA0}} \color[HTML]{F1F1F1} 0.0307 & {\cellcolor[HTML]{DADAEB}} \color[HTML]{000000} 0.0099 & {\cellcolor[HTML]{807CBA}} \color[HTML]{F1F1F1} 0.0251 & {\cellcolor[HTML]{C4C5E0}} \color[HTML]{000000} 0.0137 & {\cellcolor[HTML]{7262AC}} \color[HTML]{F1F1F1} 0.0280 \\
& Segm-VGG19$^{\dagger}$~\cite{vigneshNovelFacialEmotion2023} & {\cellcolor[HTML]{A7A4CE}} \color[HTML]{F1F1F1} 0.0186 & {\cellcolor[HTML]{FAF8FB}} \color[HTML]{000000} 0.0010 & {\cellcolor[HTML]{AEACD2}} \color[HTML]{000000} 0.0174 & {\cellcolor[HTML]{8784BD}} \color[HTML]{F1F1F1} 0.0238 & {\cellcolor[HTML]{918EC2}} \color[HTML]{F1F1F1} 0.0221 & {\cellcolor[HTML]{9B97C6}} \color[HTML]{F1F1F1} 0.0206 \\

\midrule
\multirow[c]{3}{*}{RAFDB}
& DAN~\cite{wenDistractYourAttention2023} & {\cellcolor[HTML]{61409B}} \color[HTML]{F1F1F1} 0.0319 & {\cellcolor[HTML]{7B72B4}} \color[HTML]{F1F1F1} 0.0262 & {\cellcolor[HTML]{F8F7FA}} \color[HTML]{000000} 0.0017 & {\cellcolor[HTML]{9B97C6}} \color[HTML]{F1F1F1} 0.0205 & {\cellcolor[HTML]{827FBB}} \color[HTML]{F1F1F1} 0.0246 & {\cellcolor[HTML]{63449D}} \color[HTML]{F1F1F1} 0.0314 \\
 & DDAMFN++~\cite{zhangDualDirectionAttentionMixed2023} & {\cellcolor[HTML]{F9F7FB}} \color[HTML]{000000} 0.0013 & {\cellcolor[HTML]{F9F7FB}} \color[HTML]{000000} 0.0013 & {\cellcolor[HTML]{FCFBFD}} \color[HTML]{000000} 0.0002 & {\cellcolor[HTML]{C6C7E1}} \color[HTML]{000000} 0.0134 & {\cellcolor[HTML]{D0D1E6}} \color[HTML]{000000} 0.0117 & {\cellcolor[HTML]{AAA8D0}} \color[HTML]{000000} 0.0181 \\
 & PosterV2~\cite{maoPOSTERSimplerStronger2023} & {\cellcolor[HTML]{5E3B98}} \color[HTML]{F1F1F1} 0.0326 & {\cellcolor[HTML]{8D89C0}} \color[HTML]{F1F1F1} 0.0228 & {\cellcolor[HTML]{EEECF4}} \color[HTML]{000000} 0.0054 & {\cellcolor[HTML]{64459E}} \color[HTML]{F1F1F1} 0.0313 & {\cellcolor[HTML]{B2B2D5}} \color[HTML]{000000} 0.0166 & {\cellcolor[HTML]{4A1587}} \color[HTML]{F1F1F1} 0.0373 \\
\bottomrule
\end{tabular}

    \label{tab:symmetry_scores}
\end{table}

We go one step further and quantitatively measure the impact of facial symmetry using our proposed score (\Cref{sec:score}).
\Cref{tab:symmetry_scores} summarizes these results averaged over all individuals $\mathfrak{I}$.
A positive score corresponds to increased activations for increased symmetry.
Further, we report the ratio of significant systematic changes over the set of individuals $\mathfrak{I}$ in Table~1 of the supplementary material.
In most cases, we observe a significant impact of facial symmetry for all classifiers and expressions.
This enables us to interpret the patterns we observe in \Cref{tab:symmetry_scores} concerning the expressions and training datasets.
We note the highest impact of $0.0373$ for \surprise of PosterV2~\cite{maoPOSTERSimplerStronger2023}.
This score indicates that, on average, over the complete \surprise onset, increasing the symmetry by one step in our simulation increased the softmax output of PosterV2 by $3.7$ percentage points.
However, while all 17 classifiers are significantly impacted by changes in facial symmetry, the effect size can still be small (see, for example, \fear in \Cref{fig:next_symmetry_at_time}).

We start with broad insights about the results in \Cref{tab:symmetry_scores}, before focusing on specific models:
First, all $\hat{\mathcal{S}}(\eclass_{\ecweights}^{(\emo)})$ contained in \Cref{tab:symmetry_scores} are positive.
Hence, expression classifiers show, on average, lower logit activations for decreased symmetry.
This result provides \emph{interventional evidence} for our previously stated hypothesis.
Second, similar to \Cref{fig:models_t1_s}, we see low scores for \fear expressions irrespective of the classifier and dataset.
However, this is likely due to the often lower activations for \fear (\Cref{fig:models_t1_s} and supplementary material).
The FLAME expression space may limit accurately modeling {\color{fear}\emph{fear}}.
The shift in model outputs is, nevertheless, significant.
In contrast to {\color{fear}\emph{fear}}, the overall high scores for {\color{happy}\emph{happy}}, {\color{surprise}\emph{surprise}}, and \angry suggest stronger changes in model behavior for these expressions.

Seen in \Cref{tab:symmetry_scores}, models trained on the same dataset often show similar $\hat{\mathcal{S}}(\eclass_{\ecweights}^{(\emo)})$, likely due to the latent training data distribution~\cite{lesson}.
For FER2013~\cite{dumitru2013fer} classifiers, facial symmetry has a lower impact on the \disgust expression, excluding ResidualMaskingNet~\cite{pham2021facial}.
Similarly, models trained on this dataset display lower scores for \sad and {\color{angry}\emph{angry}}.
We analyze different EmoNeXt~\cite{boundori2023EmoNext} model sizes. 
\emph{Large} being impacted the most, indicating higher capacity could consolidate biases in the training data.
In contrast, DDAMFN++\cite{chenStaticDynamicAdapting2023} shows a small effect size irrespective of the dataset.
Visualizing the graphs at $t=1.0$ for different $s$ and the classification accuracies on the real-world data (both in the supplementary material), we assume that the model is likely overfitting the training data.
Thus, low output activations result in smaller $\nabla_s$.

Nevertheless, we conclude that \underline{all} analyzed classifiers show significant increases in output activations for higher facial symmetry, confirming our hypothesis.
Given that we use interventions beyond the associational level, we verify a causal link between facial symmetry and expression classifiers' behavior. 

\section{Limitations and Social Impact}
Our work relies on the statistical shape model, inducing a domain shift and limiting possible expressions.
Some classifiers advertise the out-of-domain usage, e.g.,~\cite{savchenko2022classifying, Savchenko_2022_CVPRW, savchenko2023facial, pham2021facial}, and we optimize the synthetic faces regarding model and expression (see \Cref{sec:eval_frame_work}).
Factors like camera angle could jointly influence the behavior. 
Secondly, other facial features, e.g., age or skin color, could impact classifier performance and should be considered.
In our current framework, we cannot account for all possible forms of facial asymmetry, e.g., synkinetic effects.

Regarding societal impact, we investigate existing expression classifiers only.
We move from the associational level to causal interventions to better understand how these black-box models operate.
This could benefit other disciplines, primarily psychological and medical applications.
We provide our experiments' framework and evaluation code so that researchers can evaluate their models.

\section{Conclusion}

Emotional expressiveness is crucial for communicating our internal state and for understanding other people~\cite{thompson1987empathy,roberts1996empathy}.
In this work, we investigate the impact of facial symmetry on 17 different expression classifiers trained on four different datasets~\cite{dumitru2013fer, li2017reliable, li2019reliable, Mollahosseini2019affectnet}.
Extending empirical analysis, we try to answer an interventional question \cite{bareinboim2022onpearl} by following insights from causal inference and explainability \cite{reimers2020determining} and using an SCM (\Cref{fig:scm}) together with a generative framework.
We control expression and facial symmetry using a modified statistical shape model~\cite{liLearningModelFacial2017} to measure systematic changes with a proposed interpretable score.

We tested our hypothesis on real-world data using associational methods~\cite{reimers2020determining,penzel2023analyzing,buechner2024power}.
Here, we saw that facial palsy and the similarity between the face halves led to 33 out of 34 tests being significant.
To verify these results in a controlled manner, moving up the causal hierarchy, we employed our interventional framework to test the impact of symmetry on the models' behavior.
We observed that many classifiers, on average, decrease logit activations for lower facial symmetry.

While, in retrospect, our results align with the pre-specified intuition, we stress that our framework provides a structured way to test such hypotheses.
Further, it could be extended to other features, e.g., age or skin color, given the controllable nature of statistical shape models.
These insights could also be used to grade facial palsy or to correct the classier output for patients posthoc. 
Hence, we hope that our work can help researchers understand the prediction behavior of their trained expression classifiers beyond simple performance metrics.

\subsubsection{Ethics Approval}
Written consent was obtained from all participants.
The Jena University Hospital ethics committee approved the study (No. 2019-1539).

\subsubsection{Acknowledgement}
Partially supported by Deutsche Forschungsgemeinschaft (DFG - German Research Foundation) project 427899908 BRIDGING THE GAP: MIMICS AND MUSCLES (DE 735/15-1 and GU 463/12-1).
\bibliographystyle{splncs04}
\bibliography{references}

\begin{thebibliography}{100}
\providecommand{\url}[1]{\texttt{#1}}
\providecommand{\urlprefix}{URL }
\providecommand{\doi}[1]{https://doi.org/#1}

\bibitem{baltrusaitisOpenFaceOpenSource2016}
Baltru{\v s}aitis, T., Robinson, P., Morency, L.: {{OpenFace}}: {{An}} open source facial behavior analysis toolkit. In: 2016 {{IEEE Winter Conference}} on {{Applications}} of {{Computer Vision}} ({{WACV}}). pp. 1--10 (Mar 2016). \doi{10.1109/WACV.2016.7477553}

\bibitem{baltrusaitisOpenFaceOpenSource2018}
Baltrusaitis, T., Zadeh, A., Lim, Y.C., Morency, L.P.: {{OpenFace}} 2.0: {{Facial}} behavior analysis toolkit. In: 2018 13th {{IEEE}} International Conference on Automatic Face \& Gesture Recognition ({{FG}} 2018). pp. 59--66 (2018). \doi{10.1109/FG.2018.00019}

\bibitem{banksClinicianGradedElectronicFacial2015}
Banks, C.A., Bhama, P.K., Park, J., Hadlock, C.R., Hadlock, T.A.: Clinician-{{Graded Electronic Facial Paralysis Assessment}}: {{The eFACE}}. Plastic and Reconstructive Surgery  \textbf{136}(2), ~223e (Aug 2015). \doi{10.1097/PRS.0000000000001447}

\bibitem{bareinboim2022onpearl}
Bareinboim, E., Correa, J.D., Ibeling, D., Icard, T.F.: On pearl’s hierarchy and the foundations of causal inference. Probabilistic and Causal Inference  (2022)

\bibitem{blanzMorphableModelSynthesis1999}
Blanz, V., Vetter, T.: A morphable model for the synthesis of {{3D}} faces. In: Proceedings of the 26th Annual Conference on {{Computer}} Graphics and Interactive Techniques - {{SIGGRAPH}} '99. pp. 187--194. ACM Press, Not Known (1999). \doi{10.1145/311535.311556}

\bibitem{buechner2023improved}
B{\"u}chner, T., Guntinas-Lichius, O., Denzler, J.: Improved obstructed facial feature reconstruction for emotion recognition with minimal change cyclegans. In: Advanced Concepts for Intelligent Vision Systems (Acivs). pp. 262--274. SpringerNature (august 2023). \doi{10.1007/978-3-031-45382-3_22}

\bibitem{buechner2023lets}
B{\"u}chner, T., Sickert, S., Volk, G.F., Anders, C., Guntinas-Lichius, O., Denzler, J.: Let’s get the facs straight - reconstructing obstructed facial features. In: International Conference on Computer Vision Theory and Applications (VISAPP). SciTePress (march 2023). \doi{10.5220/0011619900003417}

\bibitem{buchnerFacesVolumesMeasuring2023}
B{\"u}chner, T., Sickert, S., Volk, G.F., {Guntinas-Lichius}, O., Denzler, J.: From {{Faces}} to~{{Volumes}} - {{Measuring Volumetric Asymmetry}} in~{{3D Facial Palsy Scans}}. In: Advances in {{Visual Computing}}. Lecture {{Notes}} in {{Computer Science}}, {Springer Nature Switzerland} (2023). \doi{10.1007/978-3-031-47969-4_10}

\bibitem{buechner2024power}
Büchner, T., Penzel, N., Guntinas-Lichius, O., Denzler, J.: The power of properties: Uncovering the influential factors in emotion classification. In: International Conference on Pattern Recognition and Artificial Intelligence (ICPRAI) (2024), \url{https://arxiv.org/abs/2404.07867}, (accepted)

\bibitem{Buechner20232D3DFace}
Büchner, T., Sickert, S., Graßme, R., Anders, C., Guntinas-Lichius, O., Denzler, J.: Using 2d and 3d face representations to generate comprehensive facial electromyography intensity maps. In: International Symposium on Visual Computing (ISVC). pp. 136--147 (2023). \doi{10.1007/978-3-031-47966-3_11}, \url{https://link.springer.com/chapter/10.1007/978-3-031-47966-3_11}

\bibitem{chalupka2018fast}
Chalupka, K., Perona, P., Eberhardt, F.: Fast conditional independence test for vector variables with large sample sizes. arXiv preprint arXiv:1804.02747  (2018)

\bibitem{chenStaticDynamicAdapting2023}
Chen, Y., Li, J., Shan, S., Wang, M., Hong, R.: From {{Static}} to {{Dynamic}}: {{Adapting Landmark-Aware Image Models}} for {{Facial Expression Recognition}} in {{Videos}} (Dec 2023)

\bibitem{chen2019learning}
Chen, Y., Li, W., Chen, X., Gool, L.V.: Learning semantic segmentation from synthetic data: A geometrically guided input-output adaptation approach. In: Proceedings of the IEEE/CVF conference on computer vision and pattern recognition. pp. 1841--1850 (2019)

\bibitem{choi2018stargan}
Choi, Y., Choi, M., Kim, M., Ha, J.W., Kim, S., Choo, J.: Stargan: {{Unified}} generative adversarial networks for multi-domain image-to-image translation. In: Proceedings of the {{IEEE}} Conference on Computer Vision and Pattern Recognition. pp. 8789--8797 (2018)

\bibitem{choithwaniPoseBiasDatasetBias2023}
Choithwani, M., Almeida, S., Egger, B.: {{PoseBias}}: {{On Dataset Bias}} and {{Task Difficulty}} - {{Is}} there an {{Optimal Camera Position}} for {{Facial Image Analysis}}? In: 2023 {{IEEE}}/{{CVF International Conference}} on {{Computer Vision Workshops}} ({{ICCVW}}). pp. 3088--3096. IEEE, Paris, France (Oct 2023). \doi{10.1109/ICCVW60793.2023.00334}

\bibitem{cootes2001active}
Cootes, T.F., Edwards, G.J., Taylor, C.J.: Active appearance models. IEEE Transactions on pattern analysis and machine intelligence  \textbf{23}(6),  681--685 (2001)

\bibitem{daneVcVekEMOCAEmotionDriven2022}
Dane{\textasciicaron}c{\textasciicaron}ek, R., Black, M.J., Bolkart, T.: {{EMOCA}}: {{Emotion Driven Monocular Face Capture}} and {{Animation}}. CVPR p.~12 (2022)

\bibitem{demecoQuantitativeAnalysisMovements2021}
Demeco, A., Marotta, N., Moggio, L., Pino, I., Marinaro, C., Barletta, M., Petraroli, A., Palumbo, A., Ammendolia, A.: Quantitative analysis of movements in facial nerve palsy with surface electromyography and kinematic analysis. Journal of Electromyography and Kinesiology  \textbf{56},  102485 (Feb 2021). \doi{10.1016/j.jelekin.2020.102485}

\bibitem{deng2020disentangled}
Deng, Y., Yang, J., Chen, D., Wen, F., Tong, X.: Disentangled and controllable face image generation via 3d imitative-contrastive learning. In: Proceedings of the {{IEEE}}/{{CVF}} Conference on Computer Vision and Pattern Recognition. pp. 5154--5163 (2020)

\bibitem{dumitru2013fer}
Dumitru, Goodfellow, I., Cukierski, W., Bengio, Y.: Challenges in representation learning: Facial expression recognition challenge (2013), \url{https://kaggle.com/competitions/challenges-in-representation-learning-facial-expression-recognition-challenge}

\bibitem{egger3DMorphableFace2020}
Egger, B., Smith, W.A.P., Tewari, A., Wuhrer, S., Zollhoefer, M., Beeler, T., Bernard, F., Bolkart, T., Kortylewski, A., Romdhani, S., Theobalt, C., Blanz, V., Vetter, T.: {{3D Morphable Face Models-Past}}, {{Present}}, and {{Future}}. ACM Transactions on Graphics  \textbf{39}(5),  157:1--157:38 (Jun 2020). \doi{10.1145/3395208}

\bibitem{eggerIdentityExpressionAmbiguity3D2021a}
Egger, B., Sutherland, S., Medin, S.C., Tenenbaum, J.: Identity-{{Expression Ambiguity}} in {{3D Morphable Face Models}}. In: 2021 16th {{IEEE International Conference}} on {{Automatic Face}} and {{Gesture Recognition}} ({{FG}} 2021). pp.~1--7. IEEE Press, Jodhpur, India (Dec 2021). \doi{10.1109/FG52635.2021.9667002}

\bibitem{ekman1992argument}
Ekman, P.: An argument for basic emotions. Cognition and Emotion  \textbf{6}(3-4),  169--200 (1992). \doi{10.1080/02699939208411068}

\bibitem{boundori2023EmoNext}
El~Boudouri, Y., Bohi, A.: Emonext: an adapted convnext for facial emotion recognition. In: 2023 IEEE 25th International Workshop on Multimedia Signal Processing (MMSP). pp.~1--6 (2023). \doi{10.1109/MMSP59012.2023.10337732}

\bibitem{fengLearningAnimatableDetailed2021c}
Feng, Y., Feng, H., Black, M.J., Bolkart, T.: Learning an animatable detailed {{3D}} face model from in-the-wild images. ACM Transactions on Graphics  \textbf{40}(4),  1--13 (Aug 2021). \doi{10.1145/3450626.3459936}

\bibitem{fornberg1988generation}
Fornberg, B.: Generation of finite difference formulas on arbitrarily spaced grids. Mathematics of Computation  \textbf{51},  699--706 (1988), \url{https://api.semanticscholar.org/CorpusID:119513587}

\bibitem{fukumizu2007kernel}
Fukumizu, K., Gretton, A., Sun, X., Sch{\"o}lkopf, B.: Kernel measures of conditional dependence. Advances in neural information processing systems  \textbf{20} (2007)

\bibitem{gao2010review}
Gao, X., Su, Y., Li, X., Tao, D.: A review of active appearance models. IEEE Transactions on Systems, Man, and Cybernetics, Part C (Applications and Reviews)  \textbf{40}(2),  145--158 (2010)

\bibitem{gerigMorphableFaceModels2018}
Gerig, T., {Morel-Forster}, A., Blumer, C., Egger, B., Luthi, M., Schoenborn, S., Vetter, T.: Morphable {{Face Models}} - {{An Open Framework}}. In: 2018 13th {{IEEE International Conference}} on {{Automatic Face}} \& {{Gesture Recognition}} ({{FG}} 2018). pp. 75--82. IEEE, Xi'an (May 2018). \doi{10.1109/FG.2018.00021}

\bibitem{good2013permutation}
Good, P.: Permutation Tests: A Practical Guide to Resampling Methods for Testing Hypotheses. Springer Series in Statistics, Springer New York (2013), \url{https://books.google.de/books?id=pK3hBwAAQBAJ}

\bibitem{gretton2006kernel}
Gretton, A., Borgwardt, K., Rasch, M., Sch{\"o}lkopf, B., Smola, A.: A kernel method for the two-sample-problem. Adv. Neural. Inf. Process. Syst.  \textbf{19} (2006)

\bibitem{guntinas2023high}
Guntinas-Lichius, O., Trentzsch, V., Mueller, N., Heinrich, M., Kuttenreich, A.M., Dobel, C., et~al.: High-resolution surface electromyographic activities of facial muscles during the six basic emotional expressions in healthy adults: a prospective observational study. Scientific Reports  \textbf{13}(1),  19214 (2023)

\bibitem{guo2020towards}
Guo, J., Zhu, X., Yang, Y., Yang, F., Lei, Z., Li, S.Z.: Towards fast, accurate and stable {{3D}} dense face alignment. In: Proceedings of the European Conference on Computer Vision ({{ECCV}}) (2020)

\bibitem{haase2014instance}
Haase, D., Rodner, E., Denzler, J.: Instance-weighted transfer learning of active appearance models. In: Proceedings of the {{IEEE}} Conference on Computer Vision and Pattern Recognition. pp. 1426--1433 (2014)

\bibitem{harris2020array}
Harris, C.R., Millman, K.J., van~der Walt, S.J., Gommers, R., Virtanen, P., Cournapeau, D., Wieser, E., Taylor, J., Berg, S., Smith, N.J., Kern, R., Picus, M., Hoyer, S., van Kerkwijk, M.H., Brett, M., Haldane, A., del R{\'{i}}o, J.F., Wiebe, M., Peterson, P., G{\'{e}}rard-Marchant, P., Sheppard, K., Reddy, T., Weckesser, W., Abbasi, H., Gohlke, C., Oliphant, T.E.: Array programming with {NumPy}. Nature  \textbf{585}(7825),  357--362 (Sep 2020). \doi{10.1038/s41586-020-2649-2}, \url{https://doi.org/10.1038/s41586-020-2649-2}

\bibitem{holm1979simple}
Holm, S.: A simple sequentially rejective multiple test procedure. Scandinavian Journal of Statistics  \textbf{6}(2),  65--70 (1979), \url{http://www.jstor.org/stable/4615733}

\bibitem{hu2021sail}
Hu, Y.T., Wang, J., Yeh, R.A., Schwing, A.G.: Sail-vos 3d: A synthetic dataset and baselines for object detection and 3d mesh reconstruction from video data. In: Proceedings of the IEEE/CVF Conference on Computer Vision and Pattern Recognition. pp. 1418--1428 (2021)

\bibitem{josifovski2018object}
Josifovski, J., Kerzel, M., Pregizer, C., Posniak, L., Wermter, S.: Object detection and pose estimation based on convolutional neural networks trained with synthetic data. In: 2018 IEEE/RSJ international conference on intelligent robots and systems (IROS). pp. 6269--6276. IEEE (2018)

\bibitem{karrasStyleBasedGeneratorArchitecture2019}
Karras, T., Laine, S., Aila, T.: A {{Style-Based Generator Architecture}} for {{Generative Adversarial Networks}} (Mar 2019). \doi{10.48550/arXiv.1812.04948}

\bibitem{katsumiQuantitativeAnalysisFacial2015}
Katsumi, S., Esaki, S., Hattori, K., Yamano, K., Umezaki, T., Murakami, S.: Quantitative analysis of facial palsy using a three-dimensional facial motion measurement system. Auris Nasus Larynx  \textbf{42}(4),  275--283 (Aug 2015). \doi{10.1016/j.anl.2015.01.002}

\bibitem{kim2018interpretability}
Kim, B., Wattenberg, M., Gilmer, J., Cai, C., Wexler, J., Viegas, F., et~al.: Interpretability beyond feature attribution: Quantitative testing with concept activation vectors (tcav). In: International conference on machine learning. pp. 2668--2677. PMLR (2018)

\bibitem{knoedlerReliableRapidAutomated2022}
Knoedler, L., Baecher, H., {Kauke-Navarro}, M., Prantl, L., Machens, H.G., Scheuermann, P., Palm, C., Baumann, R., Kehrer, A., Panayi, A.C., Knoedler, S.: Towards a {{Reliable}} and {{Rapid Automated Grading System}} in {{Facial Palsy Patients}}: {{Facial Palsy Surgery Meets Computer Science}}. Journal of Clinical Medicine  \textbf{11}(17), ~4998 (Aug 2022). \doi{10.3390/jcm11174998}

\bibitem{kortylewskiEmpiricallyAnalyzingEffect2018}
Kortylewski, A., Egger, B., Schneider, A., Gerig, T., {Morel-Forster}, A., Vetter, T.: Empirically {{Analyzing}} the {{Effect}} of {{Dataset Biases}} on {{Deep Face Recognition Systems}}. In: 2018 {{IEEE}}/{{CVF Conference}} on {{Computer Vision}} and {{Pattern Recognition Workshops}} ({{CVPRW}}). pp. 2174--217409. IEEE, Salt Lake City, UT, USA (Jun 2018). \doi{10.1109/CVPRW.2018.00283}

\bibitem{lapuschkin2019unmasking}
Lapuschkin, S., W{\"a}ldchen, S., Binder, A., Montavon, G., Samek, W., M{\"u}ller, K.R.: Unmasking clever hans predictors and assessing what machines really learn. Nature communications  \textbf{10}(1), ~1096 (2019)

\bibitem{lewis2000posespace}
Lewis, J.P., Cordner, M., Fong, N.: Pose space deformation: A unified approach to shape interpolation and skeleton-driven deformation. In: Proceedings of the 27th Annual Conference on Computer Graphics and Interactive Techniques. pp. 165--172. Siggraph '00, ACM Press/Addison-Wesley Publishing Co., USA (2000). \doi{10.1145/344779.344862}

\bibitem{li2019reliable}
Li, S., Deng, W.: Reliable crowdsourcing and deep locality-preserving learning for unconstrained facial expression recognition. IEEE Transactions on Image Processing  \textbf{28}(1),  356--370 (2019)

\bibitem{li2017reliable}
Li, S., Deng, W., Du, J.: Reliable crowdsourcing and deep locality-preserving learning for expression recognition in the wild. In: 2017 {{IEEE}} Conference on Computer Vision and Pattern Recognition ({{CVPR}}). pp. 2584--2593. IEEE (2017)

\bibitem{liLearningModelFacial2017}
Li, T., Bolkart, T., Black, M.J., Li, H., Romero, J.: Learning a model of facial shape and expression from {{4D}} scans. ACM Transactions on Graphics  \textbf{36}(6),  1--17 (Nov 2017). \doi{10.1145/3130800.3130813}

\bibitem{linSingleShotImplicitMorphable2023}
Lin, C.Z., Nagano, K., Kautz, J., Chan, E.R., Iqbal, U., Guibas, L., Wetzstein, G., Khamis, S.: Single-{{Shot Implicit Morphable Faces}} with {{Consistent Texture Parameterization}}. In: Special {{Interest Group}} on {{Computer Graphics}} and {{Interactive Techniques Conference Conference Proceedings}}. pp. 1--12 (Jul 2023). \doi{10.1145/3588432.3591494}

\bibitem{maoPOSTERSimplerStronger2023}
Mao, J., Xu, R., Yin, X., Chang, Y., Nie, B., Huang, A.: {{POSTER}}++: {{A}} simpler and stronger facial expression recognition network (Feb 2023)

\bibitem{matthews2004active}
Matthews, I., Baker, S.: Active appearance models revisited. International journal of computer vision  \textbf{60},  135--164 (2004)

\bibitem{medinMOSTGAN3DMorphable2022}
Medin, S.C., Egger, B., Cherian, A., Wang, Y., Tenenbaum, J.B., Liu, X., Marks, T.K.: {{MOST-GAN}}: {{3D Morphable StyleGAN}} for {{Disentangled Face Image Manipulation}}. Proceedings of the AAAI Conference on Artificial Intelligence  \textbf{36}(2),  1962--1971 (Jun 2022). \doi{10.1609/aaai.v36i2.20091}

\bibitem{Mollahosseini2019affectnet}
Mollahosseini, A., Hasani, B., Mahoor, M.H.: Affectnet: A database for facial expression, valence, and arousal computing in the wild. IEEE Transactions on Affective Computing  \textbf{10}(1),  18--31 (2019). \doi{10.1109/TAFFC.2017.2740923}

\bibitem{Nachbar1994TheAR}
Nachbar, F., Stolz, W., Merkle, T., Cognetta, A.B., Vogt, T., Landthaler, M., Bilek, P., Braun-Falco, O., Plewig, G.: The abcd rule of dermatoscopy. high prospective value in the diagnosis of doubtful melanocytic skin lesions. Journal of the American Academy of Dermatology  \textbf{30 4},  551--9 (1994), \url{https://api.semanticscholar.org/CorpusID:4860343}

\bibitem{neumannValidierungDeutschenVersion2016}
Neumann, T., Lorenz, A., Volk, G., Hamzei, F., Schulz, S., {Guntinas-Lichius}, O.: {Validierung einer Deutschen Version des Sunnybrook Facial Grading Systems}. Laryngo-Rhino-Otologie  \textbf{96}(03),  168--174 (Nov 2016). \doi{10.1055/s-0042-111512}

\bibitem{nowruzi2019much}
Nowruzi, F.E., Kapoor, P., Kolhatkar, D., Hassanat, F.A., Laganiere, R., Rebut, J.: How much real data do we actually need: Analyzing object detection performance using synthetic and real data. arXiv preprint arXiv:1907.07061  (2019)

\bibitem{ozsoyThreedimensionalObjectiveEvaluation2021}
{\"O}zsoy, U., Uysal, H., Hizay, A., Sekerci, R., Yildirim, Y.: Three-dimensional objective evaluation of facial palsy and follow-up of recovery with a handheld scanner. Journal of Plastic, Reconstructive \& Aesthetic Surgery p. S1748681521002552 (Jun 2021). \doi{10.1016/j.bjps.2021.05.003}

\bibitem{patelFacialAsymmetryAssessment2015}
Patel, A., Islam, S.M.S., Murray, K., Goonewardene, M.S.: Facial asymmetry assessment in adults using three-dimensional surface imaging. Progress in Orthodontics  \textbf{16}(1), ~36 (Oct 2015). \doi{10.1186/s40510-015-0106-9}

\bibitem{paysan3DFaceModel2009}
Paysan, P., Knothe, R., Amberg, B., Romdhani, S., Vetter, T.: A {{3D Face Model}} for {{Pose}} and {{Illumination Invariant Face Recognition}}. In: 2009 {{Sixth IEEE International Conference}} on {{Advanced Video}} and {{Signal Based Surveillance}}. pp. 296--301. IEEE, Genova, Italy (Sep 2009). \doi{10.1109/AVSS.2009.58}

\bibitem{pearl2009causality}
Pearl, J.: Causality. Cambridge university press (2009)

\bibitem{penzel2023analyzing}
Penzel, N., Kierdorf, J., Roscher, R., Denzler, J.: Analyzing the behavior of cauliflower harvest-readiness models by investigating feature relevances. In: 2023 IEEE/CVF International Conference on Computer Vision Workshops (ICCVW). pp. 572--581. IEEE (2023)

\bibitem{penzel2022investigating}
Penzel, N., Reimers, C., Bodesheim, P., Denzler, J.: Investigating neural network training on a feature level using conditional independence. In: European Conference on Computer Vision. pp. 383--399. Springer (2022)

\bibitem{perarnau2016invertible}
Perarnau, G., Van De~Weijer, J., Raducanu, B., {\'A}lvarez, J.M.: Invertible conditional gans for image editing. arXiv preprint arXiv:1611.06355  (2016)

\bibitem{peters2017elements}
Peters, J., Janzing, D., Schlkopf, B.: Elements of Causal Inference: Foundations and Learning Algorithms. The MIT Press (2017)

\bibitem{pham2021facial}
Pham, L., Vu, T.H., Tran, T.A.: Facial expression recognition using residual masking network. In: 2020 25th International Conference on Pattern Recognition (ICPR). pp. 4513--4519 (2021). \doi{10.1109/ICPR48806.2021.9411919}

\bibitem{piao2021inverting}
Piao, J., Sun, K., Wang, Q., Lin, K.Y., Li, H.: Inverting generative adversarial renderer for face reconstruction. In: Proceedings of the {{IEEE}}/{{CVF}} Conference on Computer Vision and Pattern Recognition. pp. 15619--15628 (2021)

\bibitem{pumarola2018GANimation}
Pumarola, A., Agudo, A., Martinez, A.M., Sanfeliu, A., {Moreno-Noguer}, F.: {{GANimation}}: {{Anatomically-aware}} facial animation from a single image. In: Ferrari, V., Hebert, M., Sminchisescu, C., Weiss, Y. (eds.) Computer Vision -- {{ECCV}} 2018. pp. 835--851. Springer International Publishing, Cham (2018)

\bibitem{pumarola2018unsupervised}
Pumarola, A., Agudo, A., Sanfeliu, A., {Moreno-Noguer}, F.: Unsupervised person image synthesis in arbitrary poses. In: Proceedings of the {{IEEE}} Conference on Computer Vision and Pattern Recognition. pp. 8620--8628 (2018)

\bibitem{qiuSCULPTORSkeletonConsistentFace2022}
Qiu, Z., Li, Y., He, D., Zhang, Q., Zhang, L., Zhang, Y., Wang, J., Xu, L., Wang, X., Zhang, Y., Yu, J.: {{SCULPTOR}}: {{Skeleton-Consistent Face Creation Using}} a {{Learned Parametric Generator}}. ACM Transactions on Graphics  \textbf{41}(6),  213:1--213:17 (Nov 2022). \doi{10.1145/3550454.3555462}

\bibitem{reichenbach1956direction}
Reichenbach, H.: The direction of time, vol.~65. Univ of California Press (1956)

\bibitem{reimers2021debiasing}
Reimers, C., Bodesheim, P., Runge, J., Denzler, J.: Conditional adversarial debiasing: Towards learning unbiased classifiers from biased data. In: DAGM German Conference on Pattern Recognition. pp. 48--62. Springer (2021)

\bibitem{reimers2021conditional}
Reimers, C., Penzel, N., Bodesheim, P., Runge, J., Denzler, J.: Conditional dependence tests reveal the usage of abcd rule features and bias variables in automatic skin lesion classification. In: Proceedings of the IEEE/CVF Conference on Computer Vision and Pattern Recognition. pp. 1810--1819 (2021)

\bibitem{reimers2020determining}
Reimers, C., Runge, J., Denzler, J.: Determining the relevance of features for deep neural networks. In: European Conference on Computer Vision. Springer (2020)

\bibitem{Ribeiro2016WhySI}
Ribeiro, M.T., Singh, S., Guestrin, C.: “why should i trust you?”: Explaining the predictions of any classifier. Proceedings of the 22nd ACM SIGKDD International Conference on Knowledge Discovery and Data Mining  (2016), \url{https://api.semanticscholar.org/CorpusID:13029170}

\bibitem{richardson20163d}
Richardson, E., Sela, M., Kimmel, R.: 3d face reconstruction by learning from synthetic data. In: 2016 fourth international conference on 3D vision (3DV). pp. 460--469. IEEE (2016)

\bibitem{roberts1996empathy}
Roberts, W., Strayer, J.: Empathy, emotional expressiveness, and prosocial behavior. Child development  \textbf{67}(2),  449--470 (1996)

\bibitem{ros2016synthia}
Ros, G., Sellart, L., Materzynska, J., Vazquez, D., Lopez, A.M.: The synthia dataset: A large collection of synthetic images for semantic segmentation of urban scenes. In: Proceedings of the IEEE conference on computer vision and pattern recognition. pp. 3234--3243 (2016)

\bibitem{rossDevelopmentSensitiveClinical1996}
Ross, B.G., Fradet, G., Nedzelski, J.M.: Development of a {{Sensitive Clinical Facial Grading System}}. Otolaryngology--Head and Neck Surgery  \textbf{114}(3),  380--386 (Mar 1996). \doi{10.1016/S0194-59989670206-1}

\bibitem{runge2018conditional}
Runge, J.: Conditional independence testing based on a nearest-neighbor estimator of conditional mutual information. In: International Conference on Artificial Intelligence and Statistics. PMLR (2018)

\bibitem{saleh2018effective}
Saleh, F.S., Aliakbarian, M.S., Salzmann, M., Petersson, L., Alvarez, J.M.: Effective use of synthetic data for urban scene semantic segmentation. In: Proceedings of the European Conference on Computer Vision (ECCV). pp. 84--100 (2018)

\bibitem{sankaranarayanan2018learning}
Sankaranarayanan, S., Balaji, Y., Jain, A., Lim, S.N., Chellappa, R.: Learning from synthetic data: Addressing domain shift for semantic segmentation. In: Proceedings of the IEEE conference on computer vision and pattern recognition. pp. 3752--3761 (2018)

\bibitem{savchenko2023facial}
Savchenko, A.: Facial expression recognition with adaptive frame rate based on multiple testing correction. In: International Conference on Machine Learning. vol.~202. PMLR (2023), \url{https://proceedings.mlr.press/v202/savchenko23a.html}

\bibitem{Savchenko_2022_CVPRW}
Savchenko, A.V.: Video-based frame-level facial analysis of affective behavior on mobile devices using {{EfficientNets}}. In: Proceedings of the {{IEEE}}/{{CVF}} Conference on Computer Vision and Pattern Recognition ({{CVPR}}) Workshops. pp. 2359--2366 (Jun 2022)

\bibitem{savchenko2022classifying}
Savchenko, A.V., Savchenko, L.V., Makarov, I.: Classifying emotions and engagement in online learning based on a single facial expression recognition neural network. IEEE Transactions on Affective Computing  (2022)

\bibitem{Selvaraju2016GradCAMVE}
Selvaraju, R.R., Das, A., Vedantam, R., Cogswell, M., Parikh, D., Batra, D.: Grad-cam: Visual explanations from deep networks via gradient-based localization. International Journal of Computer Vision  \textbf{128},  336 -- 359 (2016), \url{https://api.semanticscholar.org/CorpusID:15019293}

\bibitem{shah2020hardness}
Shah, R.D., Peters, J.: The hardness of conditional independence testing and the generalised covariance measure. The Annals of Statistics  \textbf{48}(3),  1514--1538 (2020)

\bibitem{Smilkov2017SmoothGradRN}
Smilkov, D., Thorat, N., Kim, B., Vi{\'e}gas, F.B., Wattenberg, M.: Smoothgrad: removing noise by adding noise. ArXiv  \textbf{abs/1706.03825} (2017), \url{https://api.semanticscholar.org/CorpusID:11695878}

\bibitem{Springenberg2014StrivingFS}
Springenberg, J.T., Dosovitskiy, A., Brox, T., Riedmiller, M.A.: Striving for simplicity: The all convolutional net. CoRR  \textbf{abs/1412.6806} (2014), \url{https://api.semanticscholar.org/CorpusID:12998557}

\bibitem{stornDifferentialEvolutionSimple1997}
Storn, R., Price, K.: Differential {{Evolution}} -- {{A Simple}} and {{Efficient Heuristic}} for global {{Optimization}} over {{Continuous Spaces}}. Journal of Global Optimization  \textbf{11}(4),  341--359 (Dec 1997). \doi{10.1023/A:1008202821328}

\bibitem{Sundararajan2017AxiomaticAF}
Sundararajan, M., Taly, A., Yan, Q.: Axiomatic attribution for deep networks. In: International Conference on Machine Learning (2017), \url{https://api.semanticscholar.org/CorpusID:16747630}

\bibitem{lesson}
Sutton, R.: The bitter lesson (2019)

\bibitem{takmaz20233d}
Takmaz, A., Schult, J., Kaftan, I., Ak{\c{c}}ay, M., Leibe, B., Sumner, R., Engelmann, F., Tang, S.: 3d segmentation of humans in point clouds with synthetic data. In: Proceedings of the IEEE/CVF International Conference on Computer Vision. pp. 1292--1304 (2023)

\bibitem{tewari2020stylerig}
Tewari, A., Elgharib, M., Bharaj, G., Bernard, F., Seidel, H.P., P{\'e}rez, P., Z{\"o}llhofer, M., Theobalt, C.: {{StyleRig}}: {{Rigging StyleGAN}} for {{3D}} control over portrait images, {{CVPR}} 2020. In: {{IEEE}} Conference on Computer Vision and Pattern Recognition ({{CVPR}}). IEEE (Jun 2020)

\bibitem{thalhammer2019sydpose}
Thalhammer, S., Patten, T., Vincze, M.: Sydpose: Object detection and pose estimation in cluttered real-world depth images trained using only synthetic data. In: 2019 International Conference on 3D Vision (3DV). pp. 106--115. IEEE (2019)

\bibitem{thompson1987empathy}
Thompson, R.A.: Empathy and emotional understanding: The early development of empathy. Empathy and its development  \textbf{119}, ~145 (1987)

\bibitem{tremblay2018falling}
Tremblay, J., To, T., Birchfield, S.: Falling things: A synthetic dataset for 3d object detection and pose estimation. In: Proceedings of the IEEE Conference on Computer Vision and Pattern Recognition Workshops. pp. 2038--2041 (2018)

\bibitem{vanherle2022analysis}
Vanherle, B., Moonen, S., Van~Reeth, F., Michiels, N.: Analysis of training object detection models with synthetic data. arXiv preprint arXiv:2211.16066  (2022)

\bibitem{vigneshNovelFacialEmotion2023}
Vignesh, S., Savithadevi, M., Sridevi, M., Sridhar, R.: A novel facial emotion recognition model using segmentation {{VGG-19}} architecture. International Journal of Information Technology  \textbf{15}(4),  1777--1787 (Apr 2023). \doi{10.1007/s41870-023-01184-z}

\bibitem{wagnerSoftDECAComputationallyEfficient2023}
Wagner, N., Botsch, M., Schwanecke, U.: {{SoftDECA}}: {{Computationally Efficient Physics-Based Facial Animations}}. In: Proceedings of the 16th {{ACM SIGGRAPH Conference}} on {{Motion}}, {{Interaction}} and {{Games}}. pp. 1--11. {{MIG}} '23, Association for Computing Machinery, New York, NY, USA (Nov 2023). \doi{10.1145/3623264.3624439}

\bibitem{wasiARBExAttentiveFeature2023}
Wasi, A.T., {\v S}erbetar, K., Islam, R., Rafi, T.H., Chae, D.K.: {{ARBEx}}: {{Attentive Feature Extraction}} with {{Reliability Balancing}} for {{Robust Facial Expression Learning}} (Jul 2023)

\bibitem{weihererApproximatingIntersectionsDifferences2024}
Weiherer, M., Klein, F., Egger, B.: Approximating {{Intersections}} and {{Differences Between Linear Statistical Shape Models Using Markov Chain Monte Carlo}}. In: 2024 {{IEEE}}/{{CVF Winter Conference}} on {{Applications}} of {{Computer Vision}} ({{WACV}}). pp. 6352--6361. IEEE, Waikoloa, HI, USA (Jan 2024). \doi{10.1109/WACV57701.2024.00624}

\bibitem{wenDistractYourAttention2023}
Wen, Z., Lin, W., Wang, T., Xu, G.: Distract {{Your Attention}}: {{Multi-head Cross Attention Network}} for {{Facial Expression Recognition}}. Biomimetics  \textbf{8}(2), ~199 (May 2023). \doi{10.3390/biomimetics8020199}

\bibitem{wu2022synthetic}
Wu, Z., Wang, L., Wang, W., Shi, T., Chen, C., Hao, A., Li, S.: Synthetic data supervised salient object detection. In: Proceedings of the 30th ACM international conference on multimedia. pp. 5557--5565 (2022)

\bibitem{yang2020facescape}
Yang, H., Zhu, H., Wang, Y., Huang, M., Shen, Q., Yang, R., Cao, X.: {{FaceScape}}: A large-scale high quality {{3D}} face dataset and detailed riggable {{3D}} face prediction. In: Proceedings of the {{IEEE}} Conference on Computer Vision and Pattern Recognition ({{CVPR}}) (2020)

\bibitem{yangLearningGeneralizedPhysical2024}
Yang, L., Zoss, G., Chandran, P., Gross, M., Solenthaler, B., Sifakis, E., Bradley, D.: Learning a {{Generalized Physical Face Model From Data}} (Feb 2024)

\bibitem{yang2023change}
Yang, Y., Zhang, H., Katabi, D., Ghassemi, M.: Change is hard: A closer look at subpopulation shift. arXiv preprint arXiv:2302.12254  (2023)

\bibitem{zeiler2014visualizing}
Zeiler, M.D., Fergus, R.: Visualizing and understanding convolutional networks. In: Computer Vision--ECCV 2014: 13th European Conference, Zurich, Switzerland, September 6-12, 2014, Proceedings, Part I 13. pp. 818--833. Springer (2014)

\bibitem{zhangUnreasonableEffectivenessDeep2018}
Zhang, R., Isola, P., Efros, A.A., Shechtman, E., Wang, O.: The {{Unreasonable Effectiveness}} of {{Deep Features}} as a {{Perceptual Metric}}. Proceedings of the IEEE Conference on Computer Vision and Pattern Recognition  (Apr 2018). \doi{10.48550/arXiv.1801.03924}

\bibitem{zhang2018unreasonable}
Zhang, R., Isola, P., Efros, A.A., Shechtman, E., Wang, O.: The {{Unreasonable Effectiveness}} of {{Deep Features}} as a {{Perceptual Metric}}. Proceedings of the IEEE Conference on Computer Vision and Pattern Recognition  (Apr 2018). \doi{10.48550/arXiv.1801.03924}

\bibitem{zhangDualDirectionAttentionMixed2023}
Zhang, S., Zhang, Y., Zhang, Y., Wang, Y., Song, Z.: A {{Dual-Direction Attention Mixed Feature Network}} for {{Facial Expression Recognition}}. Electronics  \textbf{12}(17), ~3595 (Jan 2023). \doi{10.3390/electronics12173595}

\bibitem{zhouExploringEmotionFeatures2019}
Zhou, H., Meng, D., Zhang, Y., Peng, X., Du, J., Wang, K., Qiao, Y.: Exploring {{Emotion Features}} and {{Fusion Strategies}} for {{Audio-Video Emotion Recognition}}. In: 2019 {{International Conference}} on {{Multimodal Interaction}}. pp. 562--566 (Oct 2019). \doi{10.1145/3340555.3355713}

\bibitem{zhu2023facescape}
Zhu, H., Yang, H., Guo, L., Zhang, Y., Wang, Y., Huang, M., Wu, Menghua {and}~Shen, Q., Yang, R., Cao, X.: {{FaceScape}}: {{3D}} facial dataset and benchmark for single-view {{3D}} face reconstruction. IEEE Transactions on Pattern Analysis and Machine Intelligence (TPAMI)  (2023)

\end{thebibliography}
\newpage
\appendix
\section{Structural Causal Models}
Here we include a technical definition of structural causal models used in our work.

\begin{definition}[Structural Causal Model in \protect{\cite[Sec. 7.1.1]{pearl2009causality}} and \protect{\cite[Def.~1]{bareinboim2022onpearl}}]
A structural causal model (SCM) is defined as a 4-tuple $M=(U, V, F, P)$, where $U$ is a set of exogenous variables describing outside factors, $V=\{V_1, ..., V_n\}$ is the set of endogenous variables we measure in our model, $F=\{f_2, ..., f_n\}$ is a set containing functions $f_i$ that describe the functional relationships, and $P$ is a joint probability distribution over $U$.
Further, each $V_i$ has a set of parents $PA_i$ that functionally determine $V_i$ together with some exogenous variables $U_i \subseteq U$. 
These parents $PA_i$ are a subset of $V\setminus\{V_i\}$. 
For settings $pa_i$ of parents $PA_i$ and $u_i$ of the exogenous variables $U_i$, $f_i$ determines the value $v_i = f_i(pa_i, u_i)$ of $V_i$.
\end{definition}

Each causal model $M$ can be visualized as directed graphs.
Here, each variable $V_i$ in $V$ defines a node, and we draw directed links from all parents $PA_i$ into $V_i$.
Using such a model $M$, we can investigate questions of the following nature: given observed evidence, e.g., $V_j=v_j$, what is the probability of a statement $A$ happening?
Further, performing a \emph{do}-action on $V_i \in V$ is equivalent to removing the dependency $f_i$ and instead forcing $V_i$ to a constant value $x$.
In other words, we set $F$ to $F_x$ with $F_x = \{f_j: V_j \neq V_i\} \cup \{V_i \leftarrow x\}$ \cite{bareinboim2022onpearl}.

\section{Measuring Systematic Change - Significance Test}

\begin{figure}
    \centering
    \includegraphics[width=\textwidth]{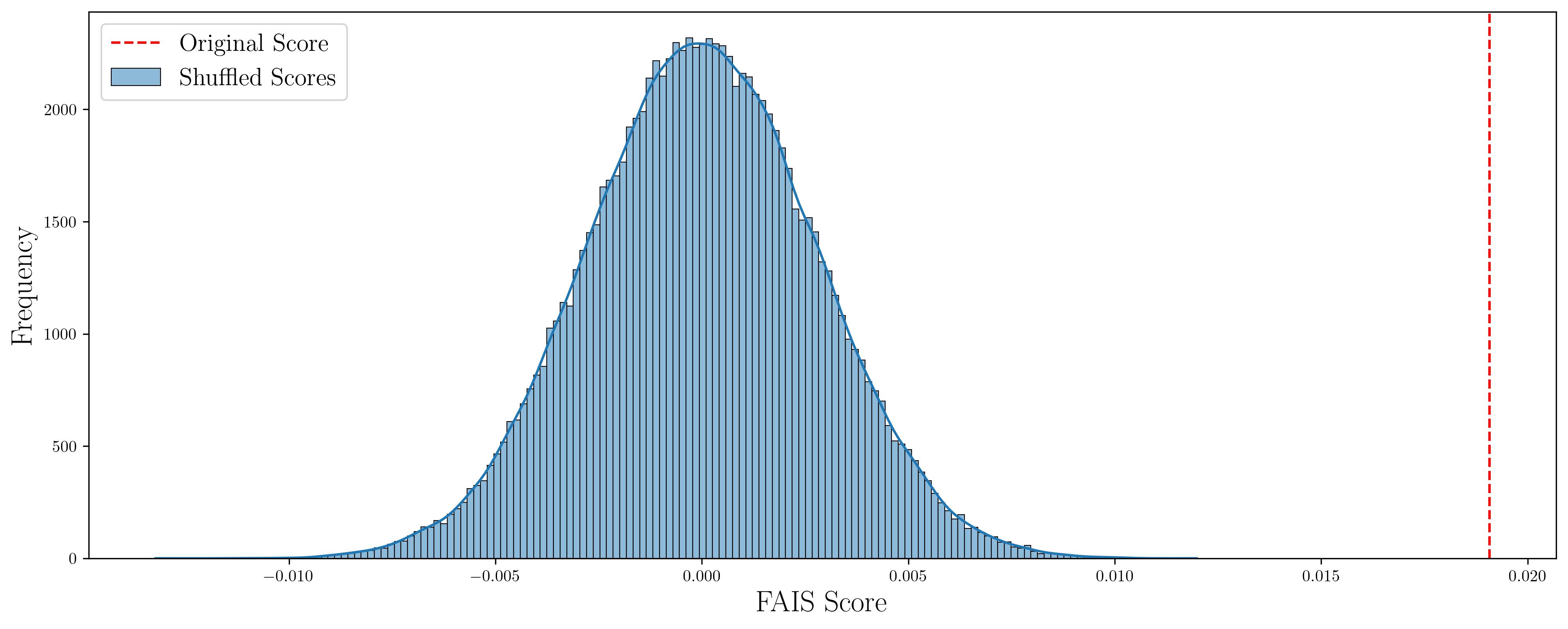}
    \caption{
        Using the shuffle test, outlined in \Cref{alg:sig}, we plot the resulting scores for $100000$ permutations in a histogram.
        Our tests use a significance threshold of $p < (0.05)$.
        The original model score for a single individual, shown as a red dashed line, lies clearly outside the computed null distribution and is thus significant.
    }
    \label{fig:sig-test}
\end{figure}

In \Cref{alg:sig}, we provide detailed pseudo code for our proposed shuffle hypothesis test regarding the significance of $\hat{\Sym}(\eclass_\ecweights^{(\emo)}|\Iphi)$.
Further in \Cref{fig:sig-test}, we visualize the estimated null-distribution as well as the originally measured score.
We see that randomly shuffling observations along the symmetry axis results in a symmetrical distribution centered around zero, i.e., no systematic dependence on the facial symmetry.
The original score, for the example in \Cref{fig:sig-test}, is not typical for the estimated null-distribution leading to a low p-value.
\Cref{tab:sigtest} contains the number of individuals per classifier and expression for which \Cref{alg:sig} together with the Holm-Bonferroni correction \cite{holm1979simple} is significant.
For this analysis, we perform the shuffle test with 10K iterations.
\underline{All} 17 classifiers show significant behavior changes concerning facial symmetry for all expressions and a majority of individuals.

\begin{algorithm}
\caption{Testing for statistical significance of $\hat{\Sym}(\eclass_\ecweights^{(\emo)}|\Iphi)$.}\label{alg:sig}
\begin{algorithmic}
\Require grid of predictions $\eclass_\ecweights^{(\emo)}(\Iphi(s,t))$ \Comment{Gridsize is $S\times T$}
\Require integer K > 0 \Comment{Number of Permutations}
\Require $\siglevel \in (0,1)$ \Comment{Significance Level}
\State $p \gets 0.0$
\State $\sigma_{orig.} \gets \hat{\Sym}(\eclass_\ecweights^{(\emo)}|\Iphi)$ \Comment{Estimate the original statistic}
\For{\texttt{$i \in \{1, ..., K\}$}}
    \State $\eclass_\ecweights^{(\emo)}(\Iphi^{(perm.)}(s,t)) \gets \texttt{permute}(\eclass_\ecweights^{(\emo)}(\Iphi(s,t)), \texttt{axis} = 0)$ \\\Comment{Shuffle along Symmetry Axis}
    \State $\sigma_{perm.} \gets \hat{\Sym}(\eclass_\ecweights^{(\emo)}|\Iphi^{(perm.)})$
    \If{$|\sigma_{perm.}| > |\sigma_{orig.}$|} \Comment{Absolutes because our statistic is two sided}
        \State $p \gets p + \nicefrac{1}{K}$   \Comment{Increment the $p$-value}
    \EndIf
\EndFor

\If{$p < \delta$}
    \State \texttt{return} $\hat{\Sym}(\eclass_\ecweights^{(\emo)}|\Iphi)$ is significant.
\Else
    \State \texttt{return} $\hat{\Sym}(\eclass_\ecweights^{(\emo)}|\Iphi)$ is not significant.
\EndIf
\end{algorithmic}
\end{algorithm}

\begin{table}[t]
    \centering
    \caption{
        We report how many of the 200 individuals, using the Holm-Bonferroni \cite{holm1979simple} corrected p-values, have been significant ($p < 0.05$).
        We can see that the majority of all results are significant, confirming our hypothesis that facial symmetry impacts the internal decision rules.
    }
    \begin{tabular}{llrrrrrr}
\toprule
Dataset & Model & \multicolumn{1}{c}{Angry} & \multicolumn{1}{c}{Disgust} & \multicolumn{1}{c}{Fear} & \multicolumn{1}{c}{Happy} & \multicolumn{1}{c}{Sad} & \multicolumn{1}{c}{Surprise} \\
\midrule
\multirow[c]{4}{*}{AffectNet7} 
& DAN~\cite{wenDistractYourAttention2023} & {\cellcolor[HTML]{3F007D}} \color[HTML]{F1F1F1} 200 & {\cellcolor[HTML]{3F007D}} \color[HTML]{F1F1F1} 200 & {\cellcolor[HTML]{3F007D}} \color[HTML]{F1F1F1} 200 & {\cellcolor[HTML]{3F007D}} \color[HTML]{F1F1F1} 200 & {\cellcolor[HTML]{3F007D}} \color[HTML]{F1F1F1} 200 & {\cellcolor[HTML]{3F007D}} \color[HTML]{F1F1F1} 200 \\
 & DDAMFN++~\cite{zhangDualDirectionAttentionMixed2023} & {\cellcolor[HTML]{3F007D}} \color[HTML]{F1F1F1} 200 & {\cellcolor[HTML]{3F007D}} \color[HTML]{F1F1F1} 200 & {\cellcolor[HTML]{3F007D}} \color[HTML]{F1F1F1} 200 & {\cellcolor[HTML]{3F007D}} \color[HTML]{F1F1F1} 200 & {\cellcolor[HTML]{3F007D}} \color[HTML]{F1F1F1} 200 & {\cellcolor[HTML]{3F007D}} \color[HTML]{F1F1F1} 200 \\
 & HSEmotion~\cite{savchenko2023facial} & {\cellcolor[HTML]{3F007D}} \color[HTML]{F1F1F1} 200 & {\cellcolor[HTML]{3F007D}} \color[HTML]{F1F1F1} 200 & {\cellcolor[HTML]{3F007D}} \color[HTML]{F1F1F1} 200 & {\cellcolor[HTML]{3F007D}} \color[HTML]{F1F1F1} 200 & {\cellcolor[HTML]{3F007D}} \color[HTML]{F1F1F1} 200 & {\cellcolor[HTML]{3F007D}} \color[HTML]{F1F1F1} 200 \\
 & PosterV2~\cite{maoPOSTERSimplerStronger2023} & {\cellcolor[HTML]{3F007D}} \color[HTML]{F1F1F1} 200 & {\cellcolor[HTML]{41047F}} \color[HTML]{F1F1F1} 199 & {\cellcolor[HTML]{3F007D}} \color[HTML]{F1F1F1} 200 & {\cellcolor[HTML]{3F007D}} \color[HTML]{F1F1F1} 200 & {\cellcolor[HTML]{3F007D}} \color[HTML]{F1F1F1} 200 & {\cellcolor[HTML]{3F007D}} \color[HTML]{F1F1F1} 200 \\
\midrule
\multirow[c]{4}{*}{AffectNet8} & DAN~\cite{wenDistractYourAttention2023} & {\cellcolor[HTML]{3F007D}} \color[HTML]{F1F1F1} 200 & {\cellcolor[HTML]{3F007D}} \color[HTML]{F1F1F1} 200 & {\cellcolor[HTML]{3F007D}} \color[HTML]{F1F1F1} 200 & {\cellcolor[HTML]{3F007D}} \color[HTML]{F1F1F1} 200 & {\cellcolor[HTML]{3F007D}} \color[HTML]{F1F1F1} 200 & {\cellcolor[HTML]{3F007D}} \color[HTML]{F1F1F1} 200 \\
 & DDAMFN++~\cite{zhangDualDirectionAttentionMixed2023} & {\cellcolor[HTML]{3F007D}} \color[HTML]{F1F1F1} 200 & {\cellcolor[HTML]{3F007D}} \color[HTML]{F1F1F1} 200 & {\cellcolor[HTML]{440981}} \color[HTML]{F1F1F1} 198 & {\cellcolor[HTML]{3F007D}} \color[HTML]{F1F1F1} 200 & {\cellcolor[HTML]{3F007D}} \color[HTML]{F1F1F1} 200 & {\cellcolor[HTML]{3F007D}} \color[HTML]{F1F1F1} 200 \\
 & HSEmotion~\cite{savchenko2023facial} & {\cellcolor[HTML]{3F007D}} \color[HTML]{F1F1F1} 200 & {\cellcolor[HTML]{3F007D}} \color[HTML]{F1F1F1} 200 & {\cellcolor[HTML]{3F007D}} \color[HTML]{F1F1F1} 200 & {\cellcolor[HTML]{3F007D}} \color[HTML]{F1F1F1} 200 & {\cellcolor[HTML]{3F007D}} \color[HTML]{F1F1F1} 200 & {\cellcolor[HTML]{3F007D}} \color[HTML]{F1F1F1} 200 \\
 & PosterV2~\cite{maoPOSTERSimplerStronger2023} & {\cellcolor[HTML]{3F007D}} \color[HTML]{F1F1F1} 200 & {\cellcolor[HTML]{3F007D}} \color[HTML]{F1F1F1} 200 & {\cellcolor[HTML]{3F007D}} \color[HTML]{F1F1F1} 200 & {\cellcolor[HTML]{3F007D}} \color[HTML]{F1F1F1} 200 & {\cellcolor[HTML]{3F007D}} \color[HTML]{F1F1F1} 200 & {\cellcolor[HTML]{3F007D}} \color[HTML]{F1F1F1} 200 \\
\midrule
\multirow[c]{6}{*}{FER2013} 
 & EmoNeXt-Tiny$^{\dagger}$~\cite{boundori2023EmoNext} & {\cellcolor[HTML]{3F007D}} \color[HTML]{F1F1F1} 200 & {\cellcolor[HTML]{3F007D}} \color[HTML]{F1F1F1} 200 & {\cellcolor[HTML]{3F007D}} \color[HTML]{F1F1F1} 200 & {\cellcolor[HTML]{3F007D}} \color[HTML]{F1F1F1} 200 & {\cellcolor[HTML]{3F007D}} \color[HTML]{F1F1F1} 200 & {\cellcolor[HTML]{3F007D}} \color[HTML]{F1F1F1} 200 \\
 & EmoNeXt-Small$^{\dagger}$~\cite{boundori2023EmoNext} & {\cellcolor[HTML]{41047F}} \color[HTML]{F1F1F1} 199 & {\cellcolor[HTML]{4D1A89}} \color[HTML]{F1F1F1} 194 & {\cellcolor[HTML]{3F007D}} \color[HTML]{F1F1F1} 200 & {\cellcolor[HTML]{3F007D}} \color[HTML]{F1F1F1} 200 & {\cellcolor[HTML]{3F007D}} \color[HTML]{F1F1F1} 200 & {\cellcolor[HTML]{3F007D}} \color[HTML]{F1F1F1} 200 \\
& EmoNeXt-Base$^{\dagger}$~\cite{boundori2023EmoNext}  & {\cellcolor[HTML]{A3A0CB}} \color[HTML]{F1F1F1} 163 & {\cellcolor[HTML]{F1EFF6}} \color[HTML]{000000} 137 & {\cellcolor[HTML]{3F007D}} \color[HTML]{F1F1F1} 200 & {\cellcolor[HTML]{3F007D}} \color[HTML]{F1F1F1} 200 & {\cellcolor[HTML]{3F007D}} \color[HTML]{F1F1F1} 200 & {\cellcolor[HTML]{3F007D}} \color[HTML]{F1F1F1} 200 \\
 & EmoNeXt-Large$^{\dagger}$~\cite{boundori2023EmoNext}  & {\cellcolor[HTML]{F6F5F9}} \color[HTML]{000000} 133 & {\cellcolor[HTML]{5E3A98}} \color[HTML]{F1F1F1} 187 & {\cellcolor[HTML]{481185}} \color[HTML]{F1F1F1} 196 & {\cellcolor[HTML]{4B1687}} \color[HTML]{F1F1F1} 195 & {\cellcolor[HTML]{440981}} \color[HTML]{F1F1F1} 198 & {\cellcolor[HTML]{5E3A98}} \color[HTML]{F1F1F1} 187 \\
 & ResidualMaskingNet~\cite{pham2021facial} & {\cellcolor[HTML]{3F007D}} \color[HTML]{F1F1F1} 200 & {\cellcolor[HTML]{4D1A89}} \color[HTML]{F1F1F1} 194 & {\cellcolor[HTML]{41047F}} \color[HTML]{F1F1F1} 199 & {\cellcolor[HTML]{3F007D}} \color[HTML]{F1F1F1} 200 & {\cellcolor[HTML]{3F007D}} \color[HTML]{F1F1F1} 200 & {\cellcolor[HTML]{3F007D}} \color[HTML]{F1F1F1} 200 \\
 & Segmentation-VGG19$^{\dagger}$~\cite{vigneshNovelFacialEmotion2023} & {\cellcolor[HTML]{41047F}} \color[HTML]{F1F1F1} 199 & {\cellcolor[HTML]{D6D6E9}} \color[HTML]{000000} 148 & {\cellcolor[HTML]{3F007D}} \color[HTML]{F1F1F1} 200 & {\cellcolor[HTML]{D6D6E9}} \color[HTML]{000000} 148 & {\cellcolor[HTML]{3F007D}} \color[HTML]{F1F1F1} 200 & {\cellcolor[HTML]{FCFBFD}} \color[HTML]{000000} 129 \\
\midrule
\multirow[c]{3}{*}{RAFDB} & DAN~\cite{wenDistractYourAttention2023} & {\cellcolor[HTML]{3F007D}} \color[HTML]{F1F1F1} 200 & {\cellcolor[HTML]{41047F}} \color[HTML]{F1F1F1} 199 & {\cellcolor[HTML]{3F007D}} \color[HTML]{F1F1F1} 200 & {\cellcolor[HTML]{3F007D}} \color[HTML]{F1F1F1} 200 & {\cellcolor[HTML]{3F007D}} \color[HTML]{F1F1F1} 200 & {\cellcolor[HTML]{3F007D}} \color[HTML]{F1F1F1} 200 \\
 & DDAMFN++~\cite{zhangDualDirectionAttentionMixed2023} & {\cellcolor[HTML]{3F007D}} \color[HTML]{F1F1F1} 200 & {\cellcolor[HTML]{3F007D}} \color[HTML]{F1F1F1} 200 & {\cellcolor[HTML]{3F007D}} \color[HTML]{F1F1F1} 200 & {\cellcolor[HTML]{3F007D}} \color[HTML]{F1F1F1} 200 & {\cellcolor[HTML]{3F007D}} \color[HTML]{F1F1F1} 200 & {\cellcolor[HTML]{3F007D}} \color[HTML]{F1F1F1} 200 \\
 & PosterV2~\cite{maoPOSTERSimplerStronger2023} & {\cellcolor[HTML]{3F007D}} \color[HTML]{F1F1F1} 200 & {\cellcolor[HTML]{4D1A89}} \color[HTML]{F1F1F1} 194 & {\cellcolor[HTML]{3F007D}} \color[HTML]{F1F1F1} 200 & {\cellcolor[HTML]{3F007D}} \color[HTML]{F1F1F1} 200 & {\cellcolor[HTML]{4B1687}} \color[HTML]{F1F1F1} 195 & {\cellcolor[HTML]{3F007D}} \color[HTML]{F1F1F1} 200 \\
\bottomrule
\end{tabular}

    \label{tab:sigtest}
\end{table}

\section{Additional Details Experiment 1}

This section gives an overview of the prediction accuracy of all 17 expression classifiers achieved on our real-world data, consisting of healthy probands and patients with unilateral facial palsy.
Further, we detail the hyperparameter choices in our experiments regarding the associational methods to infer whether a causal link exists between facial symmetry and model prediction behavior. 
Lastly, we include some additional visualizations regarding the symmetry features.

\subsection{Real-World Prediction Accuracy}
\begin{table}[t]
    \centering
    \caption{
        We evaluated each classifier on the faces of the healthy probands and patients with unilateral facial palsy mimicking the \happy facial expression.
        Low accuracy is displayed in a \colorbox{lowacc}{\color[HTML]{F1F1F1} darks shade}, and high accuracy is displayed in a \colorbox{highacc}{\color[HTML]{000000} light shade}.
        Models trained on FER2013 especially seem to work well on our data set.
        Models trained on RAFDB seem to be less suitable.
        Further, we provide the mean accuracy of all models per data set (with at least one correct classification).
    }
    \begin{tabular}{llrrrrr}
\toprule
Dataset & Model & \multicolumn{1}{c}{$s=0.0$} & \multicolumn{1}{c}{$s=0.5$} & \multicolumn{1}{c}{$s=1.0$} & \multicolumn{1}{c}{Probands} & \multicolumn{1}{c}{Patients} \\

\midrule

\multirow[c]{5}{*}{AffectNet7} 
& DAN~\cite{wenDistractYourAttention2023} & {\cellcolor[HTML]{1FA287}} \color[HTML]{F1F1F1} 57.50\% & {\cellcolor[HTML]{F8E621}} \color[HTML]{000000} 99.00\% & {\cellcolor[HTML]{FDE725}} \color[HTML]{000000} 100.00\% & {\cellcolor[HTML]{DAE319}} \color[HTML]{000000} 94.44\% & {\cellcolor[HTML]{28AE80}} \color[HTML]{F1F1F1} 62.82\% \\
& DDAMFN++~\cite{zhangDualDirectionAttentionMixed2023} & {\cellcolor[HTML]{440154}} \color[HTML]{F1F1F1} 0.00\% & {\cellcolor[HTML]{440154}} \color[HTML]{F1F1F1} 0.00\% & {\cellcolor[HTML]{424186}} \color[HTML]{F1F1F1} 19.50\% & {\cellcolor[HTML]{440154}} \color[HTML]{F1F1F1} 0.35\% & {\cellcolor[HTML]{440154}} \color[HTML]{F1F1F1} 0.20\% \\
 & HSEmotion~\cite{savchenko2023facial} & {\cellcolor[HTML]{E5E419}} \color[HTML]{000000} 96.00\% & {\cellcolor[HTML]{FDE725}} \color[HTML]{000000} 100.00\% & {\cellcolor[HTML]{FDE725}} \color[HTML]{000000} 100.00\% & {\cellcolor[HTML]{440154}} \color[HTML]{F1F1F1} 0.00\% & {\cellcolor[HTML]{440154}} \color[HTML]{F1F1F1} 0.00\% \\
& PosterV2~\cite{maoPOSTERSimplerStronger2023} & {\cellcolor[HTML]{A8DB34}} \color[HTML]{000000} 87.00\% & {\cellcolor[HTML]{FDE725}} \color[HTML]{000000} 100.00\% & {\cellcolor[HTML]{FDE725}} \color[HTML]{000000} 100.00\% & {\cellcolor[HTML]{F1E51D}} \color[HTML]{000000} 97.92\% & {\cellcolor[HTML]{25AB82}} \color[HTML]{F1F1F1} 61.23\% \\
\cline{2-6}
& Average & {\cellcolor[HTML]{7CD250}} \color[HTML]{000000} 80.17\% & {\cellcolor[HTML]{FDE725}} \color[HTML]{000000} 99.67\% & {\cellcolor[HTML]{7AD151}} \color[HTML]{000000} 79.88\% & {\cellcolor[HTML]{2DB27D}} \color[HTML]{F1F1F1} 64.24\% & {\cellcolor[HTML]{287C8E}} \color[HTML]{F1F1F1} 41.42\% \\

\midrule

\multirow[c]{5}{*}{AffectNet8} 
& DAN~\cite{wenDistractYourAttention2023} & {\cellcolor[HTML]{440154}} \color[HTML]{F1F1F1} 0.00\% & {\cellcolor[HTML]{34618D}} \color[HTML]{F1F1F1} 30.50\% & {\cellcolor[HTML]{B2DD2D}} \color[HTML]{000000} 88.50\% & {\cellcolor[HTML]{C5E021}} \color[HTML]{000000} 91.32\% & {\cellcolor[HTML]{24868E}} \color[HTML]{F1F1F1} 45.92\% \\
& DDAMFN++~\cite{zhangDualDirectionAttentionMixed2023} & {\cellcolor[HTML]{440154}} \color[HTML]{F1F1F1} 0.00\% & {\cellcolor[HTML]{481D6F}} \color[HTML]{F1F1F1} 8.00\% & {\cellcolor[HTML]{25848E}} \color[HTML]{F1F1F1} 45.00\% & {\cellcolor[HTML]{440154}} \color[HTML]{F1F1F1} 0.35\% & {\cellcolor[HTML]{48186A}} \color[HTML]{F1F1F1} 6.56\% \\
& HSEmotion~\cite{savchenko2023facial} & {\cellcolor[HTML]{440154}} \color[HTML]{F1F1F1} 0.00\% & {\cellcolor[HTML]{440154}} \color[HTML]{F1F1F1} 0.00\% & {\cellcolor[HTML]{EFE51C}} \color[HTML]{000000} 97.50\% & {\cellcolor[HTML]{440154}} \color[HTML]{F1F1F1} 0.00\% & {\cellcolor[HTML]{440154}} \color[HTML]{F1F1F1} 0.00\% \\
& PosterV2~\cite{maoPOSTERSimplerStronger2023} & {\cellcolor[HTML]{440154}} \color[HTML]{F1F1F1} 0.00\% & {\cellcolor[HTML]{404588}} \color[HTML]{F1F1F1} 20.50\% & {\cellcolor[HTML]{F8E621}} \color[HTML]{000000} 99.00\% & {\cellcolor[HTML]{BDDF26}} \color[HTML]{000000} 89.93\% & {\cellcolor[HTML]{2D708E}} \color[HTML]{F1F1F1} 36.58\% \\
\cline{2-6}
& Average & {\cellcolor[HTML]{440154}} \color[HTML]{F1F1F1} 0.00\% & {\cellcolor[HTML]{414287}} \color[HTML]{F1F1F1} 19.67\% & {\cellcolor[HTML]{8BD646}} \color[HTML]{000000} 82.50\% & {\cellcolor[HTML]{23A983}} \color[HTML]{F1F1F1} 60.53\% & {\cellcolor[HTML]{355F8D}} \color[HTML]{F1F1F1} 29.69\% \\

\midrule

\multirow[c]{7}{*}{FER2013} 
& EmoNeXt-Base$^{\dagger}$~\cite{boundori2023EmoNext}  & {\cellcolor[HTML]{482576}} \color[HTML]{F1F1F1} 10.50\% & {\cellcolor[HTML]{48C16E}} \color[HTML]{F1F1F1} 71.00\% & {\cellcolor[HTML]{EFE51C}} \color[HTML]{000000} 97.50\% & {\cellcolor[HTML]{FDE725}} \color[HTML]{000000} 99.65\% & {\cellcolor[HTML]{46C06F}} \color[HTML]{F1F1F1} 70.38\% \\
& EmoNeXt-large$^{\dagger}$~\cite{boundori2023EmoNext}  & {\cellcolor[HTML]{375B8D}} \color[HTML]{F1F1F1} 28.50\% & {\cellcolor[HTML]{6CCD5A}} \color[HTML]{000000} 77.50\% & {\cellcolor[HTML]{FDE725}} \color[HTML]{000000} 100.00\% & {\cellcolor[HTML]{F4E61E}} \color[HTML]{000000} 98.26\% & {\cellcolor[HTML]{2AB07F}} \color[HTML]{F1F1F1} 63.42\% \\
& EmoNeXt-small$^{\dagger}$~\cite{boundori2023EmoNext} & {\cellcolor[HTML]{482576}} \color[HTML]{F1F1F1} 10.50\% & {\cellcolor[HTML]{3DBC74}} \color[HTML]{F1F1F1} 68.50\% & {\cellcolor[HTML]{FDE725}} \color[HTML]{000000} 100.00\% & {\cellcolor[HTML]{F8E621}} \color[HTML]{000000} 98.96\% & {\cellcolor[HTML]{34B679}} \color[HTML]{F1F1F1} 66.40\% \\
& EmoNeXt-tiny$^{\dagger}$~\cite{boundori2023EmoNext} & {\cellcolor[HTML]{440154}} \color[HTML]{F1F1F1} 0.00\% & {\cellcolor[HTML]{472D7B}} \color[HTML]{F1F1F1} 12.50\% & {\cellcolor[HTML]{A8DB34}} \color[HTML]{000000} 87.00\% & {\cellcolor[HTML]{F1E51D}} \color[HTML]{000000} 97.92\% & {\cellcolor[HTML]{20A486}} \color[HTML]{F1F1F1} 58.45\% \\
& ResidualMaskingNet~\cite{pham2021facial} & {\cellcolor[HTML]{31688E}} \color[HTML]{F1F1F1} 33.50\% & {\cellcolor[HTML]{B5DE2B}} \color[HTML]{000000} 89.00\% & {\cellcolor[HTML]{FDE725}} \color[HTML]{000000} 100.00\% & {\cellcolor[HTML]{CDE11D}} \color[HTML]{000000} 92.36\% & {\cellcolor[HTML]{21A685}} \color[HTML]{F1F1F1} 59.05\% \\
& Segmentation-VGG19$^{\dagger}$~\cite{vigneshNovelFacialEmotion2023} & {\cellcolor[HTML]{46085C}} \color[HTML]{F1F1F1} 2.00\% & {\cellcolor[HTML]{2A778E}} \color[HTML]{F1F1F1} 39.50\% & {\cellcolor[HTML]{F8E621}} \color[HTML]{000000} 99.00\% & {\cellcolor[HTML]{ECE51B}} \color[HTML]{000000} 97.22\% & {\cellcolor[HTML]{32648E}} \color[HTML]{F1F1F1} 31.81\% \\
\cline{2-6}
& Average & {\cellcolor[HTML]{443A83}} \color[HTML]{F1F1F1} 17.00\% & {\cellcolor[HTML]{22A785}} \color[HTML]{F1F1F1} 59.67\% & {\cellcolor[HTML]{ECE51B}} \color[HTML]{000000} 97.25\% & {\cellcolor[HTML]{EFE51C}} \color[HTML]{000000} 97.40\% & {\cellcolor[HTML]{20A486}} \color[HTML]{F1F1F1} 58.25\% \\

\midrule

\multirow[c]{4}{*}{RAFDB}
& DAN~\cite{wenDistractYourAttention2023} & {\cellcolor[HTML]{355F8D}} \color[HTML]{F1F1F1} 30.00\% & {\cellcolor[HTML]{31B57B}} \color[HTML]{F1F1F1} 65.50\% & {\cellcolor[HTML]{DAE319}} \color[HTML]{000000} 94.50\% & {\cellcolor[HTML]{32658E}} \color[HTML]{F1F1F1} 32.29\% & {\cellcolor[HTML]{1F9A8A}} \color[HTML]{F1F1F1} 54.27\% \\
& DDAMFN++~\cite{zhangDualDirectionAttentionMixed2023} & {\cellcolor[HTML]{3FBC73}} \color[HTML]{F1F1F1} 69.00\% & {\cellcolor[HTML]{F8E621}} \color[HTML]{000000} 99.00\% & {\cellcolor[HTML]{FDE725}} \color[HTML]{000000} 100.00\% & {\cellcolor[HTML]{C2DF23}} \color[HTML]{000000} 90.97\% & {\cellcolor[HTML]{65CB5E}} \color[HTML]{000000} 76.34\% \\
& PosterV2~\cite{maoPOSTERSimplerStronger2023} & {\cellcolor[HTML]{3A538B}} \color[HTML]{F1F1F1} 25.50\% & {\cellcolor[HTML]{7AD151}} \color[HTML]{000000} 80.00\% & {\cellcolor[HTML]{FDE725}} \color[HTML]{000000} 100.00\% & {\cellcolor[HTML]{482071}} \color[HTML]{F1F1F1} 8.68\% & {\cellcolor[HTML]{2C738E}} \color[HTML]{F1F1F1} 38.17\% \\
\cline{2-6}
& Average & {\cellcolor[HTML]{287C8E}} \color[HTML]{F1F1F1} 41.50\% & {\cellcolor[HTML]{84D44B}} \color[HTML]{000000} 81.50\% & {\cellcolor[HTML]{F4E61E}} \color[HTML]{000000} 98.17\% & {\cellcolor[HTML]{26828E}} \color[HTML]{F1F1F1} 43.98\% & {\cellcolor[HTML]{1FA088}} \color[HTML]{F1F1F1} 56.26\% \\

\midrule
Total & Average & {\cellcolor[HTML]{297A8E}} \color[HTML]{F1F1F1} 40.91\% & {\cellcolor[HTML]{2CB17E}} \color[HTML]{F1F1F1} 64.03\% & {\cellcolor[HTML]{BDDF26}} \color[HTML]{000000} 89.85\% & {\cellcolor[HTML]{52C569}} \color[HTML]{000000} 72.71\% & {\cellcolor[HTML]{228D8D}} \color[HTML]{F1F1F1} 48.77\% \\
\bottomrule
\end{tabular}
    \label{tab:accuracy_probands_patients}
\end{table}

We are interested in the overall prediction accuracy of the model on our real-world data set consisting of $36$ healthy probands and $36$ patients with unilateral facial palsy.
Both were instructed to mimic a \happy expression.
The probands repeated the information four times in two sessions, yielding 288 images.
The patients followed the same instruction video during a ten-day biofeedback training at the hospital.
They also repeated the exercise four times during a session on the first, third, and last day of therapy.
An additional fourth session was offered after six months but was not followed up by some patients.
Thus, we obtained 503 images for the patients.

In \Cref{tab:accuracy_probands_patients}, we display the prediction accuracy of the \happy emotion.
We see strong differences per model and group.
Therefore, we also denote the average accuracy per model and dataset to understand how we can see a particular trend per dataset.
As expected and shown in the main paper, the performance of the models degrades for images that contain some form of facial asymmetry (either simulated at $s=0.0$, $s=0.5$, or actual facial palsy).
Thus, we assume that facial symmetry is the underlying cause impacting the internal decision rules of the black box classifiers.
We also see that the DDAMFN++ model trained on the AffectNet similarly performs worse on our real-world data than on the synthetic data we use for our intervention framework.
Interestingly, the RAFDB-provided checkpoints seem more robust, at least in the case of {\color{happy}\emph{happy}}.

Given that we also follow a similar experimental setup as in FER2013, models trained on it have the best performance on our data, observable in the table.
Several reasons could be involved; either mimicry and \emph{natural} facial expression have some inherent differences, the model source on public data (and human annotated) cannot differentiate, or the impact of confounding factors like camera pose.
Lighting ensures that the models focus more on facial expressions.

Models such as PosterV2 perform well in our synthetic framework (likely due to the optimized expression parameters).
Still, they seemed to overfit on the training data RAFDB as they performed worse on the probands but somehow better on the patients.

\subsection{Feature Attribution Hyperparameter Choices}

Our main experiment 2 tests the statistical dependence between expression classifier outputs and facial symmetry.
We focus on the \happy logit and find that most models change their behavior significantly for variations in facial symmetry.
We employ the feature attribution method described in \cite{reimers2020determining} toward this goal.
This method frames supervised learning as an SCM \cite{pearl2009causality} and tests whether network predictions and a pre-defined feature (facial symmetry) are conditionally independent given the reference annotation.
If we have to discard this null hypothesis, we know that the classifier output values vary significantly for changes in the investigated feature.
This procedure is motivated by Reichenbach's common cause principle \cite{reichenbach1956direction}.

Clearly, the choice of conditional independence test is an important hyperparameter choice to ensure that the results are reliable.
Further, Shah and Peters \cite{shah2020hardness} prove that there is no optimal test that can control type-I errors, i.e., false positives, irrespective of the joint latent distribution in the non-parametric case.
Because we have no knowledge about the joint distribution of all variables important in our analysis, we are exactly in the non-parametric case.
Here, we follow previous work \cite{reimers2021conditional,penzel2022investigating,penzel2023analyzing,buechner2024power}  and select multiple non-linear tests.
Specifically, we select conditional HSIC \cite{fukumizu2007kernel}, CMIknn \cite{runge2018conditional}, and FCIT \cite{chalupka2018fast}.
We consider the result from all three tests and report the majority decision \cite{reimers2021conditional}.

The selected conditional independence tests themselves have different hyperparameter choices.
First, for conditional HSIC \cite{fukumizu2007kernel}, we have to select a suitable kernel function.
We follow the suggestion of the authors and select the common radial basis functions kernel.
Additionally, we use the heuristic by Gretton et al. \cite{gretton2006kernel} to approximate suitable kernel widths for all of our three variables.
Second, similarly for CMIknn \cite{runge2018conditional}, we follow the suggested hyperparameter settings.
Specifically, we set $k_{perm.}$, i.e., the neighborhood size, to five and use ten percent of the data to estimate the conditional mutual information ($k_{CMI} = 0.1 \cot n$ for n data points).
Lastly, for FCIT \cite{chalupka2018fast}, we again follow the suggestions by the authors.
In other words, we set the number of data permutations to eight and use ten percent of the data to calculate the test statistic.

\subsection{Additional Visualizations Regarding Logit Activations}

\begin{figure}
    \centering
    \begin{subfigure}{\textwidth}
    \centering
        \includegraphics[width=0.666\textwidth, trim=0cm 0cm 0cm 1.7cm, clip]{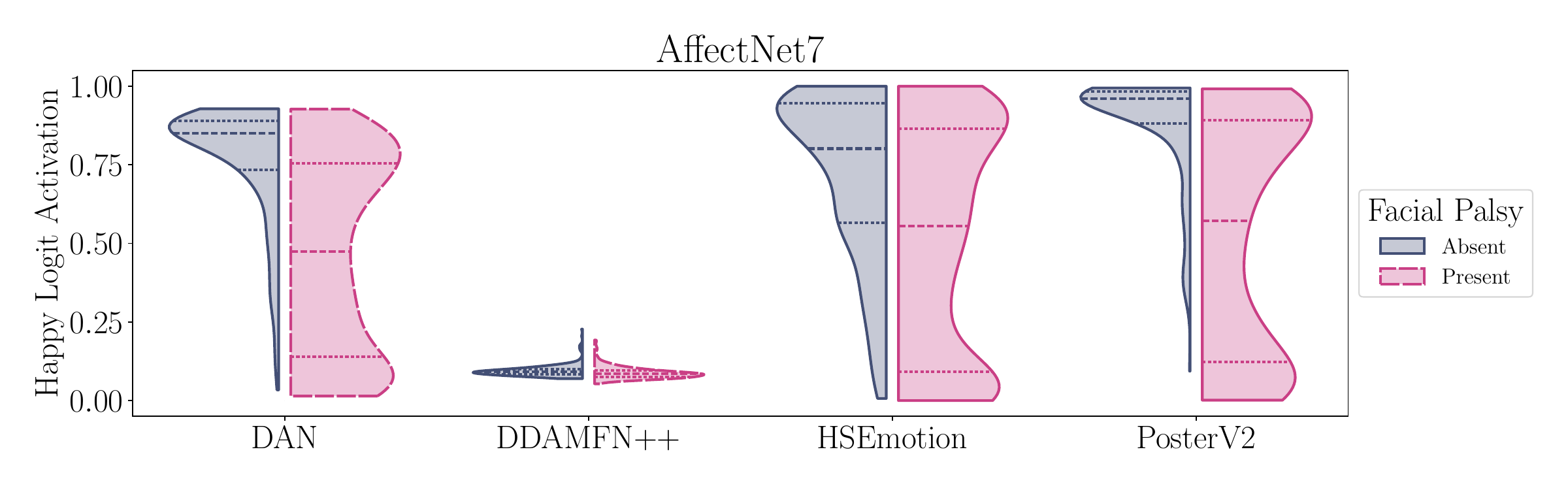}
        \caption{Shift in output behavior for classifiers trained on AffectNet7~\cite{Mollahosseini2019affectnet} with respect to facial palsy.}
        \label{fig:violin-affnet7}
    \end{subfigure}
    \begin{subfigure}{\textwidth}
    \centering
        \includegraphics[width=0.666\textwidth, trim=0cm 0cm 0cm 1.7cm, clip]{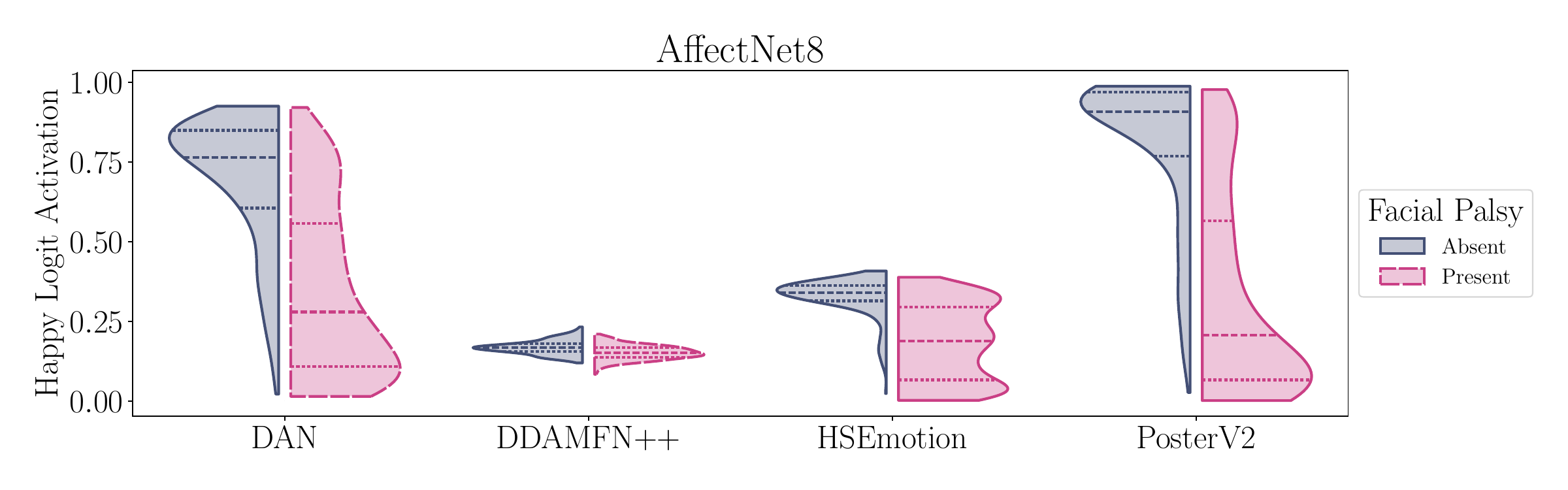}
        \caption{Shift in output behavior for classifiers trained on AffectNet8~\cite{Mollahosseini2019affectnet} with respect to facial palsy.}
        \label{fig:violin-affnet8}
    \end{subfigure}
    \begin{subfigure}{\textwidth}
    \centering
        \includegraphics[width=\textwidth, trim=0cm 0cm 0cm 1.7cm, clip]{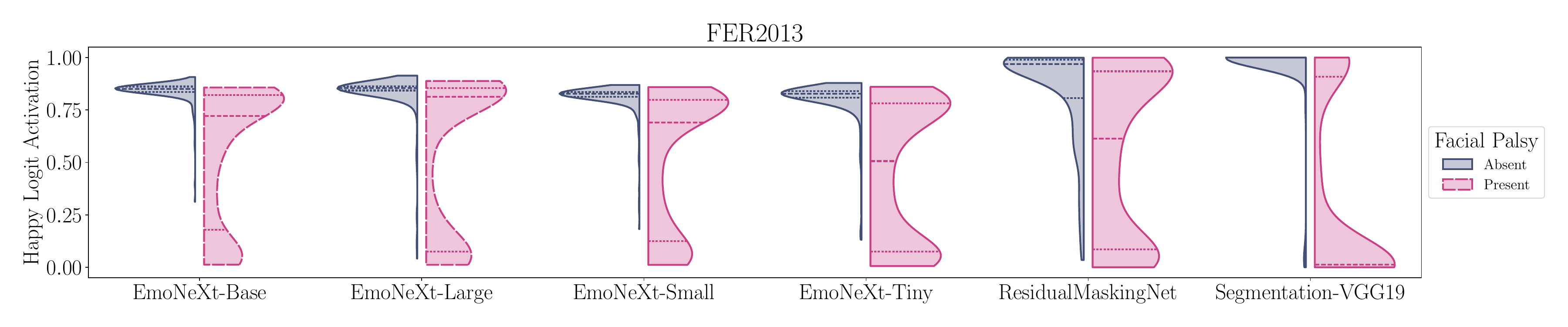}
        \caption{Shift in output behavior for classifiers trained on FER2013~\cite{dumitru2013fer} with respect to facial palsy.}
        \label{fig:violin-fer2013}
    \end{subfigure}
    \begin{subfigure}{\textwidth}
    \centering
        \includegraphics[width=0.5\textwidth, trim=0cm 0cm 0cm 1.7cm, clip]{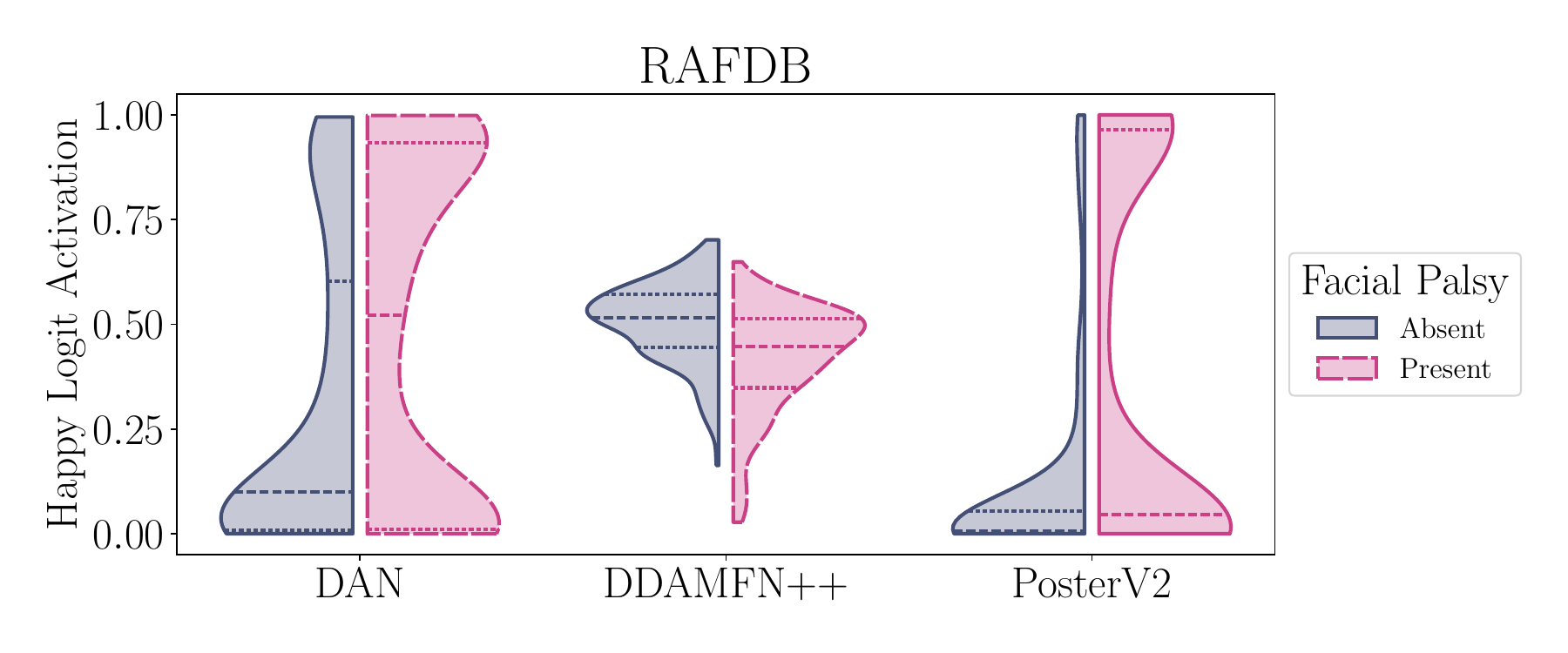}
        \caption{Shift in output behavior for classifiers trained on RAFDB~\cite{li2017reliable,li2019reliable} with respect to facial palsy. Note that we find the behavior shift for the DAN~\cite{wenDistractYourAttention2023} model is not significant.}
        \label{fig:violin-rafdb}
    \end{subfigure}
    \caption{We follow \cite{buechner2024power} and visualize the differences in the classifiers \happy logit distribution for healthy probands and facial palsy patients.
    Here \ref{fig:violin-affnet7} - \ref{fig:violin-rafdb} contain models trained on the indicated dataset respectively.
    }
    \label{fig:violins}
\end{figure}

Following previous work \cite{buechner2024power}, we visualize the difference in the \happy logit behavior between the healthy probands and facial palsy patients.
\Cref{fig:violins} contains these results split between the training datasets of the 17 models we investigate in this work.
However, these are associational investigations, i.e., of the first level of the PCH \cite{bareinboim2022onpearl}.
In other words, we do not isolate changes in facial symmetry from confounding factors and it is highly likely that other features correlate with the presence of facial palsy.
Hence, while we observe changes in classifier behavior on real data, our interventional investigation is more reliable and provides actionable insights.

Nevertheless, \Cref{fig:violins} shows a decrease in \happy activations for most models.
This is congruent with the aggregated performance results in \Cref{tab:accuracy_probands_patients}.
Further, these results are in line with our insights gained using our interventional framework: asymmetry results in lower activations for the \happy class.
Interestingly, we observe a slight deviation for models trained on the RAFDB.
Here DAN \cite{wenDistractYourAttention2023}, and PosterV2 \cite{maoPOSTERSimplerStronger2023} show higher activations and improved performance.
Nonetheless, both models still struggle with facial palsy patients and are outperformed by DDAMMFN++ \cite{zhangDualDirectionAttentionMixed2023} trained on the same dataset.

\begin{figure}
    \centering
    \begin{subfigure}{\textwidth}
    \centering
        \includegraphics[width=\textwidth, trim=0cm 0.8cm 0cm 1.7cm, clip]{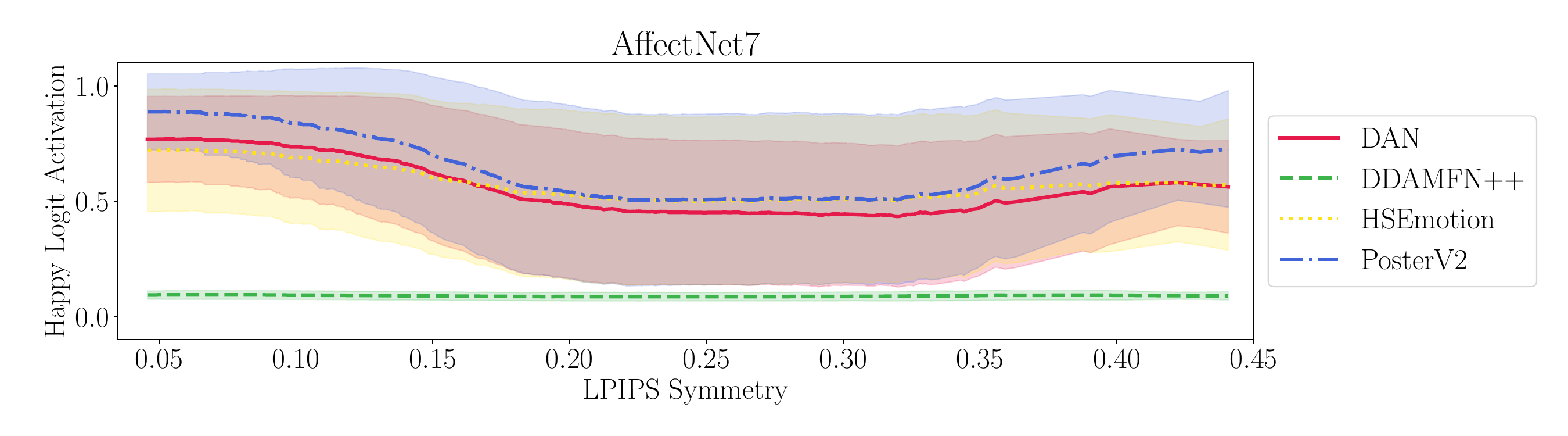}
        \caption{Shift in output for classifiers trained on AffectNet7~\cite{Mollahosseini2019affectnet} with respect to LPIPS \cite{zhang2018unreasonable} symmetry.}
        \label{fig:lpips-affnet7}
    \end{subfigure}
    \begin{subfigure}{\textwidth}
    \centering
        \includegraphics[width=\textwidth, trim=0cm 0.8cm 0cm 1.7cm, clip]{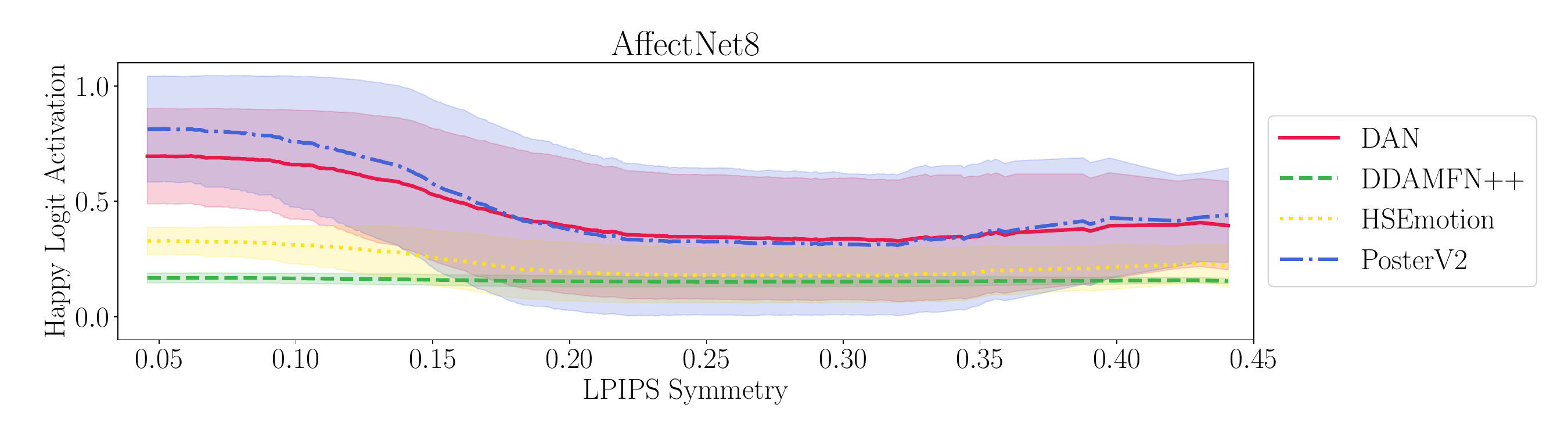}
        \caption{Shift in output for classifiers trained on AffectNet8~\cite{Mollahosseini2019affectnet} with respect to LPIPS \cite{zhang2018unreasonable} symmetry.}
        \label{fig:lpips-affnet8}
    \end{subfigure}
    \begin{subfigure}{\textwidth}
    \centering
        \includegraphics[width=\textwidth, trim=0cm 0.8cm 0cm 1.7cm, clip]{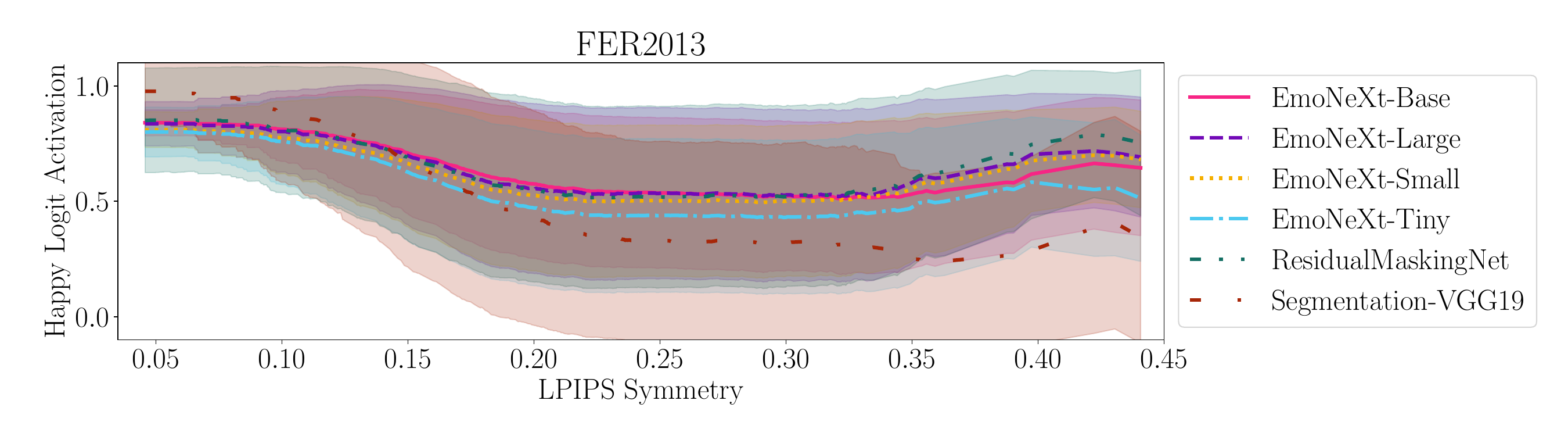}
        \caption{Shift in output for classifiers trained on FER2013~\cite{dumitru2013fer} with respect to LPIPS \cite{zhang2018unreasonable} symmetry.}
        \label{fig:lpips-fer2013}
    \end{subfigure}
    \begin{subfigure}{\textwidth}
    \centering
        \includegraphics[width=\textwidth, trim=0cm 0.8cm 0cm 1.7cm, clip]{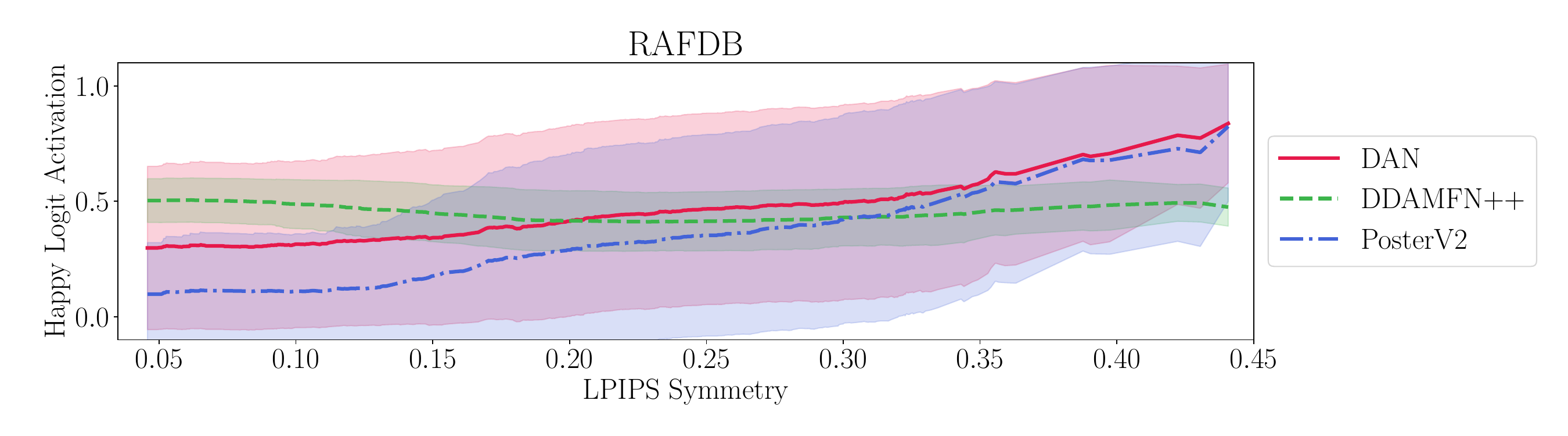}
        \caption{Shift in output for classifiers trained on RAFDB~\cite{li2017reliable,li2019reliable} with respect to LPIPS \cite{zhang2018unreasonable} symmetry. }
        \label{fig:lpips-rafdb}
    \end{subfigure}
    \caption{We follow \cite{penzel2023analyzing,buechner2024power} and regress the shift in the classifiers \happy logit distribution for measured LPIPS \cite{zhang2018unreasonable} symmetry scores of healthy probands and facial palsy patients.
    Here \ref{fig:lpips-affnet7} - \ref{fig:lpips-rafdb} contain models trained on the indicated dataset respectively.
    Note that higher LPIPS corresponds to lower symmetry \cite{zhang2018unreasonable}.
    }
    \label{fig:lpips-regression}
\end{figure}

Additionally, we also visualize the results for the continuous LPIPS \cite{zhang2018unreasonable} symmetry.
For regressing the mean and standard deviation, we use a window regression approach as described in \cite{penzel2023analyzing}.
We display these visualizations in \Cref{fig:lpips-regression}.

Overall, we observe for most models a decrease in logit activations for decreasing facial symmetry.
Hence, these results are congruent with the findings made in the main paper and \Cref{fig:violins}.
Furthermore, we again observe a very small effect size for DDAMFS++ \cite{zhangDualDirectionAttentionMixed2023} for both features.
These findings are in agreement with the noted performance in \Cref{tab:accuracy_probands_patients}.

Nevertheless, we want to highlight two additional observations:
First, in \Cref{fig:lpips-rafdb}, we observe an unexpected increase in activations.
While these are associational insights, i.e., there are many possible reasons, these increases are also visible in \Cref{fig:violin-rafdb}.
Second, while for most models in \Cref{fig:lpips-regression}, we observe a decrease in logit activations for lower facial symmetry, we note a smaller increase again for the most asymmetric faces.

\newpage
\section{Additional Details Experiment 2}

This section details the behavior analysis of the 17 expression classifiers on our synthetic intervention data.
We start with setup and information about the facial expression optimization before displaying the sampled individuals.
Afterward, we display the facial expressions achieved during the optimization per classifier for an individual.
Finally, we visualize the resulting activation surfaces.

\begin{table}[t]
    \centering
    \caption{
        Using our intervention framework, we optimized each expression classifier $\eclass_\ecweights$ using logit activation for each of the six base emotions.
        We display the average logit activation per model and emotion.
        Low activation is displayed in a \colorbox{lowacc}{\color[HTML]{F1F1F1} darks shade}, and high activation is displayed in a \colorbox{highacc}{\color[HTML]{000000} light shade}.
        We observe that \emph{fear} has a generally low activation, indicating that the models have issues classifying fear or that the FLAME expression cannot model fear. 
    }
    \begin{tabular}{llrrrrrr}
\toprule
Dataset & Model & \multicolumn{1}{c}{Angry} & \multicolumn{1}{c}{Disgust} & \multicolumn{1}{c}{Fear} & \multicolumn{1}{c}{Happy} & \multicolumn{1}{c}{Sad} & \multicolumn{1}{c}{Surprise} \\
\midrule
\multirow[c]{4}{*}{AffectNet7} & DAN~\cite{wenDistractYourAttention2023} & {\cellcolor[HTML]{A2DA37}} \color[HTML]{000000} 0.862 & {\cellcolor[HTML]{9DD93B}} \color[HTML]{000000} 0.853 & {\cellcolor[HTML]{25838E}} \color[HTML]{F1F1F1} 0.446 & {\cellcolor[HTML]{95D840}} \color[HTML]{000000} 0.842 & {\cellcolor[HTML]{44BF70}} \color[HTML]{F1F1F1} 0.702 & {\cellcolor[HTML]{C8E020}} \color[HTML]{000000} 0.917 \\
 & DDAMFN++~\cite{zhangDualDirectionAttentionMixed2023} & {\cellcolor[HTML]{2E6E8E}} \color[HTML]{F1F1F1} 0.356 & {\cellcolor[HTML]{31678E}} \color[HTML]{F1F1F1} 0.331 & {\cellcolor[HTML]{472F7D}} \color[HTML]{F1F1F1} 0.136 & {\cellcolor[HTML]{3E4989}} \color[HTML]{F1F1F1} 0.220 & {\cellcolor[HTML]{3D4D8A}} \color[HTML]{F1F1F1} 0.233 & {\cellcolor[HTML]{365D8D}} \color[HTML]{F1F1F1} 0.292 \\
 & HSEmotion~\cite{savchenko2023facial} & {\cellcolor[HTML]{C8E020}} \color[HTML]{000000} 0.915 & {\cellcolor[HTML]{C5E021}} \color[HTML]{000000} 0.913 & {\cellcolor[HTML]{29798E}} \color[HTML]{F1F1F1} 0.403 & {\cellcolor[HTML]{F1E51D}} \color[HTML]{000000} 0.979 & {\cellcolor[HTML]{89D548}} \color[HTML]{000000} 0.824 & {\cellcolor[HTML]{E2E418}} \color[HTML]{000000} 0.954 \\
 & PosterV2~\cite{maoPOSTERSimplerStronger2023} & {\cellcolor[HTML]{90D743}} \color[HTML]{000000} 0.835 & {\cellcolor[HTML]{D2E21B}} \color[HTML]{000000} 0.931 & {\cellcolor[HTML]{20928C}} \color[HTML]{F1F1F1} 0.505 & {\cellcolor[HTML]{DFE318}} \color[HTML]{000000} 0.950 & {\cellcolor[HTML]{5CC863}} \color[HTML]{000000} 0.747 & {\cellcolor[HTML]{D2E21B}} \color[HTML]{000000} 0.931 \\
 \midrule
\multirow[c]{4}{*}{AffectNet8} & DAN~\cite{wenDistractYourAttention2023} & {\cellcolor[HTML]{6CCD5A}} \color[HTML]{000000} 0.776 & {\cellcolor[HTML]{7FD34E}} \color[HTML]{000000} 0.805 & {\cellcolor[HTML]{287C8E}} \color[HTML]{F1F1F1} 0.416 & {\cellcolor[HTML]{24878E}} \color[HTML]{F1F1F1} 0.464 & {\cellcolor[HTML]{54C568}} \color[HTML]{000000} 0.732 & {\cellcolor[HTML]{B0DD2F}} \color[HTML]{000000} 0.881 \\
 & DDAMFN++~\cite{zhangDualDirectionAttentionMixed2023} & {\cellcolor[HTML]{3D4E8A}} \color[HTML]{F1F1F1} 0.237 & {\cellcolor[HTML]{404688}} \color[HTML]{F1F1F1} 0.211 & {\cellcolor[HTML]{472C7A}} \color[HTML]{F1F1F1} 0.122 & {\cellcolor[HTML]{3E4C8A}} \color[HTML]{F1F1F1} 0.228 & {\cellcolor[HTML]{3F4788}} \color[HTML]{F1F1F1} 0.212 & {\cellcolor[HTML]{33638D}} \color[HTML]{F1F1F1} 0.316 \\
 & HSEmotion~\cite{savchenko2023facial} & {\cellcolor[HTML]{22A785}} \color[HTML]{F1F1F1} 0.595 & {\cellcolor[HTML]{63CB5F}} \color[HTML]{000000} 0.759 & {\cellcolor[HTML]{2F6C8E}} \color[HTML]{F1F1F1} 0.349 & {\cellcolor[HTML]{306A8E}} \color[HTML]{F1F1F1} 0.340 & {\cellcolor[HTML]{21A685}} \color[HTML]{F1F1F1} 0.590 & {\cellcolor[HTML]{8BD646}} \color[HTML]{000000} 0.826 \\
 & PosterV2~\cite{maoPOSTERSimplerStronger2023} & {\cellcolor[HTML]{84D44B}} \color[HTML]{000000} 0.814 & {\cellcolor[HTML]{C5E021}} \color[HTML]{000000} 0.911 & {\cellcolor[HTML]{21908D}} \color[HTML]{F1F1F1} 0.499 & {\cellcolor[HTML]{35B779}} \color[HTML]{F1F1F1} 0.666 & {\cellcolor[HTML]{50C46A}} \color[HTML]{000000} 0.725 & {\cellcolor[HTML]{D0E11C}} \color[HTML]{000000} 0.928 \\
 \midrule
\multirow[c]{6}{*}{FER2013} 
 & EmoNeXt-Tiny$^{\dagger}$~\cite{boundori2023EmoNext} & {\cellcolor[HTML]{2A788E}} \color[HTML]{F1F1F1} 0.400 & {\cellcolor[HTML]{481F70}} \color[HTML]{F1F1F1} 0.083 & {\cellcolor[HTML]{39568C}} \color[HTML]{F1F1F1} 0.269 & {\cellcolor[HTML]{20928C}} \color[HTML]{F1F1F1} 0.508 & {\cellcolor[HTML]{375A8C}} \color[HTML]{F1F1F1} 0.278 & {\cellcolor[HTML]{89D548}} \color[HTML]{000000} 0.821 \\
 & EmoNeXt-Small$^{\dagger}$~\cite{boundori2023EmoNext} & {\cellcolor[HTML]{52C569}} \color[HTML]{000000} 0.727 & {\cellcolor[HTML]{482071}} \color[HTML]{F1F1F1} 0.088 & {\cellcolor[HTML]{306A8E}} \color[HTML]{F1F1F1} 0.342 & {\cellcolor[HTML]{5EC962}} \color[HTML]{000000} 0.752 & {\cellcolor[HTML]{2F6B8E}} \color[HTML]{F1F1F1} 0.345 & {\cellcolor[HTML]{B2DD2D}} \color[HTML]{000000} 0.885 \\
& EmoNeXt-Base$^{\dagger}$~\cite{boundori2023EmoNext}  & {\cellcolor[HTML]{1E9C89}} \color[HTML]{F1F1F1} 0.548 & {\cellcolor[HTML]{481C6E}} \color[HTML]{F1F1F1} 0.076 & {\cellcolor[HTML]{27808E}} \color[HTML]{F1F1F1} 0.432 & {\cellcolor[HTML]{4EC36B}} \color[HTML]{000000} 0.720 & {\cellcolor[HTML]{23888E}} \color[HTML]{F1F1F1} 0.465 & {\cellcolor[HTML]{B2DD2D}} \color[HTML]{000000} 0.884 \\
 & EmoNeXt-Large$^{\dagger}$~\cite{boundori2023EmoNext}  & {\cellcolor[HTML]{BDDF26}} \color[HTML]{000000} 0.902 & {\cellcolor[HTML]{2E6D8E}} \color[HTML]{F1F1F1} 0.352 & {\cellcolor[HTML]{2DB27D}} \color[HTML]{F1F1F1} 0.644 & {\cellcolor[HTML]{B2DD2D}} \color[HTML]{000000} 0.886 & {\cellcolor[HTML]{8ED645}} \color[HTML]{000000} 0.829 & {\cellcolor[HTML]{A2DA37}} \color[HTML]{000000} 0.862 \\
 & ResidualMaskingNet~\cite{pham2021facial} & {\cellcolor[HTML]{E5E419}} \color[HTML]{000000} 0.959 & {\cellcolor[HTML]{FBE723}} \color[HTML]{000000} 0.995 & {\cellcolor[HTML]{21918C}} \color[HTML]{F1F1F1} 0.500 & {\cellcolor[HTML]{B2DD2D}} \color[HTML]{000000} 0.884 & {\cellcolor[HTML]{20A386}} \color[HTML]{F1F1F1} 0.581 & {\cellcolor[HTML]{FDE725}} \color[HTML]{000000} 0.997 \\
 & Segmentation-VGG19$^{\dagger}$~\cite{vigneshNovelFacialEmotion2023} & {\cellcolor[HTML]{86D549}} \color[HTML]{000000} 0.818 & {\cellcolor[HTML]{481A6C}} \color[HTML]{F1F1F1} 0.067 & {\cellcolor[HTML]{50C46A}} \color[HTML]{000000} 0.725 & {\cellcolor[HTML]{EFE51C}} \color[HTML]{000000} 0.976 & {\cellcolor[HTML]{98D83E}} \color[HTML]{000000} 0.846 & {\cellcolor[HTML]{CAE11F}} \color[HTML]{000000} 0.919 \\
 \midrule
\multirow[c]{3}{*}{RAFDB} & DAN~\cite{wenDistractYourAttention2023} & {\cellcolor[HTML]{F8E621}} \color[HTML]{000000} 0.991 & {\cellcolor[HTML]{ADDC30}} \color[HTML]{000000} 0.877 & {\cellcolor[HTML]{482071}} \color[HTML]{F1F1F1} 0.088 & {\cellcolor[HTML]{AADC32}} \color[HTML]{000000} 0.874 & {\cellcolor[HTML]{DDE318}} \color[HTML]{000000} 0.947 & {\cellcolor[HTML]{FDE725}} \color[HTML]{000000} 0.999 \\
 & DDAMFN++~\cite{zhangDualDirectionAttentionMixed2023} & {\cellcolor[HTML]{481C6E}} \color[HTML]{F1F1F1} 0.077 & {\cellcolor[HTML]{481A6C}} \color[HTML]{F1F1F1} 0.070 & {\cellcolor[HTML]{450559}} \color[HTML]{F1F1F1} 0.013 & {\cellcolor[HTML]{20A486}} \color[HTML]{F1F1F1} 0.585 & {\cellcolor[HTML]{7FD34E}} \color[HTML]{000000} 0.808 & {\cellcolor[HTML]{69CD5B}} \color[HTML]{000000} 0.771 \\
 & PosterV2~\cite{maoPOSTERSimplerStronger2023} & {\cellcolor[HTML]{FBE723}} \color[HTML]{000000} 0.993 & {\cellcolor[HTML]{FBE723}} \color[HTML]{000000} 0.996 & {\cellcolor[HTML]{33628D}} \color[HTML]{F1F1F1} 0.312 & {\cellcolor[HTML]{F4E61E}} \color[HTML]{000000} 0.982 & {\cellcolor[HTML]{F6E620}} \color[HTML]{000000} 0.987 & {\cellcolor[HTML]{FDE725}} \color[HTML]{000000} 1.000 \\
 \bottomrule
\end{tabular}

    \label{tab:logit_activation}
\end{table}

\subsection{Classifier Facial Expression Optimization}
Our experiments optimized each classifier $\eclass_\ecweights$ regarding the six base emotions.
Therefore, we report the average logit activation per model and emotion reached in \Cref{tab:logit_activation}.
We can observe several interesting properties in the logit activation.
First, not all models can reach high logit activation based on facial expression changes.
This indicates that models also leverage other facial information while classifying facial expressions. 
Furthermore, we observe that fear has a low activation among all classifiers except SegmentationVgg19~\cite{vigneshNovelFacialEmotion2023}.
The \emph{surprise} facial expression has a high activation among all classifiers, whereas DDAMFN++\cite{zhangDualDirectionAttentionMixed2023} is the sole outlier; overall, reached activation is low. 

\newpage
\subsection{Individuals}
\begin{figure}[H]
    \centering
    \includegraphics[width=0.9\textwidth]{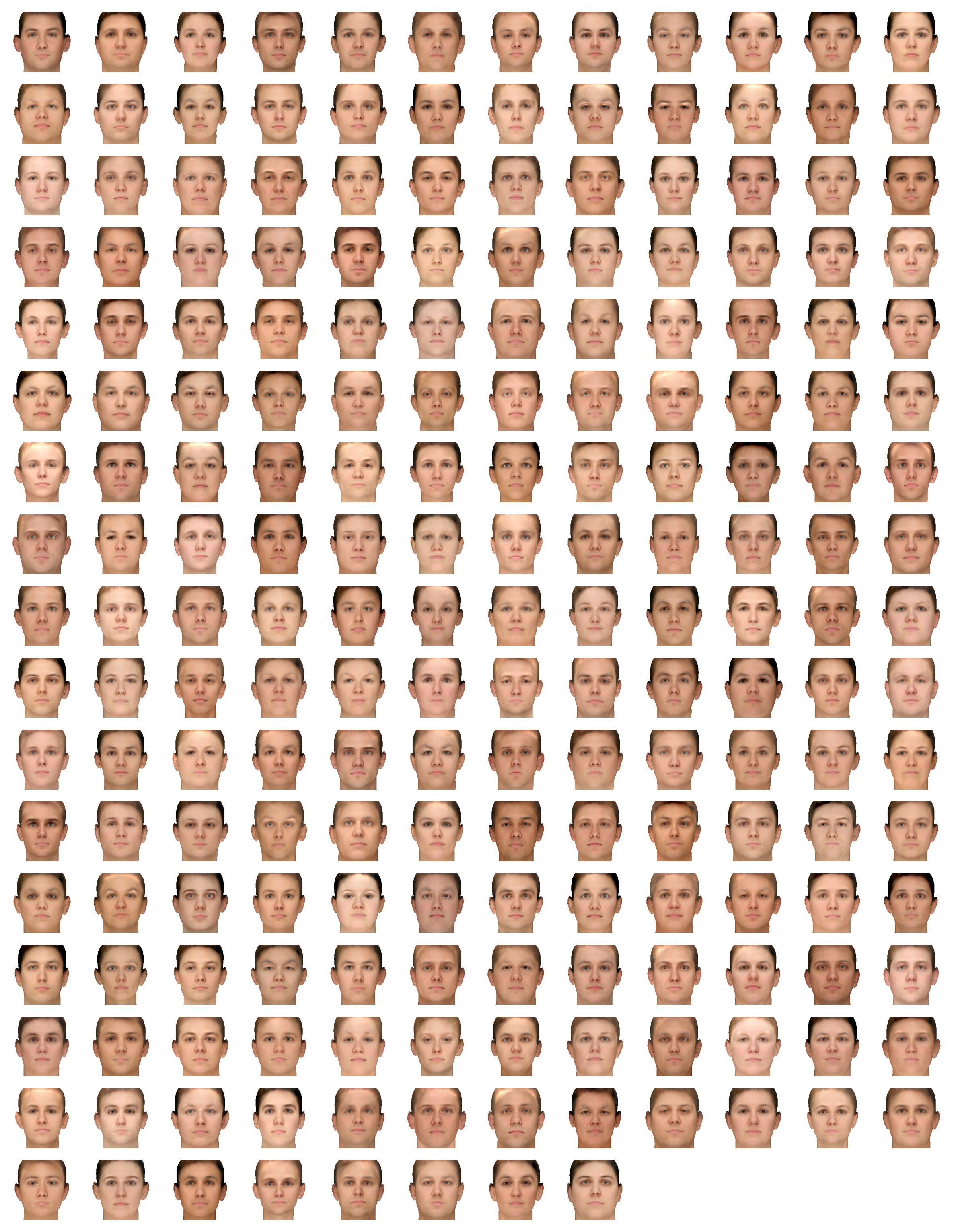}
    \caption{200 individual population $\mathfrak{I}$}
    \label{fig:population}
\end{figure}

We provide an overview of all created individuals in \Cref{fig:population}.
The data can be downloaded here:\url{https://doi.org/10.6084/m9.figshare.27074587.v1}.
All resemblance to existing people is not intended and could only result from the underlying FLAME geometry model~\cite{liLearningModelFacial2017} and the texture from the BaselFaceModel~\cite{paysan3DFaceModel2009}.

\newpage
\subsection{Average Facial Expression}
Together with the reached logit activations, see \Cref{tab:logit_activation}, we are interested in the resulting facial expression.
These should depict the internal representation of the respective emotion and give insight into what each classifier assumes.
Furthermore, we assume the underlying base dataset influences the expression.

\subsubsection{Average Facial Expression - Per Dataset}
\begin{figure}[H]
    \centering
    \includegraphics[width=0.9\textwidth]{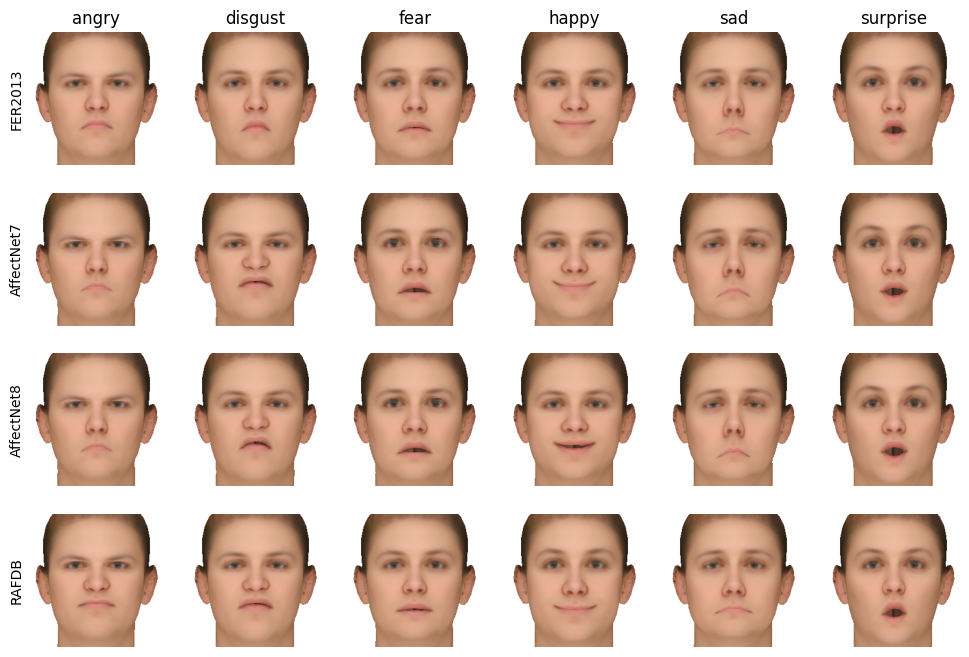}
    \caption{Average Facial Expression used for classification based on the underlying training dataset.}
    \label{fig:avg_per_dataset}
\end{figure}

Using our generative facial expression network, we can now create a representation of how different classifiers represent the underlying training dataset.
This means the expression vectors of all 200 individuals per dataset and model are averaged and shown in \Cref{fig:avg_per_dataset}.
This visualization gives an intuitive feeling about the underlying facial expression per FER benchmark~\cite{Mollahosseini2019affectnet,li2017reliable,li2019reliable,dumitru2013fer}.
Looking at the expression columns, we see that all interpretations of a face are slightly different.
For example, for {\color{angry}\emph{angry}}, the mouth frowning angles are different.
For \disgust the mouth is slightly opened compared to {\color{angry}\emph{angry}}.
For \fear we can clearly see that raising the eye brows is common.
The \happy expression varies either with a wider grin or the opening state of the eye.
The \sad expression varies in the intensity of the frowning, but the eyebrows are not activated by the corrugator muscle.
Also, the eyes are generally closed.
For the \surprise expression, we can see wide-open eyes and raised eyebrows in the shared interpretation.

Even though they are similar in their visual state, the intensity and expressiveness are different per model and could be the underlying reason for differences in the model architectures or the data used in the benchmark.

\subsubsection{Average Facial Expression - AffectNet7}
\begin{figure}[H]
    \centering
    \includegraphics[width=0.8\textwidth]{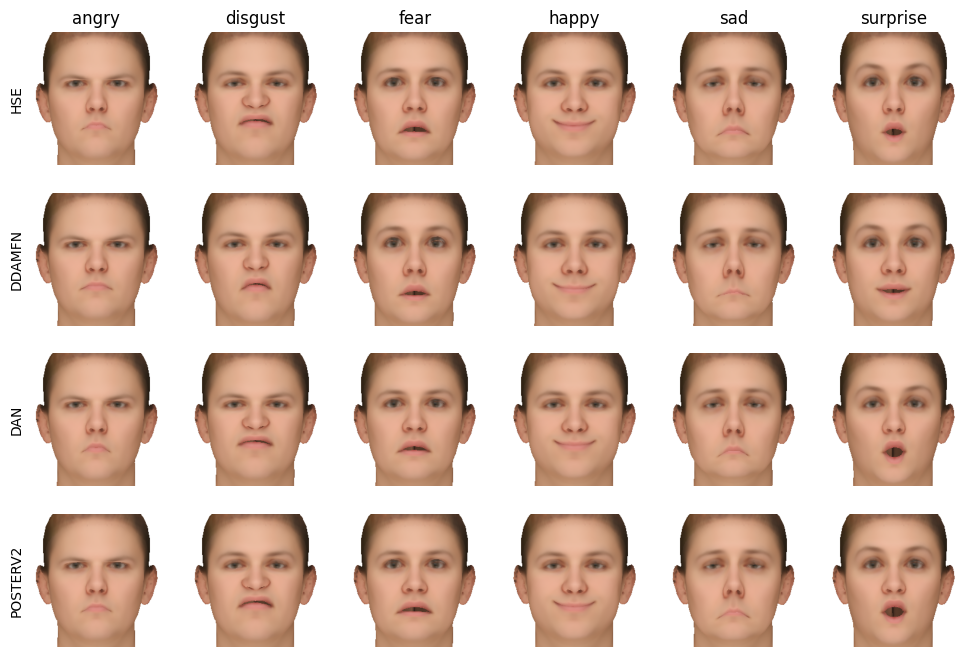}
    \caption{The average facial expressions for models trained on AffectNet7~\cite{Mollahosseini2019affectnet}}
    \label{fig:avg_affectnet7}
\end{figure}

\Cref{fig:avg_affectnet7}  contains the average facial expressions for models trained on AffectNet7~\cite{Mollahosseini2019affectnet}.

\subsubsection{Average Facial Expression - AffectNet8}
\begin{figure}[H]
    \centering
    \includegraphics[width=0.8\textwidth]{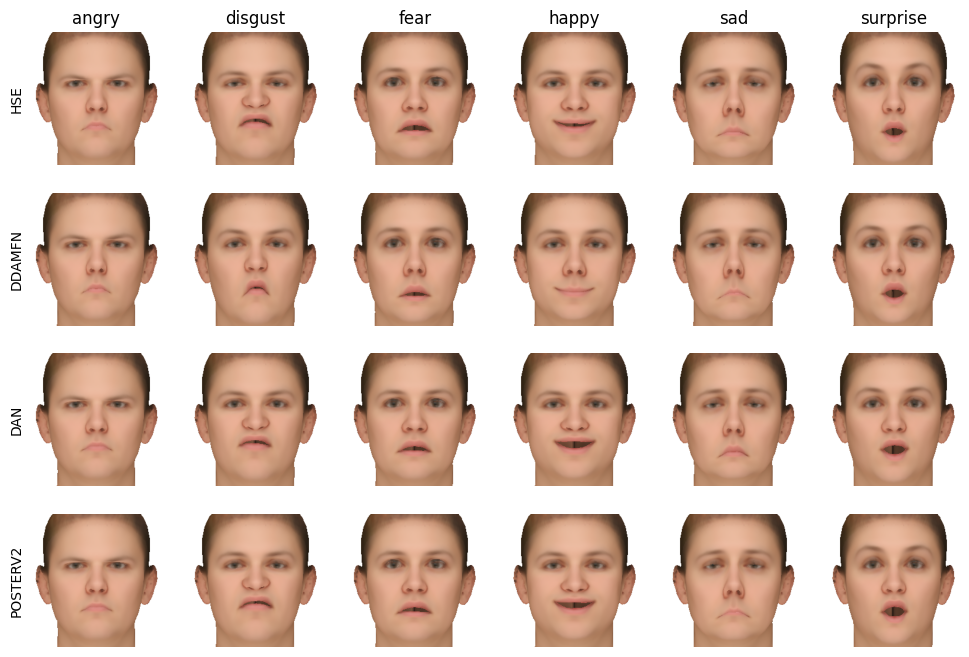}
    \caption{The average facial expressions for models trained on AffectNet8~\cite{Mollahosseini2019affectnet}}
    \label{fig:avg_affectnet8}
\end{figure}

\Cref{fig:avg_affectnet8}  contains the average facial expressions for models trained on AffectNet8~\cite{Mollahosseini2019affectnet}.

\subsubsection{Average Facial Expression - FER2013}
\begin{figure}[H]
    \centering
    \includegraphics[width=0.7\textwidth]{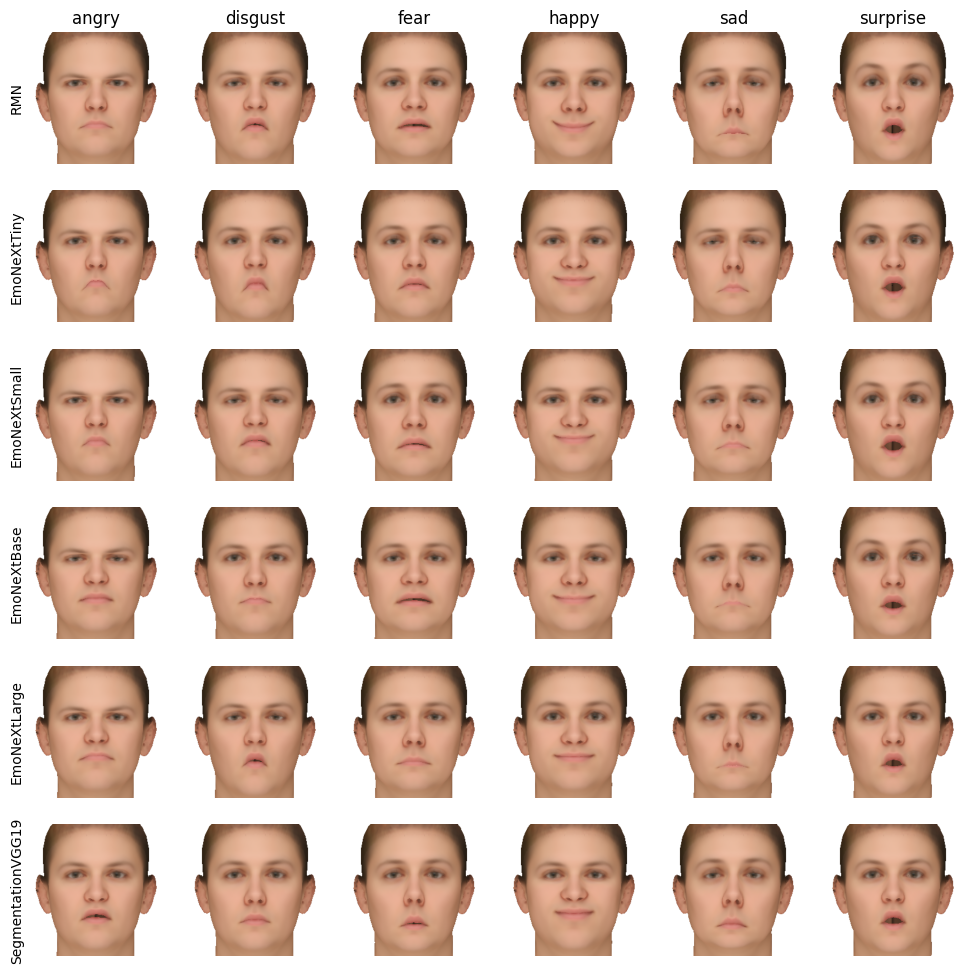}
    \caption{The average facial expressions for models trained on FER2013~\cite{dumitru2013fer}}
    \label{fig:avg_fer2013}
\end{figure}

\Cref{fig:avg_fer2013}  contains the average facial expressions for models trained on FER2013~\cite{dumitru2013fer}.

\subsubsection{Average Facial Expression - RAFDB}
\begin{figure}[H]
    \centering
    \includegraphics[width=0.75\textwidth]{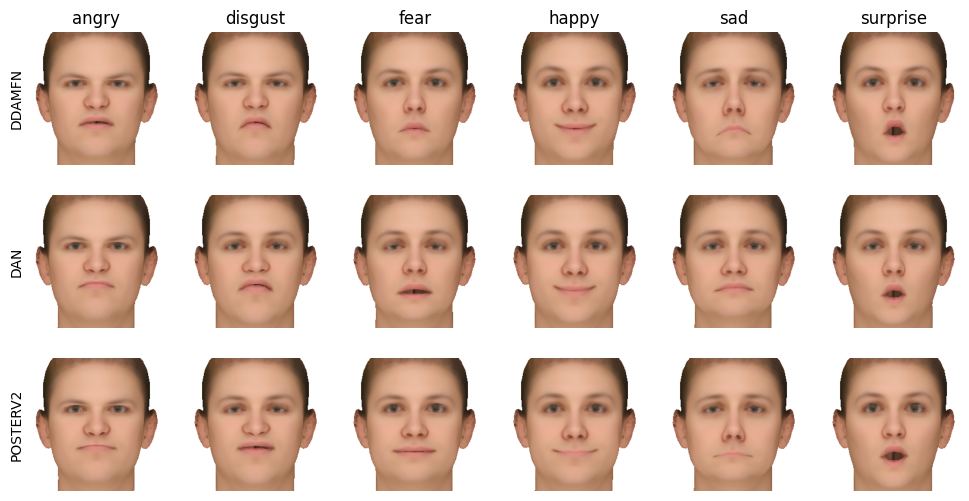}
    \caption{The average facial expressions for models trained on RAFDB~\cite{li2017reliable,li2019reliable}}
    \label{fig:avg_rafdb}
\end{figure}

\Cref{fig:avg_rafdb} contains the average facial expressions for models trained on RAFDB~\cite{li2017reliable,li2019reliable}.

\newpage
\subsection{Model Activation Surfaces}
The main paper shows that we use a finite grid over $\mathfrak{T}$ and $\mathfrak{S}$ to compute the FAIS score.
Given that we only highlighted the final time step $t=1.0$, we show here the full logit activation surfaces used to compute our score.

\begin{figure}[H]
    \centering
    \begin{subfigure}[b]{0.45\textwidth}
        \centering
        \includegraphics[width=\textwidth]{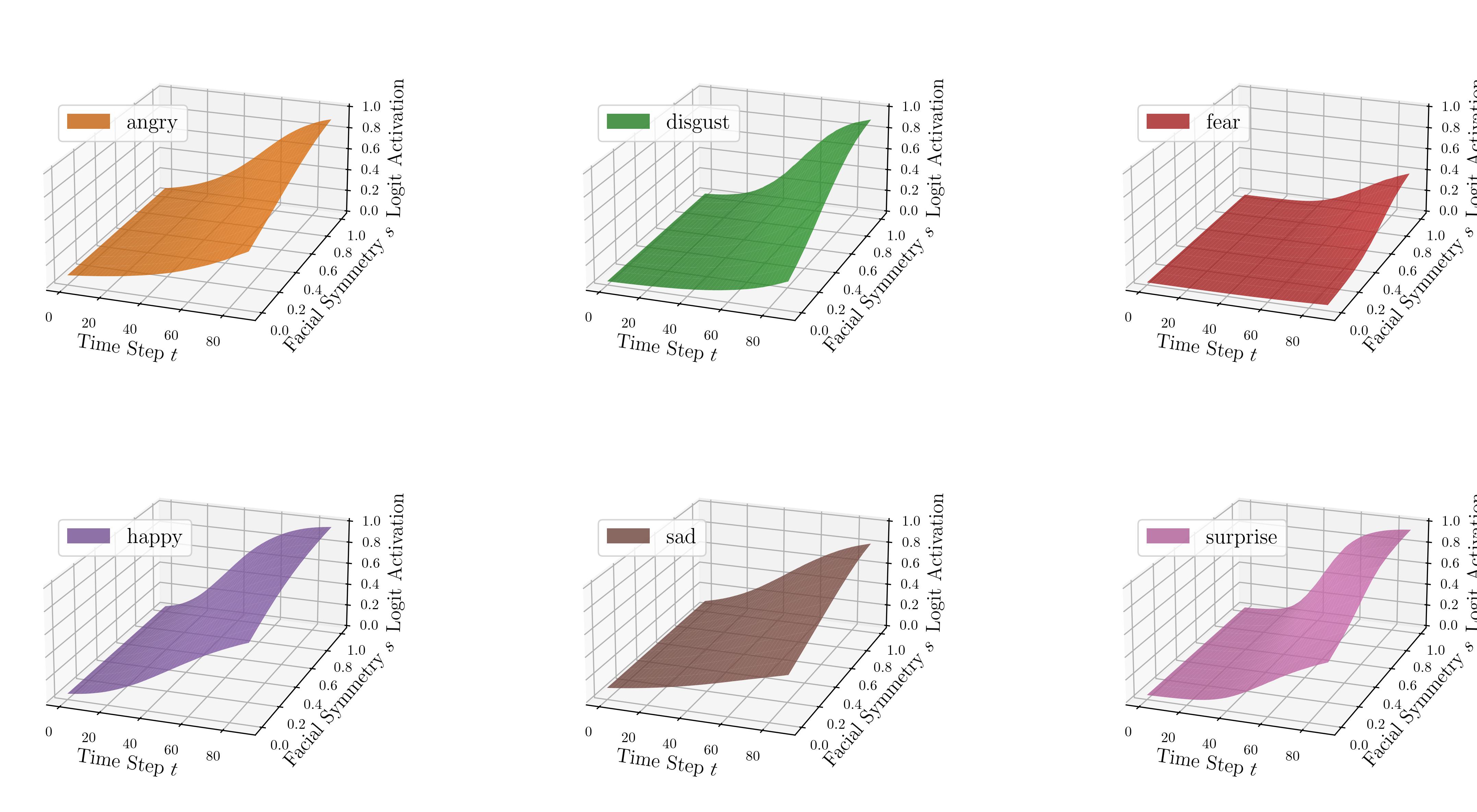}
        \caption{HSEmotion~\cite{savchenko2023facial}}
        \label{fig:sup_HSE7_surprise}
    \end{subfigure}
    \hfill
    \begin{subfigure}[b]{0.45\textwidth}
        \centering
        \includegraphics[width=\textwidth]{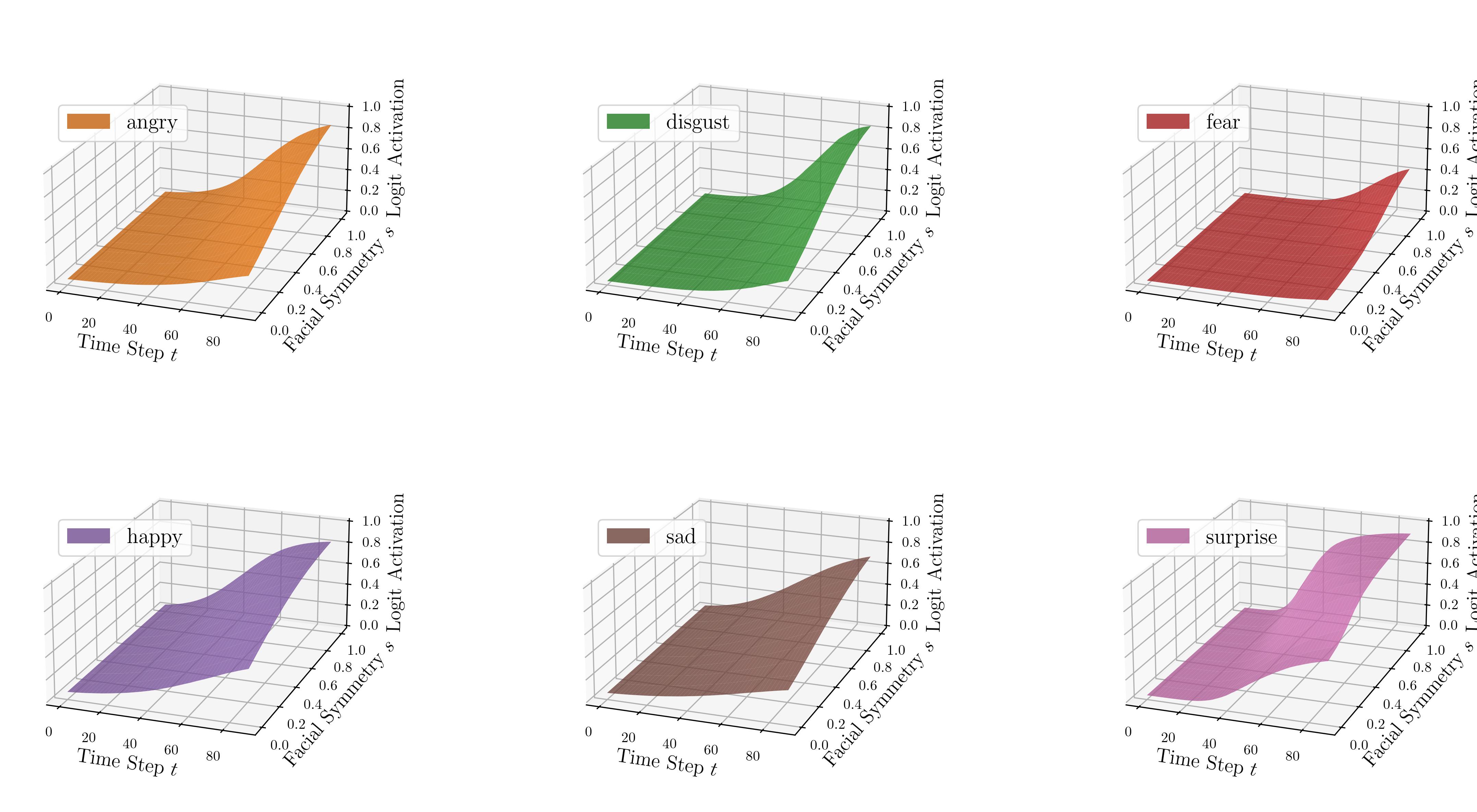}
        \caption{DAN~\cite{wenDistractYourAttention2023}}
        \label{fig:sup_DAN_aff7_surprise}
    \end{subfigure}
    \\
    \begin{subfigure}[b]{0.45\textwidth}
        \centering
        \includegraphics[width=\textwidth]{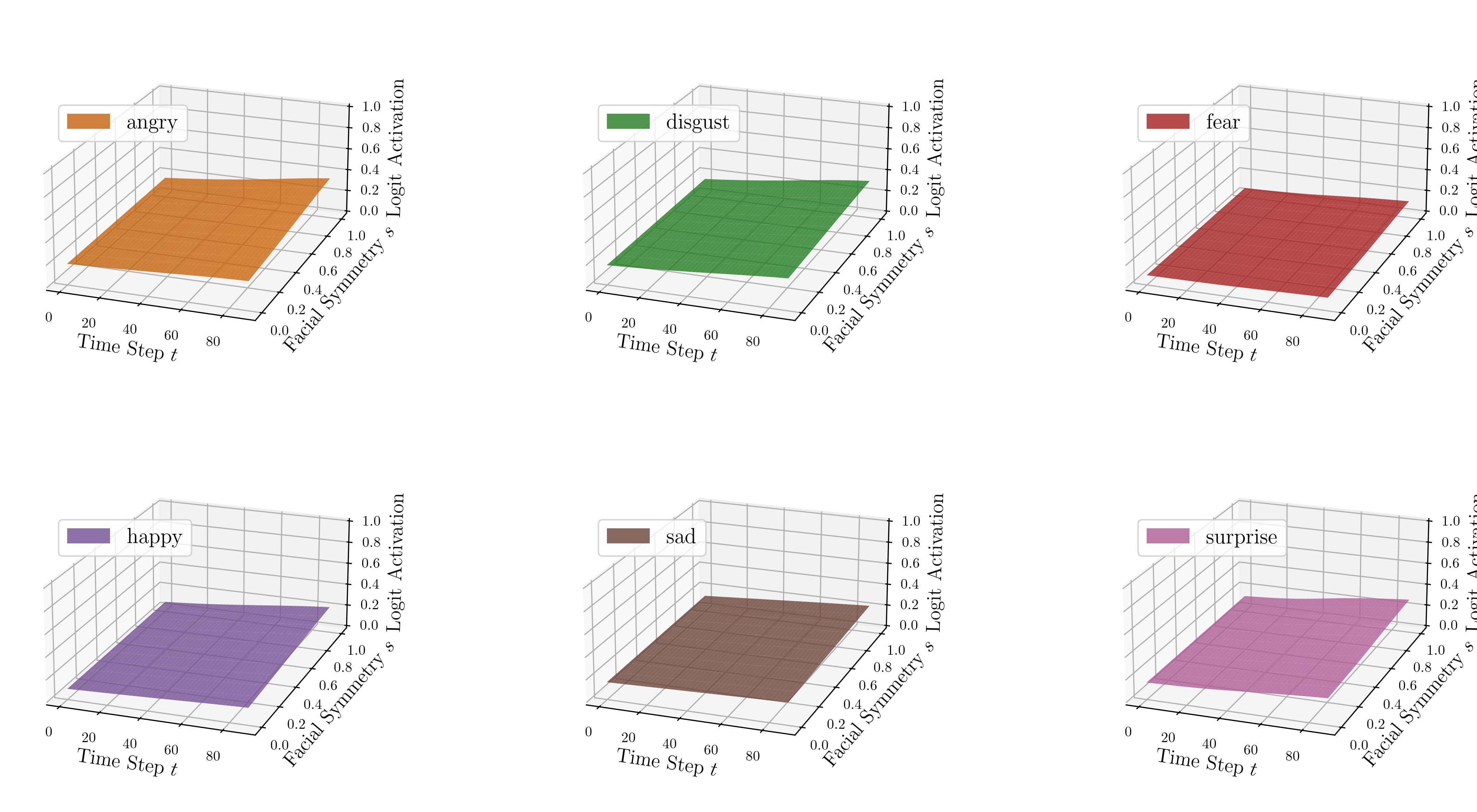}
        \caption{DDAMFN++~\cite{zhangDualDirectionAttentionMixed2023}}
        \label{fig:sup_ddamfn-aff7_surprise}
    \end{subfigure}
    \hfill
    \begin{subfigure}[b]{0.45\textwidth}
        \centering
        \includegraphics[width=\textwidth]{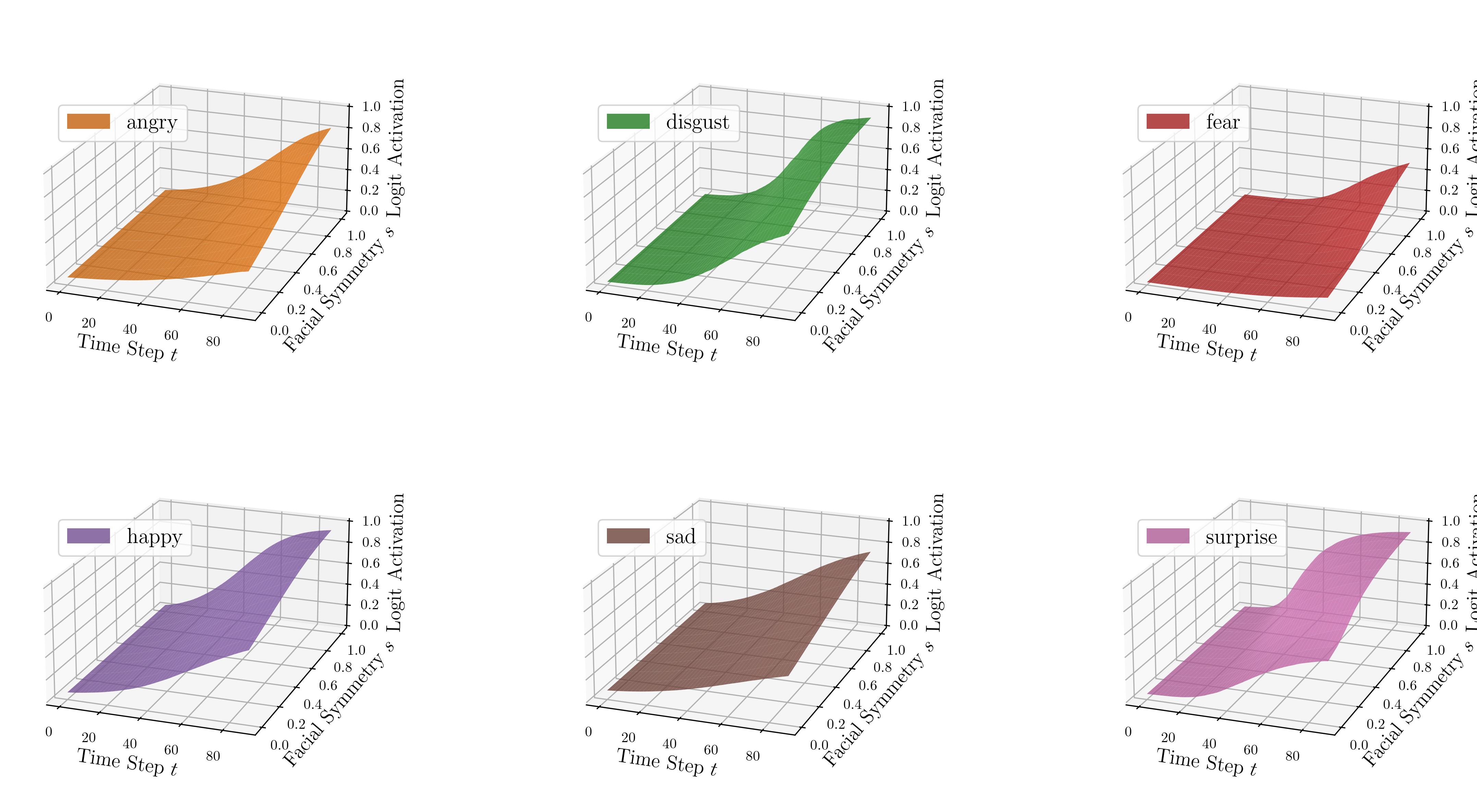}
        \caption{PosterV2~\cite{maoPOSTERSimplerStronger2023}}
        \label{fig:sup_posterv2-aff7_surprise}
    \end{subfigure}
    \caption{AffectNet7~\cite{Mollahosseini2019affectnet}}
    \label{fig:sup0}
\end{figure}

\begin{figure}[H]
    \centering
    \begin{subfigure}[b]{0.45\textwidth}
        \centering
        \includegraphics[width=\textwidth]{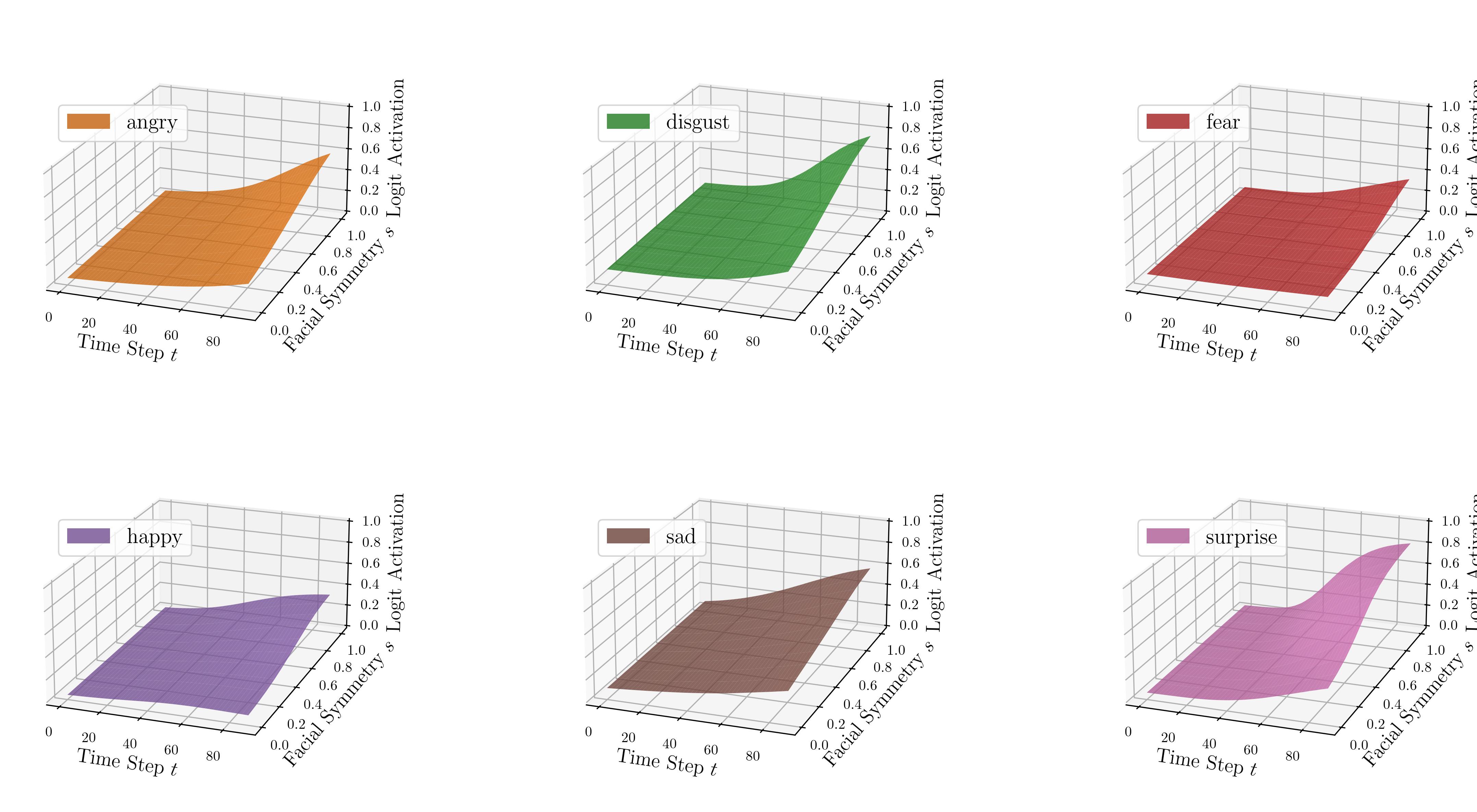}
        \caption{HSEmotion~\cite{savchenko2023facial}}
        \label{fig:sup_HSE8_surprise}
    \end{subfigure}
    \hfill
    \begin{subfigure}[b]{0.45\textwidth}
        \centering
        \includegraphics[width=\textwidth]{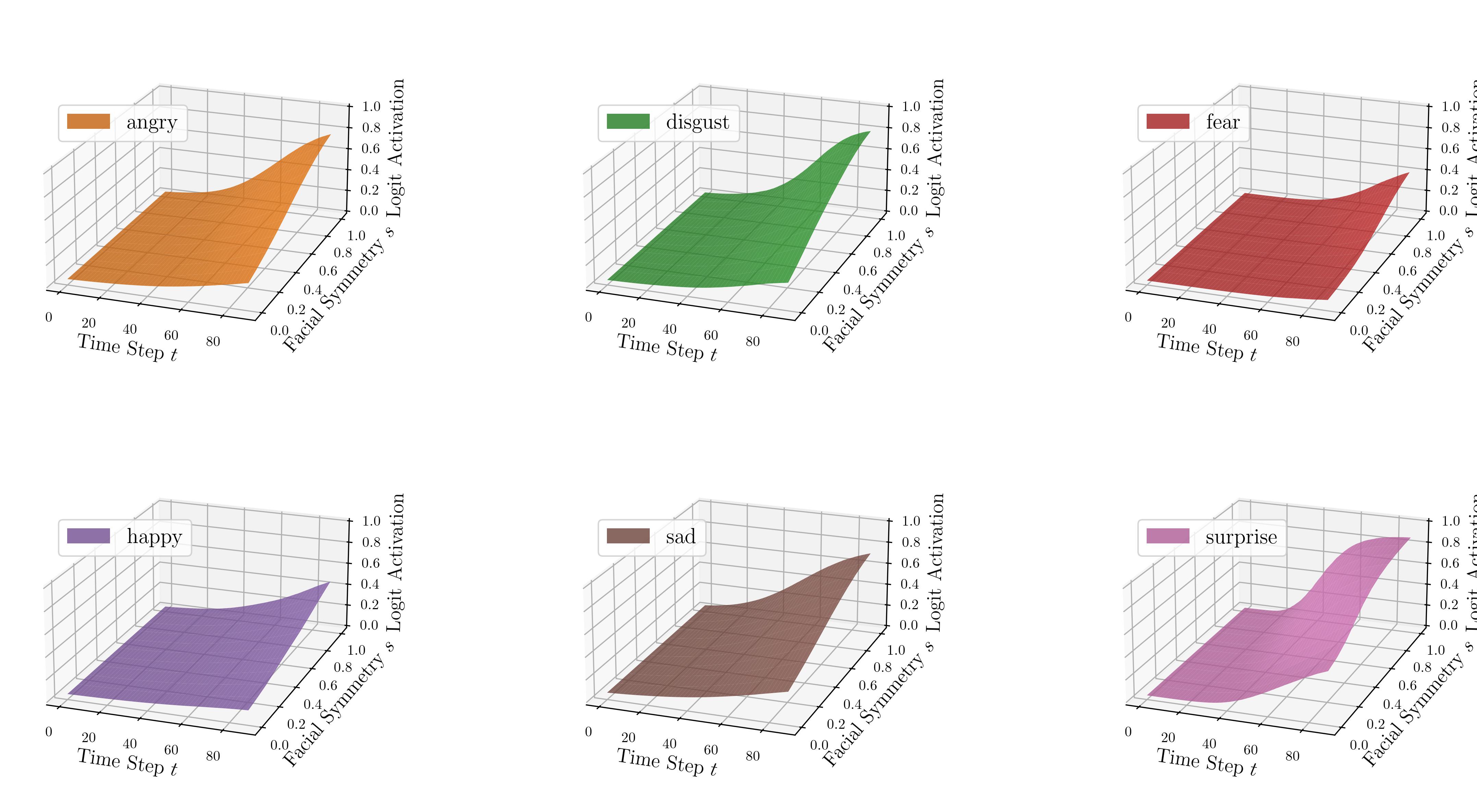}
        \caption{DAN~\cite{wenDistractYourAttention2023}}
        \label{fig:sup_dan_aff8}
    \end{subfigure}
    \\
    \begin{subfigure}[b]{0.45\textwidth}
        \centering
        \includegraphics[width=\textwidth]{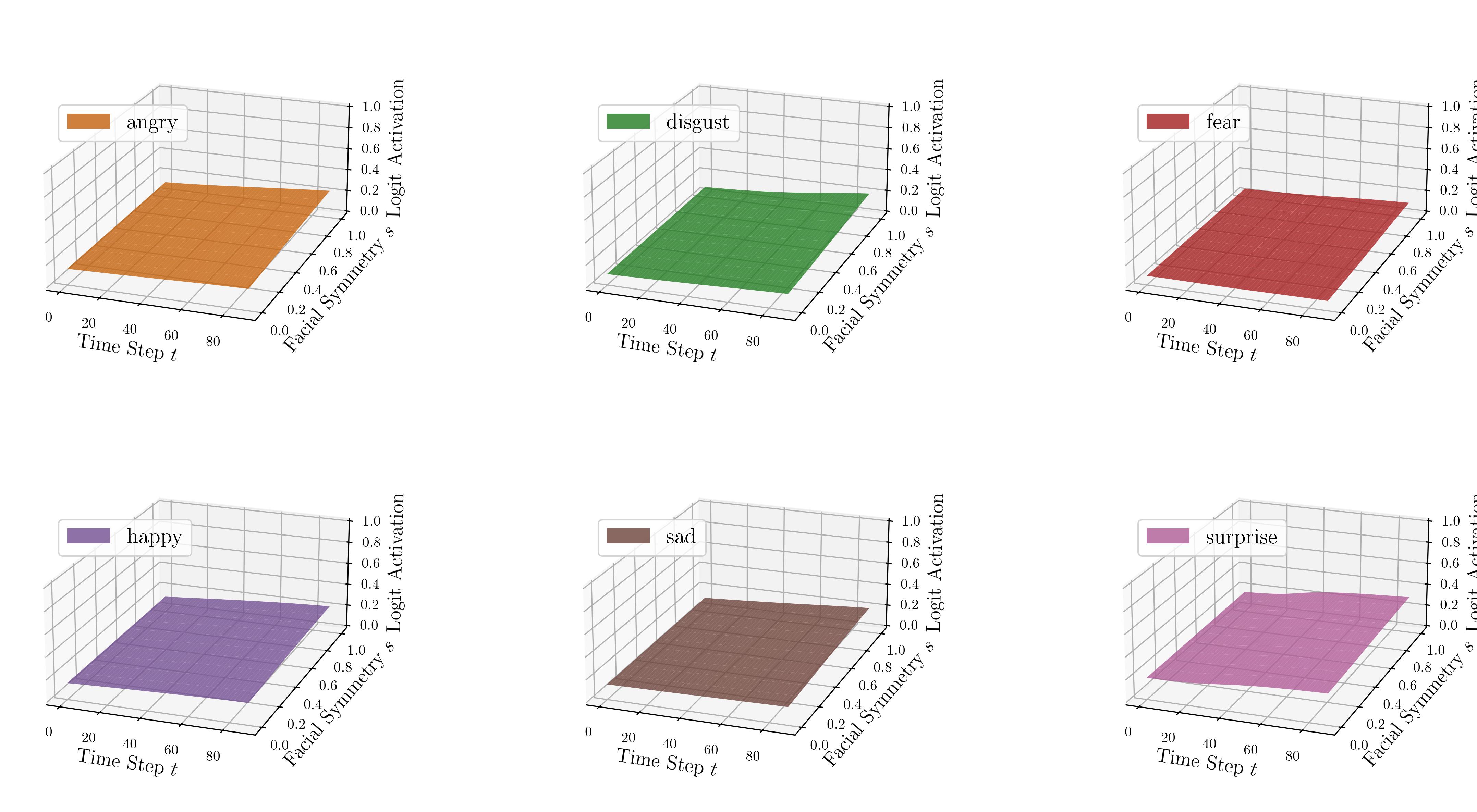}
        \caption{DDAMFN++~\cite{zhangDualDirectionAttentionMixed2023}}
        \label{fig:sup_ddafmn-aff8_surprise}
    \end{subfigure}
    \hfill
    \begin{subfigure}[b]{0.45\textwidth}
        \centering
        \includegraphics[width=\textwidth]{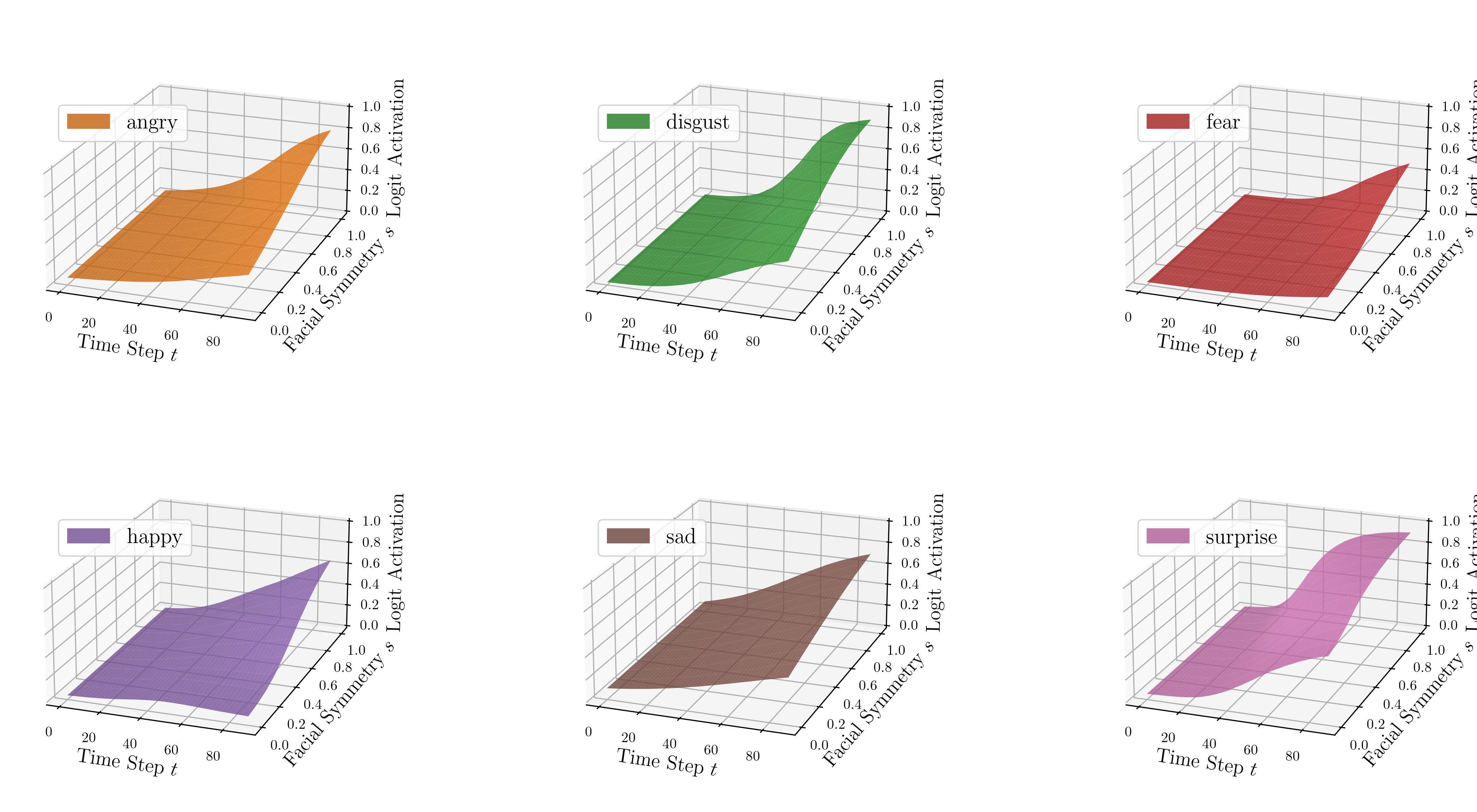}
        \caption{PosterV2~\cite{maoPOSTERSimplerStronger2023}}
        \label{fig:sup_posterv2-aff8_surprise}
    \end{subfigure}
    \caption{AffectNet8~\cite{Mollahosseini2019affectnet}}
    \label{fig:sup1}
\end{figure}

\begin{figure}[H]
    \centering
    \begin{subfigure}[b]{0.4\textwidth}
        \centering
        \includegraphics[width=\textwidth]{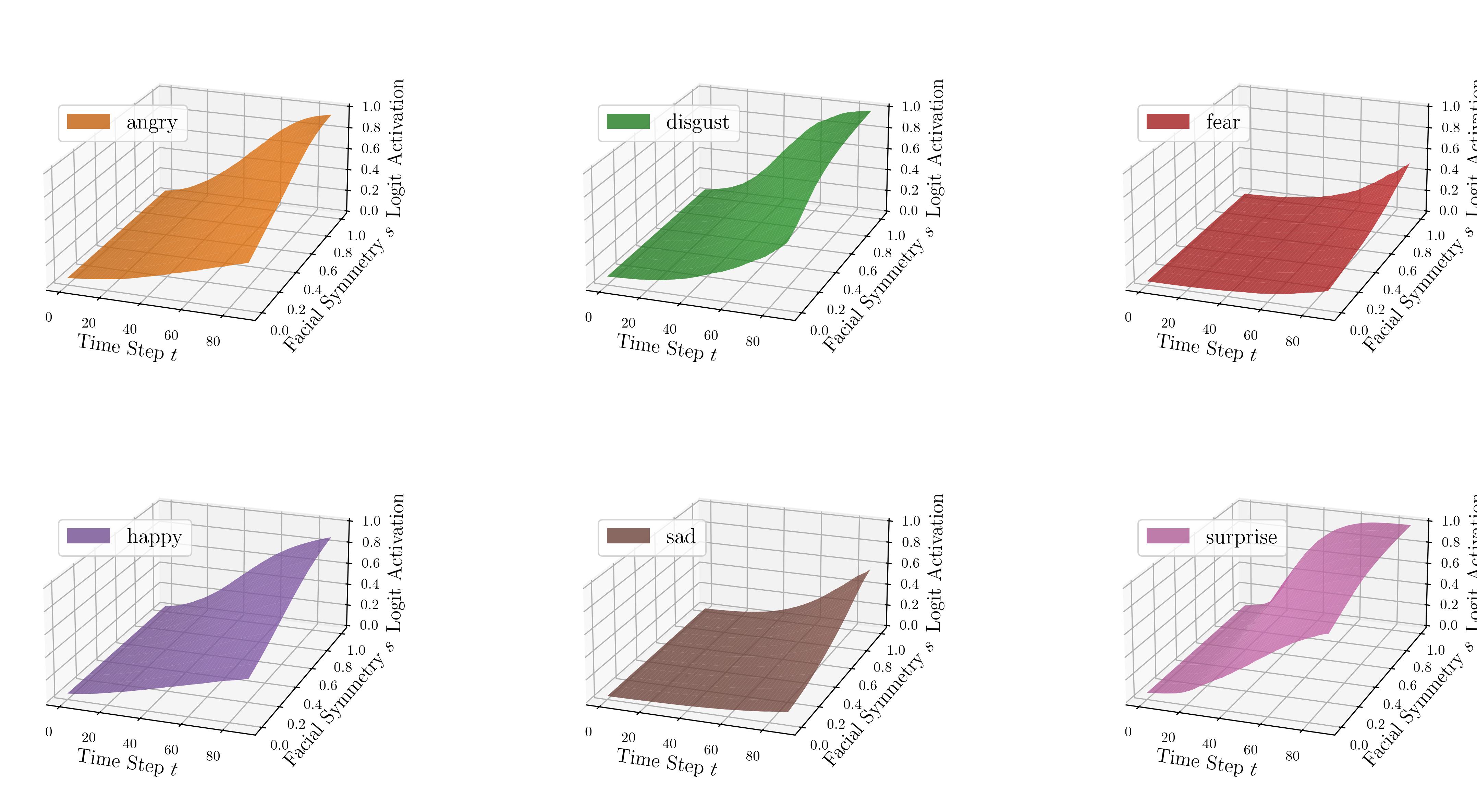}
        \caption{ResidualMaskingNet~\cite{pham2021facial}}
        \label{fig:sup_rmn_surprise}
    \end{subfigure}
    \hfill
    \begin{subfigure}[b]{0.4\textwidth}
        \centering
        \includegraphics[width=\textwidth]{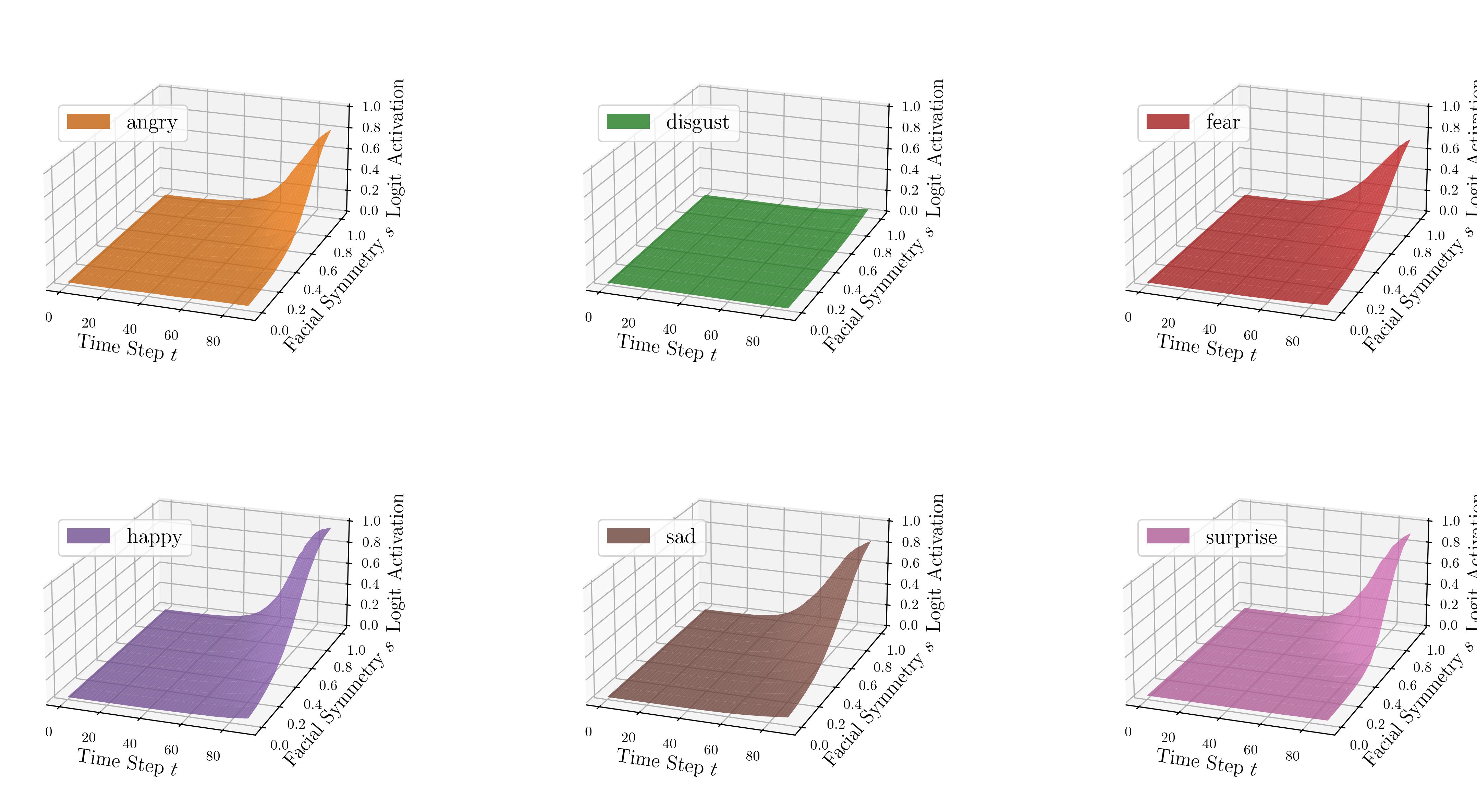}
        \caption{SegmentationVGG19~\cite{vigneshNovelFacialEmotion2023}}
        \label{fig:sup_segvgg_surprise}
    \end{subfigure}
    \\
    \begin{subfigure}[b]{0.4\textwidth}
        \centering
        \includegraphics[width=\textwidth]{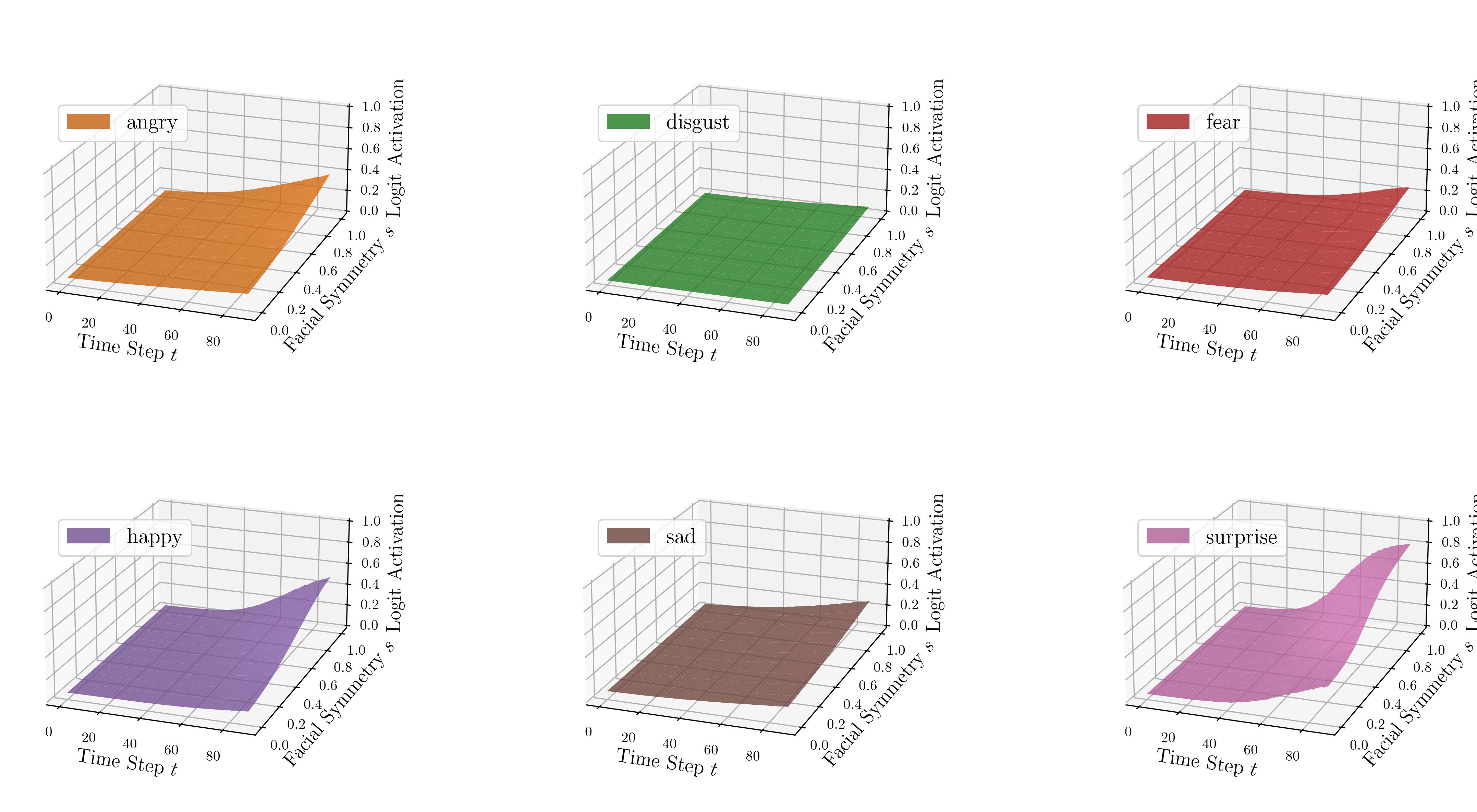}
        \caption{EmoNeXt-Tiny~\cite{boundori2023EmoNext}}
        \label{fig:sup_emonxtT_surprise}
    \end{subfigure}
    \hfill
    \begin{subfigure}[b]{0.4\textwidth}
        \centering
        \includegraphics[width=\textwidth]{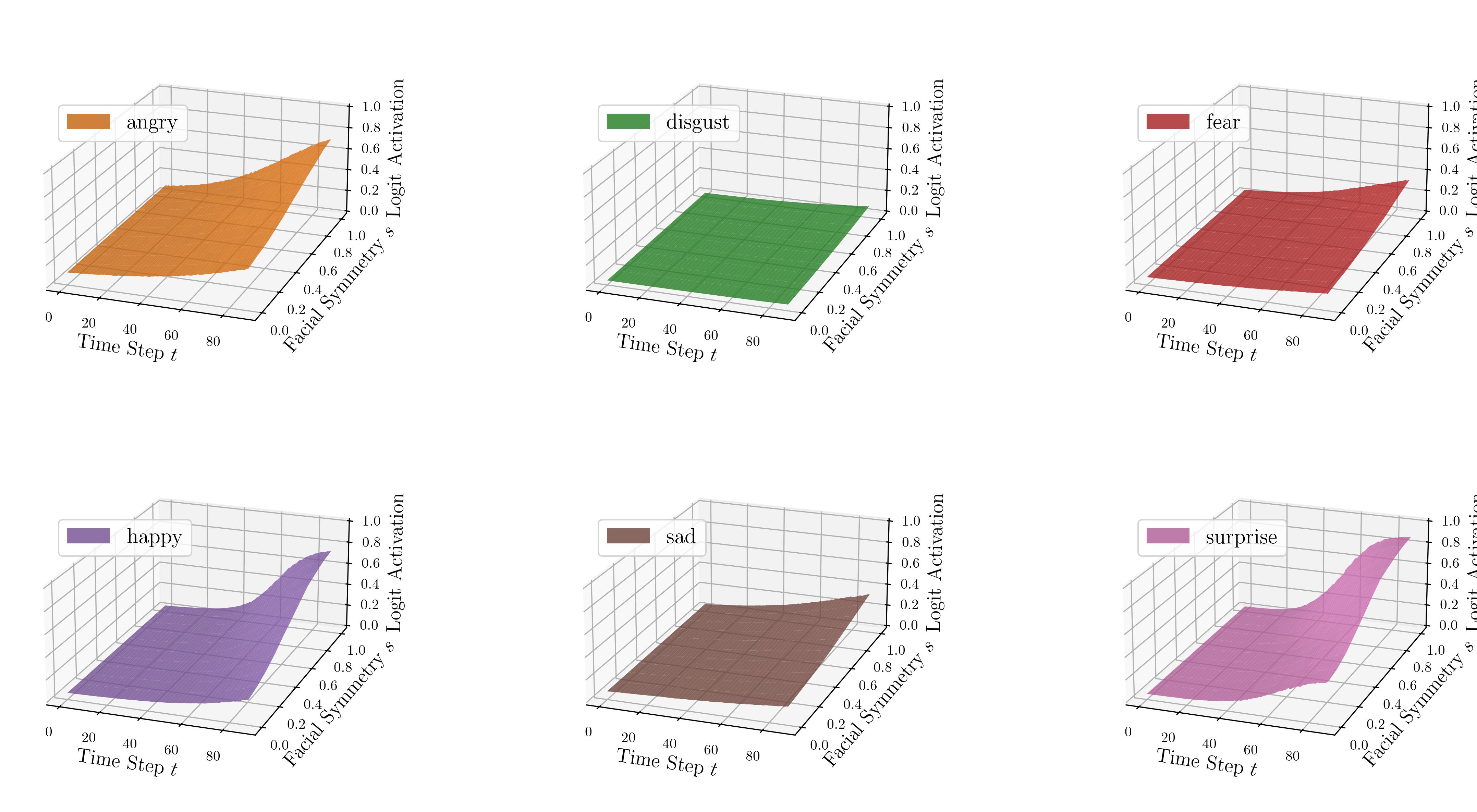}
        \caption{EmoNeXt-Small~\cite{boundori2023EmoNext}}
        \label{fig:sup_emonxtS_surprise}
    \end{subfigure}
    \\
    \begin{subfigure}[b]{0.4\textwidth}
        \centering
        \includegraphics[width=\textwidth]{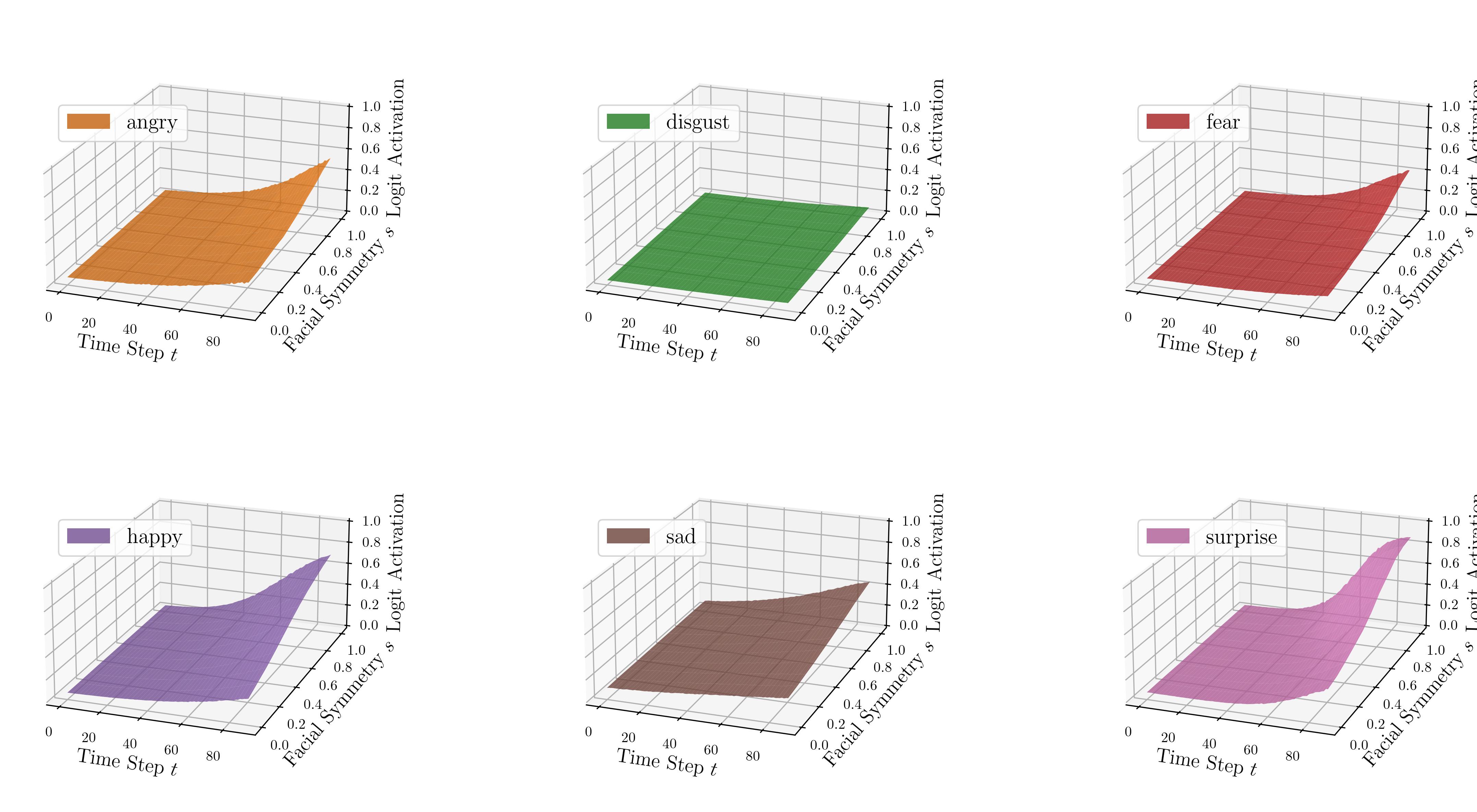}
        \caption{EmoNeXt-Base~\cite{boundori2023EmoNext}}
        \label{fig:sup_emonxtB_surprise}
    \end{subfigure}
    \hfill
    \begin{subfigure}[b]{0.4\textwidth}
        \centering
        \includegraphics[width=\textwidth]{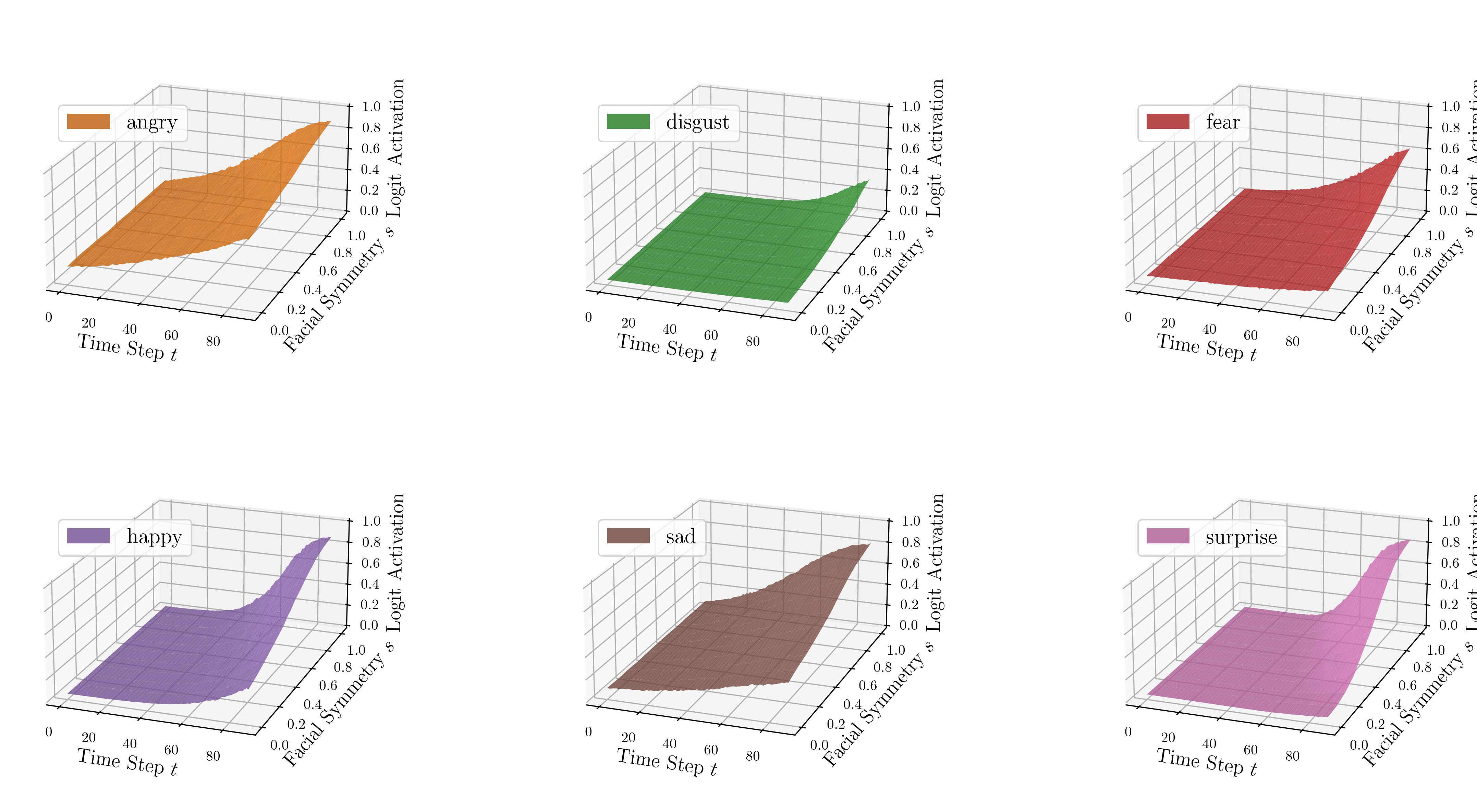}
        \caption{EmoNeXt-Large~\cite{boundori2023EmoNext}}
        \label{fig:sup_emonxtL_surprise}
    \end{subfigure}
    \caption{FER2013~\cite{dumitru2013fer}}
    \label{fig:sup7}
\end{figure}

\begin{figure}[H]
    \centering
    \begin{subfigure}[b]{0.45\textwidth}
        \centering
        \includegraphics[width=\textwidth]{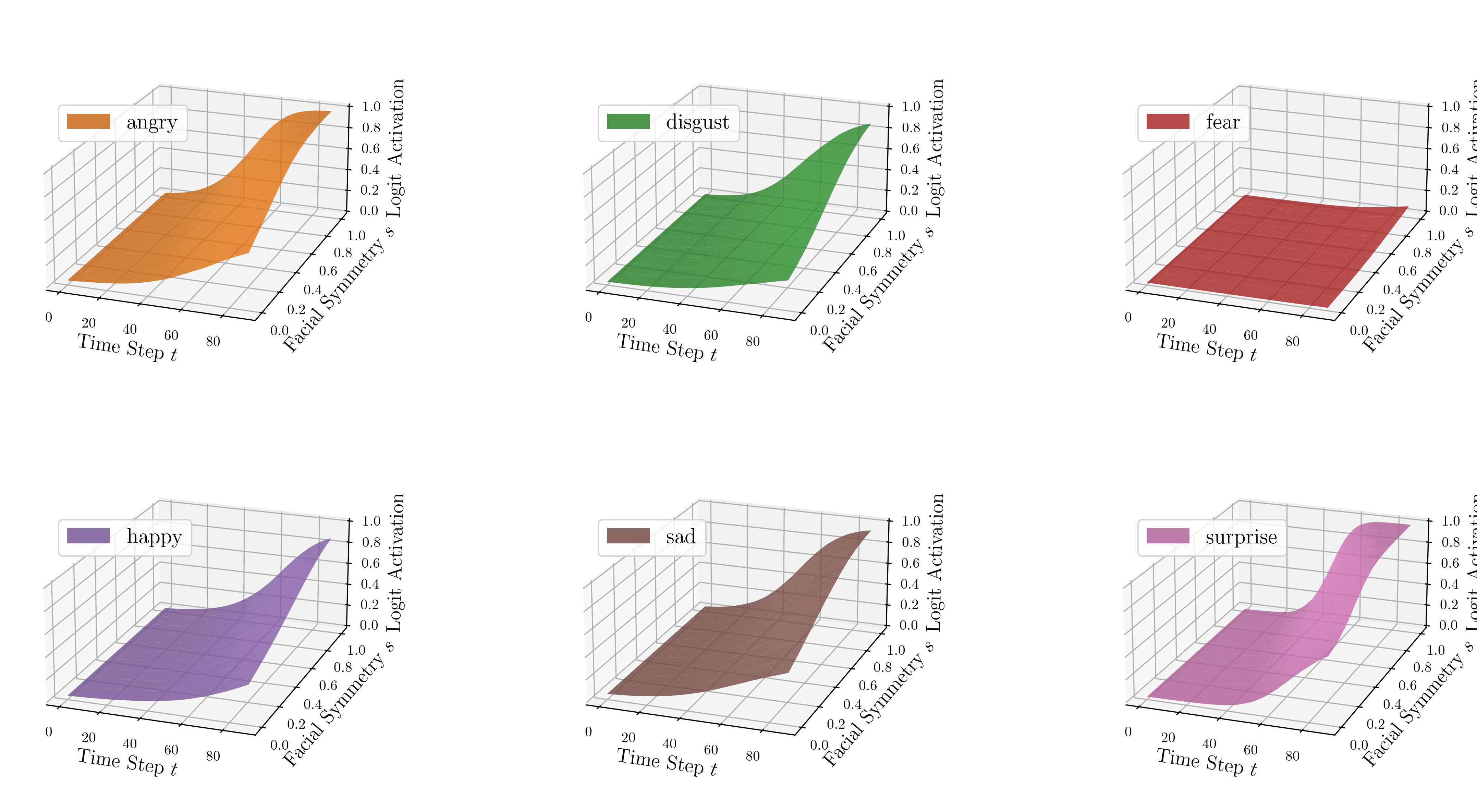}
        \caption{DAN~\cite{wenDistractYourAttention2023}}
        \label{fig:sup_DAN_surprise}
    \end{subfigure}
    \hfill
    \begin{subfigure}[b]{0.45\textwidth}
        \centering
        \includegraphics[width=\textwidth]{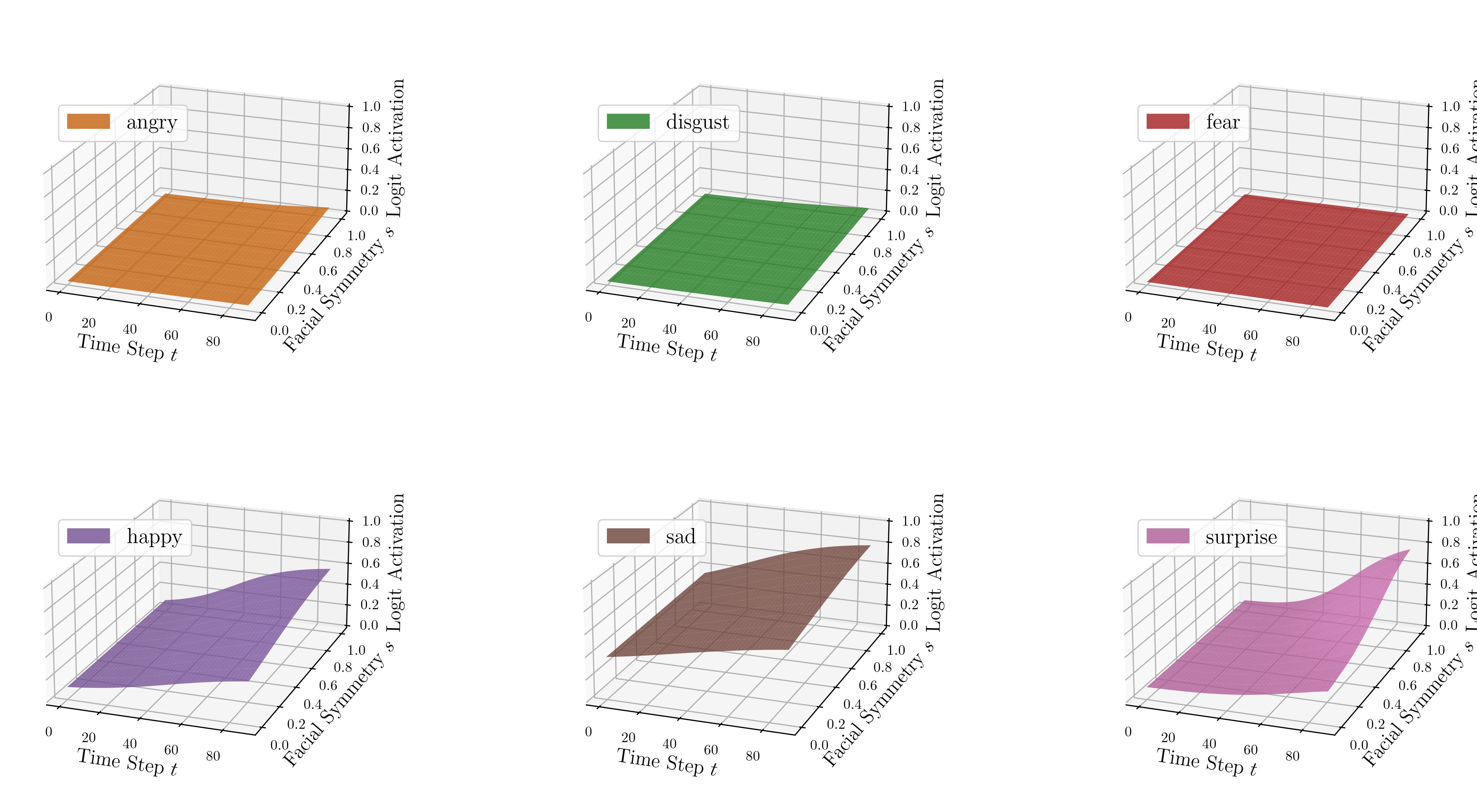}
        \caption{DDAMFN++~\cite{zhangDualDirectionAttentionMixed2023}}
        \label{fig:sup_ddamfn_surprise}
    \end{subfigure}
    \\
    \begin{subfigure}[b]{0.45\textwidth}
        \centering
        \includegraphics[width=\textwidth]{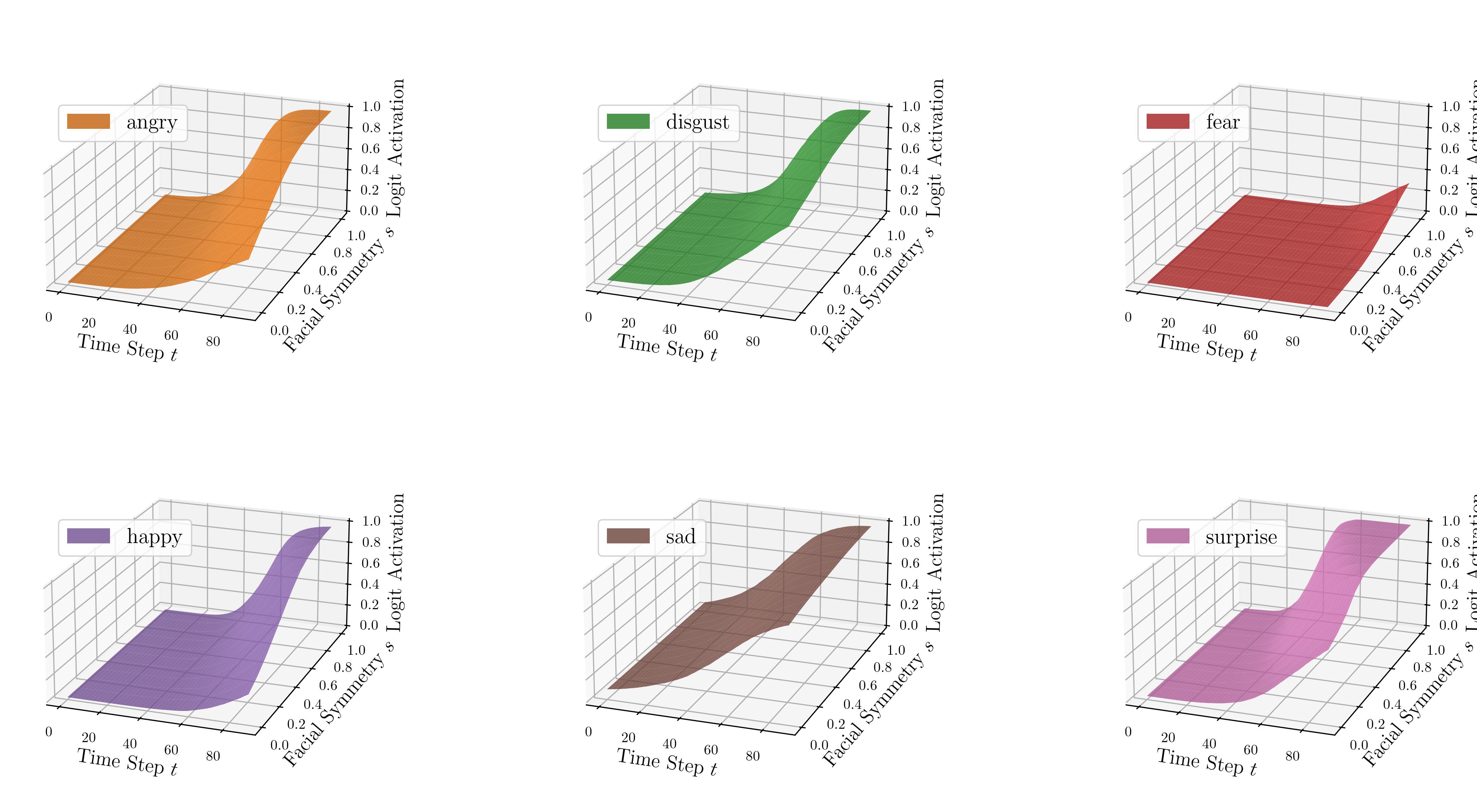}
        \caption{PosterV2~\cite{maoPOSTERSimplerStronger2023}}
        \label{fig:sup_postev2_surprise}
    \end{subfigure}
    \caption{RAFDB~\cite{li2017reliable,li2019reliable}}
    \label{fig:sup2}
\end{figure}

\newpage
\subsection{Model Symmetry Impact}
The main paper shows that we compute the interpretable asymmetry score using a finite grid over $\mathfrak{T}$ and $\mathfrak{S}$.
Given that we only highlighted the final time step $t=1.0$, we show here the full logit activation surfaces used to compute our score.

\begin{figure}
    \centering
    \begin{subfigure}[b]{0.45\textwidth}
        \centering
        \includegraphics[width=\textwidth]{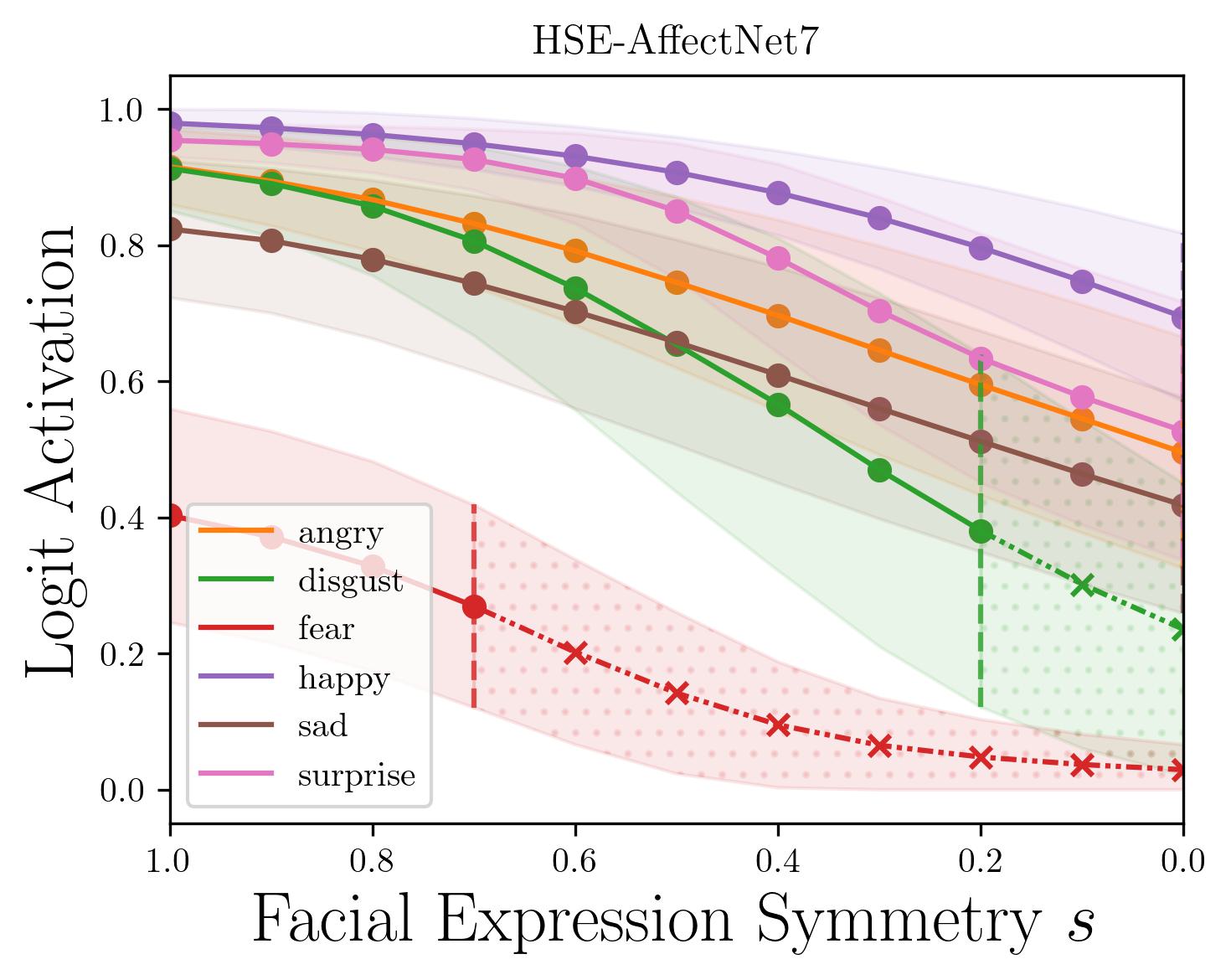}
        \caption{HSEmotion~\cite{savchenko2023facial}}
        \label{fig:sup_HSE7_sym}
    \end{subfigure}
    \hfill
    \begin{subfigure}[b]{0.45\textwidth}
        \centering
        \includegraphics[width=\textwidth]{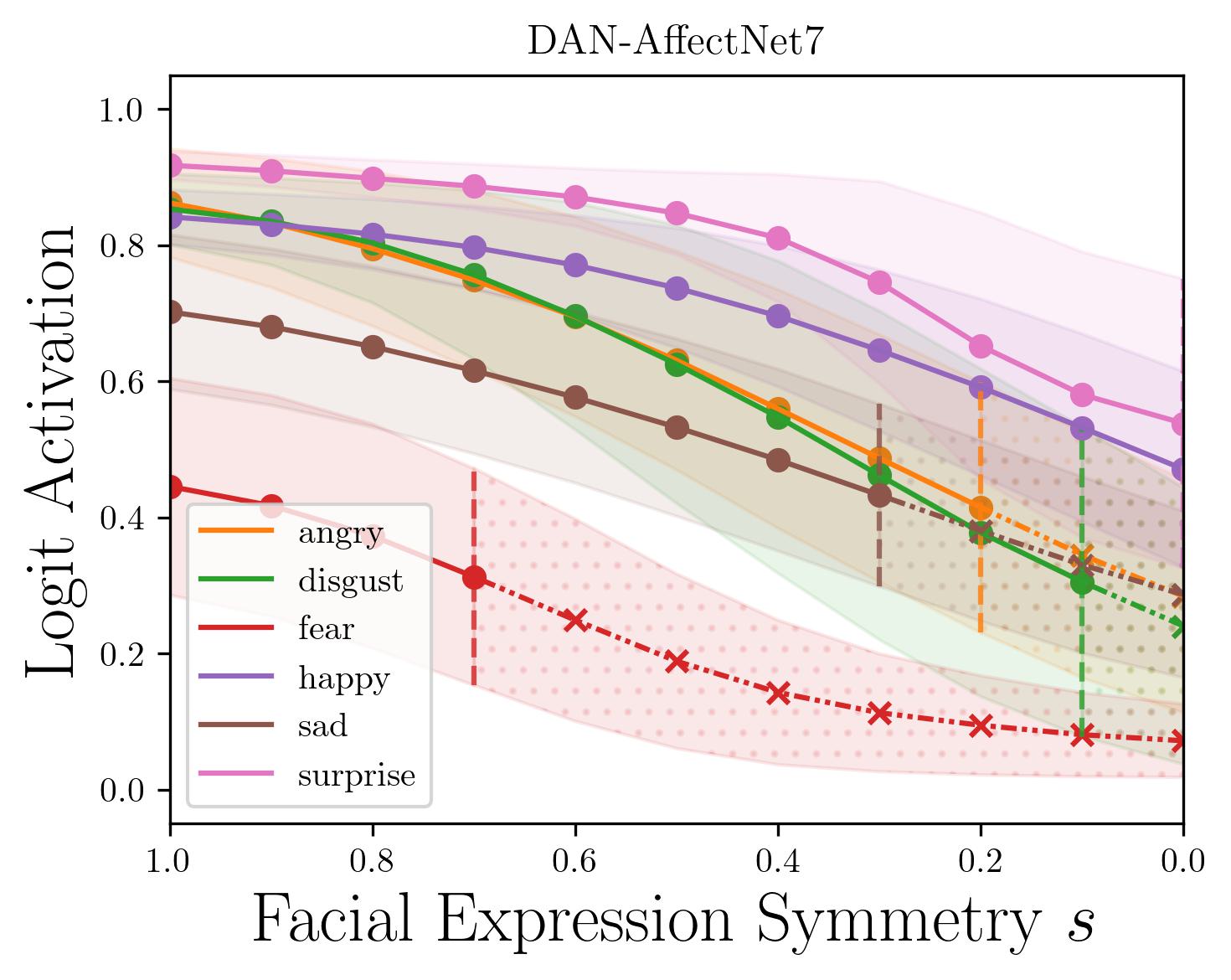}
        \caption{DAN~\cite{wenDistractYourAttention2023}}
        \label{fig:sup_dan-aff7_sym}
    \end{subfigure}
    \\
    \begin{subfigure}[b]{0.45\textwidth}
        \centering
        \includegraphics[width=\textwidth]{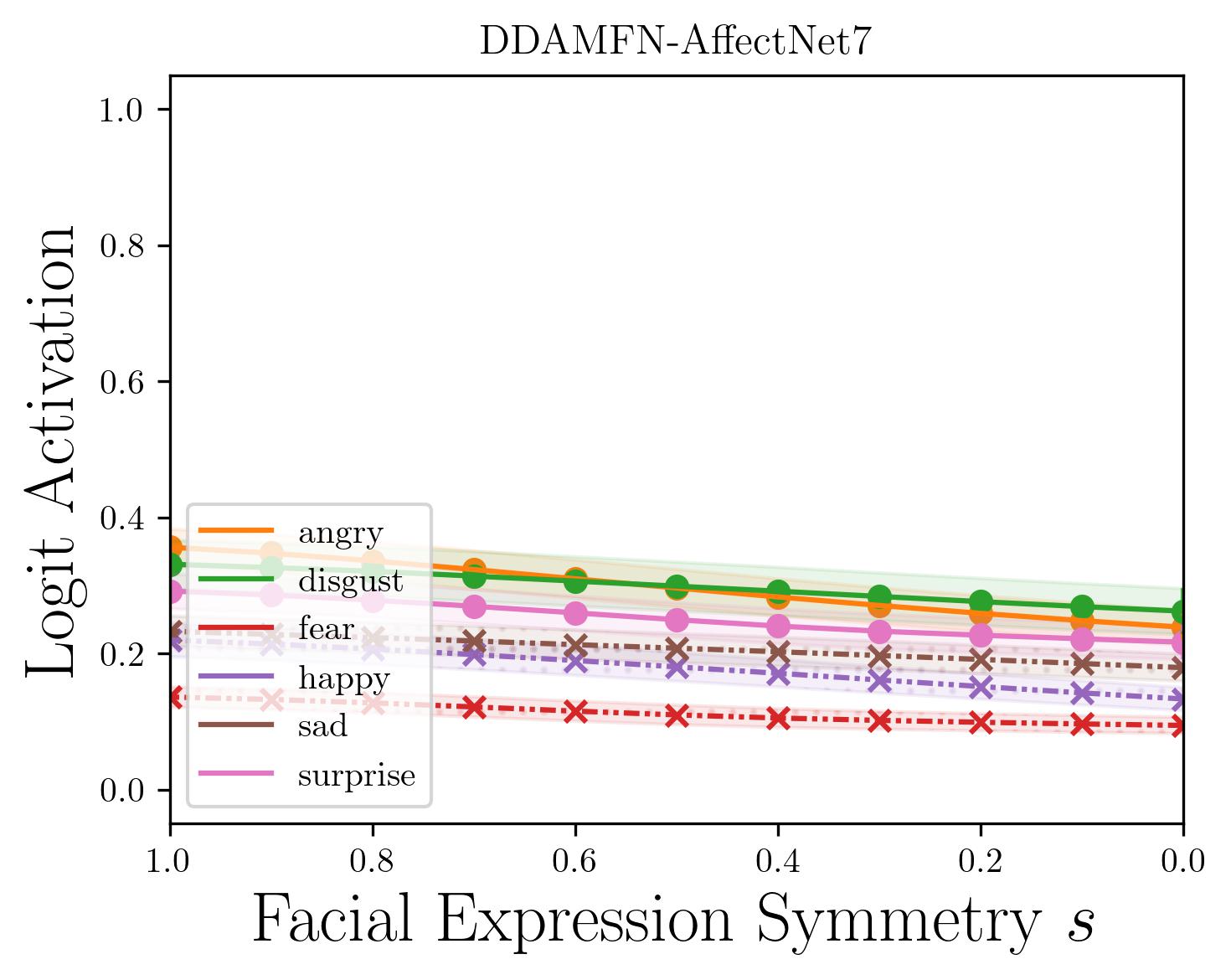}
        \caption{DDAMFN++~\cite{zhangDualDirectionAttentionMixed2023}}
        \label{fig:sup_ddamfn_sym}
    \end{subfigure}
    \hfill
    \begin{subfigure}[b]{0.45\textwidth}
        \centering
        \includegraphics[width=\textwidth]{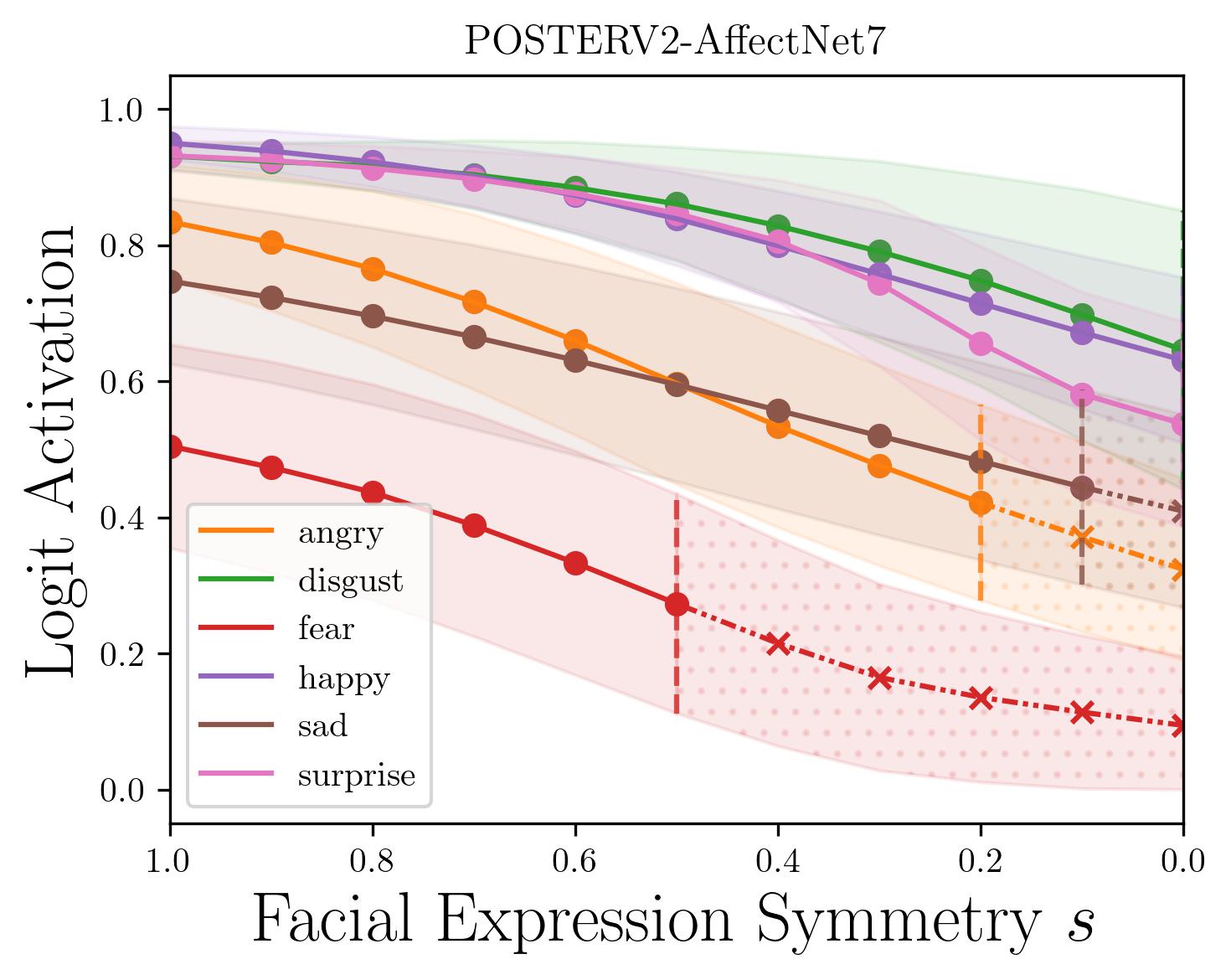}
        \caption{PosterV2~\cite{maoPOSTERSimplerStronger2023}}
        \label{fig:sup_posterv2_sym}
    \end{subfigure}
    \caption{AffectNet7~\cite{Mollahosseini2019affectnet}}
    \label{fig:sup3}
\end{figure}

\begin{figure}
    \centering
    \begin{subfigure}[b]{0.45\textwidth}
        \centering
        \includegraphics[width=\textwidth]{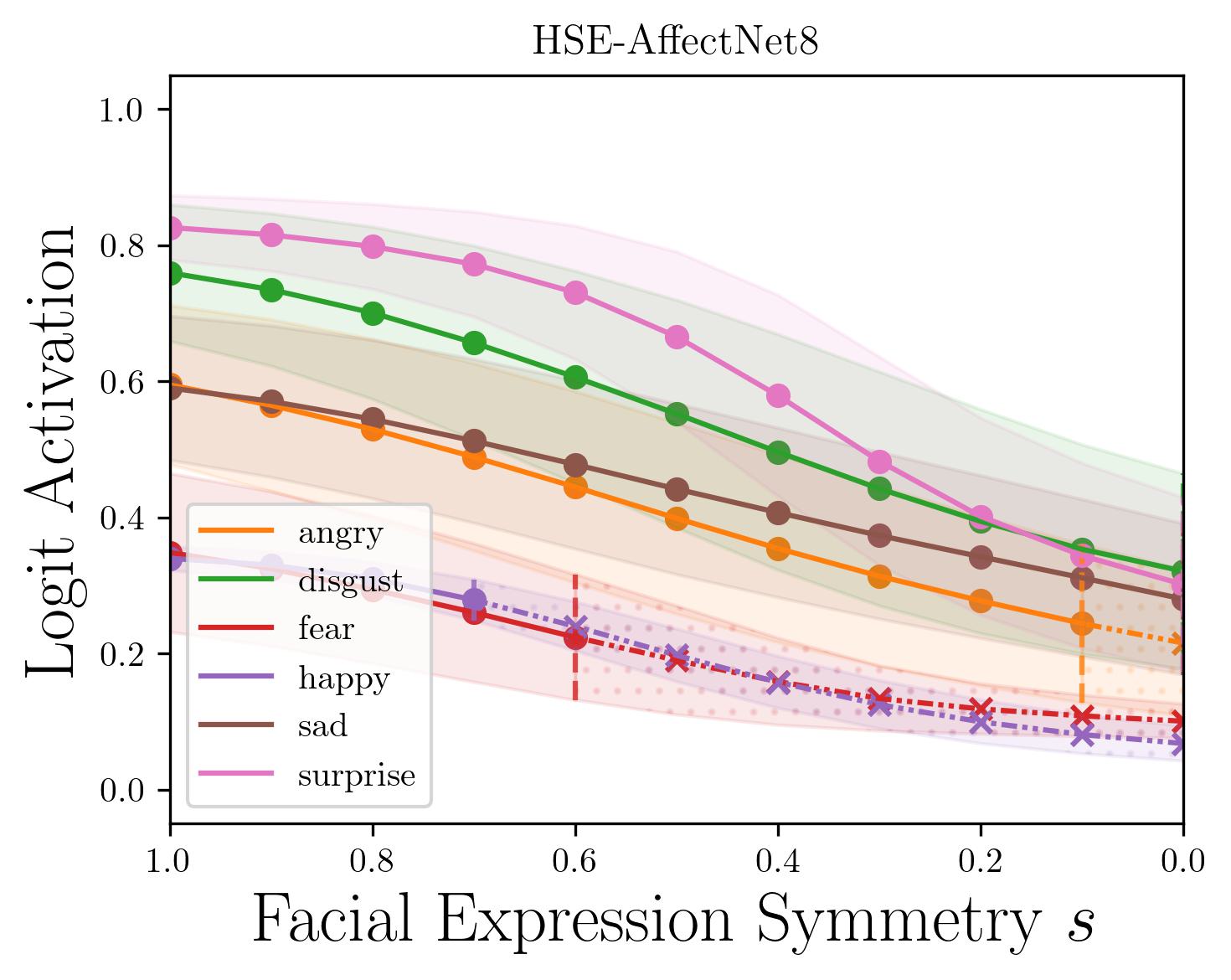}
        \caption{HSEmotion~\cite{savchenko2023facial}}
        \label{fig:sup_HSE_sym}
    \end{subfigure}
    \hfill
    \begin{subfigure}[b]{0.45\textwidth}
        \centering
        \includegraphics[width=\textwidth]{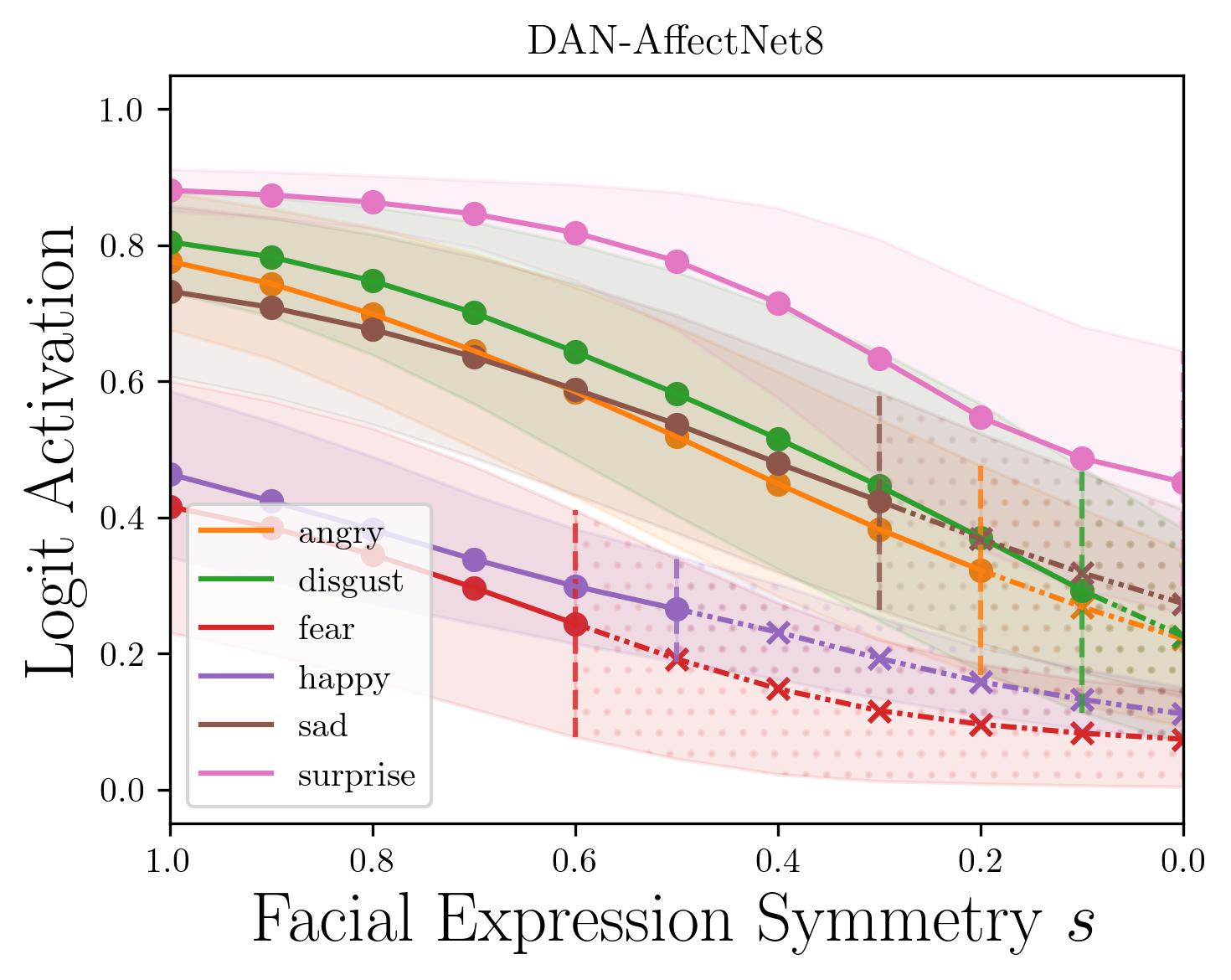}
        \caption{DAN~\cite{wenDistractYourAttention2023}}
        \label{fig:sup_dan-aff8_sym}
    \end{subfigure}
    \\
    \begin{subfigure}[b]{0.45\textwidth}
        \centering
        \includegraphics[width=\textwidth]{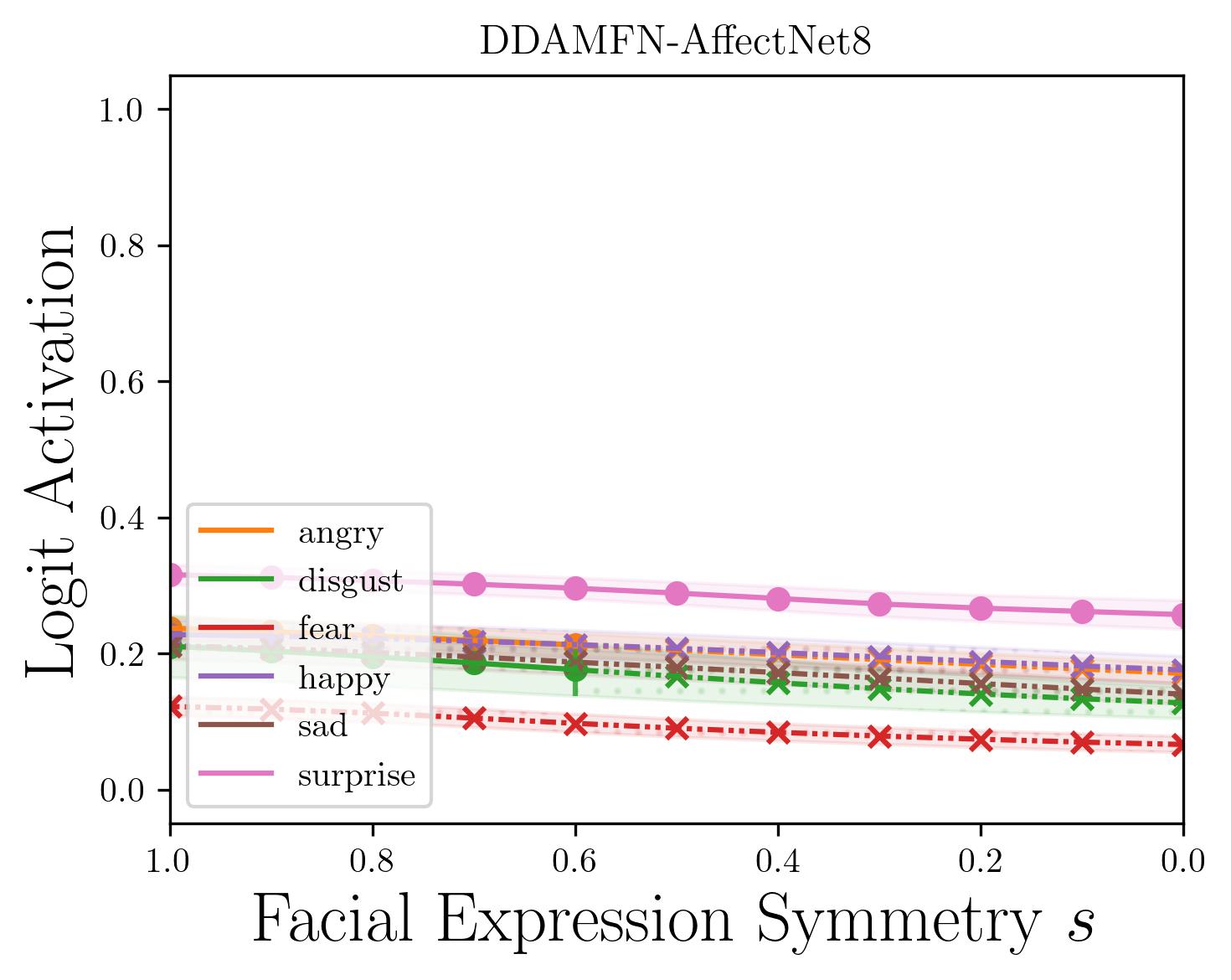}
        \caption{DDAMFN++~\cite{zhangDualDirectionAttentionMixed2023}}
        \label{fig:sup_ddafmn_aff8_sym}
    \end{subfigure}
    \hfill
    \begin{subfigure}[b]{0.45\textwidth}
        \centering
        \includegraphics[width=\textwidth]{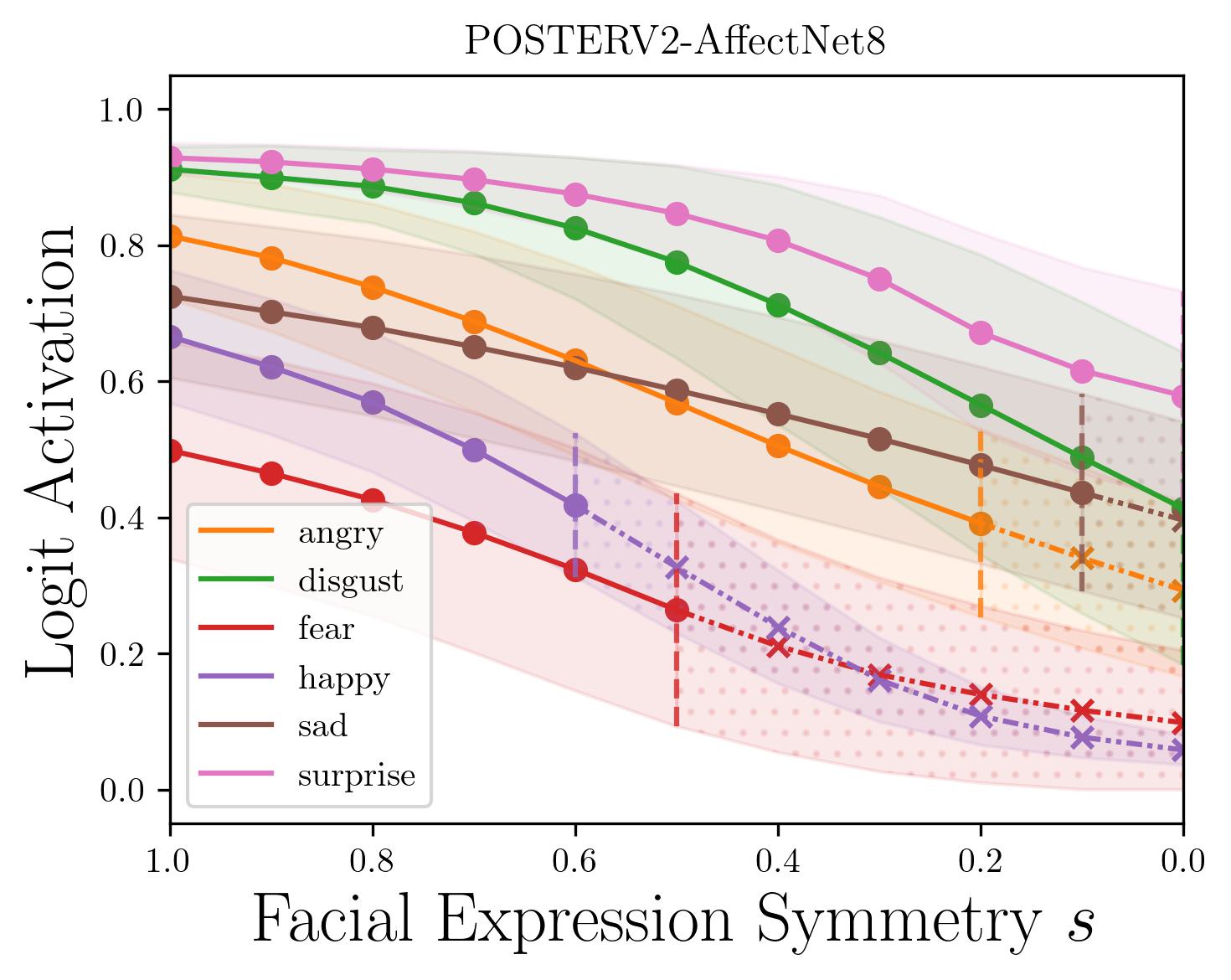}
        \caption{PosterV2~\cite{maoPOSTERSimplerStronger2023}}
        \label{fig:sup_poster_v2-aff8_sym}
    \end{subfigure}
    \caption{AffectNet8~\cite{Mollahosseini2019affectnet}}
    \label{fig:sup4}
\end{figure}

\begin{figure}
    \centering
    \begin{subfigure}[b]{0.45\textwidth}
        \centering
        \includegraphics[width=\textwidth]{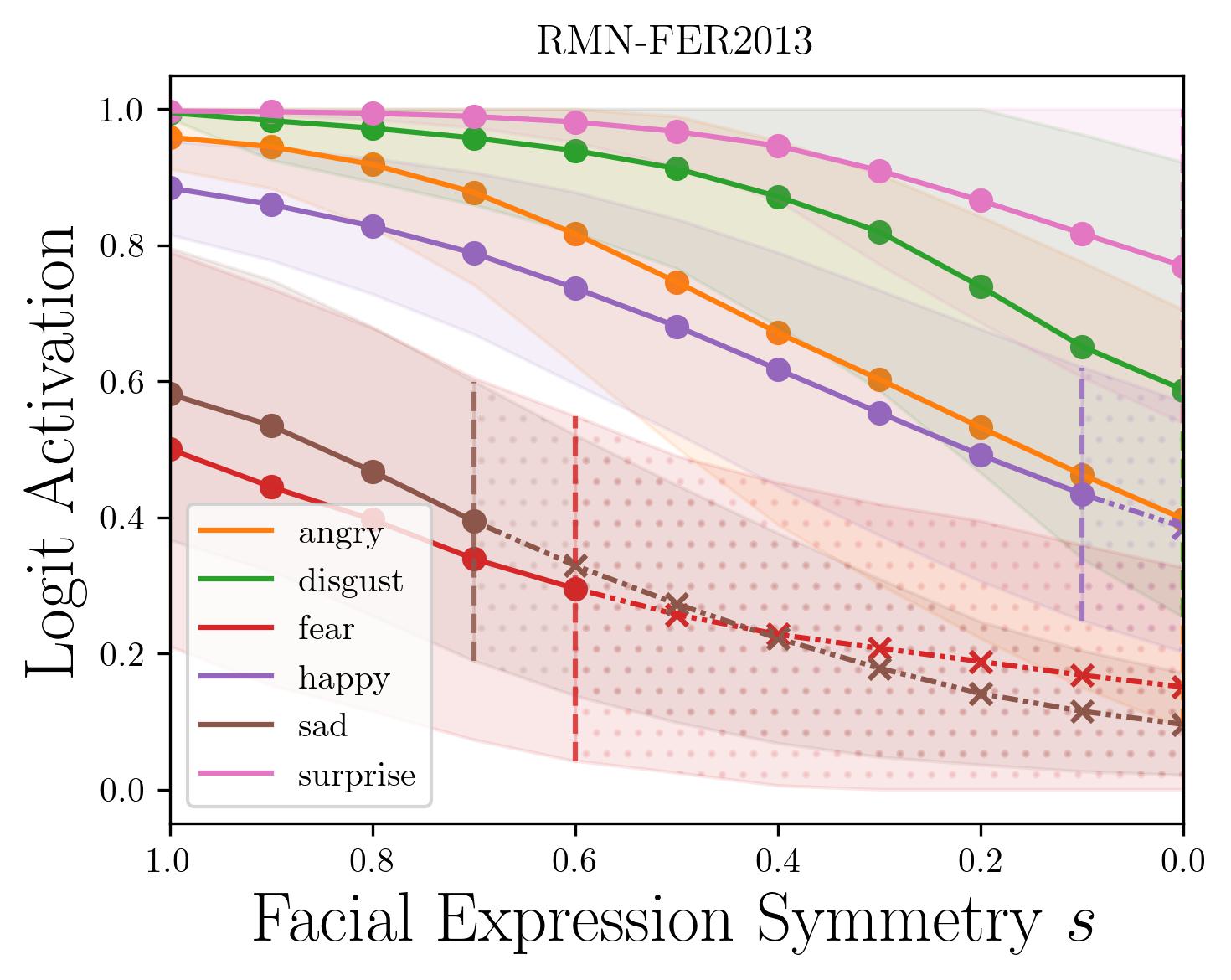}
        \caption{ResidualMaskingNet~\cite{pham2021facial}}
        \label{fig:sup_rmn_sym}
    \end{subfigure}
    \hfill
    \begin{subfigure}[b]{0.45\textwidth}
        \centering
        \includegraphics[width=\textwidth]{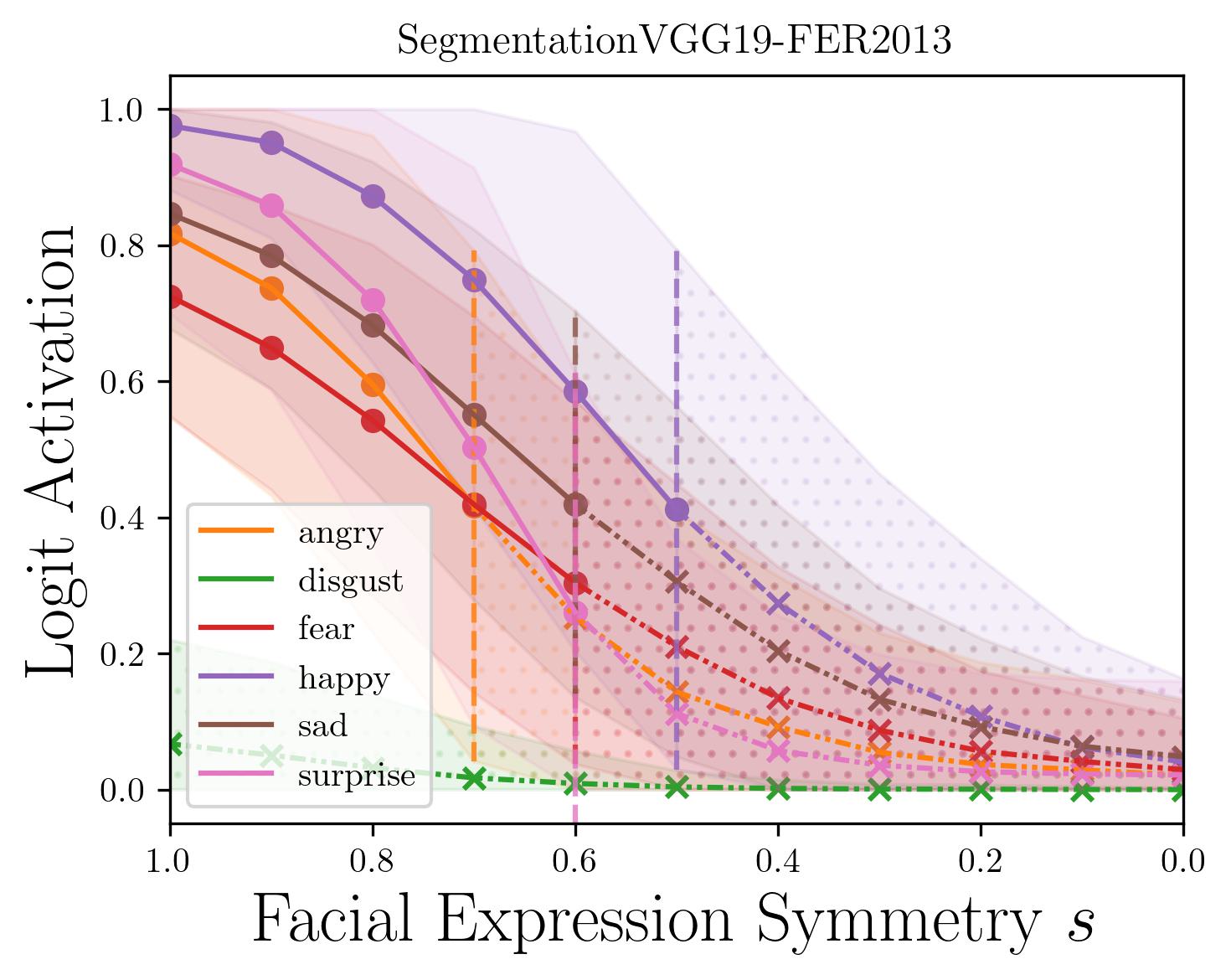}
        \caption{SegmentationVGG19~\cite{vigneshNovelFacialEmotion2023}}
        \label{fig:sup_segvgg_sym}
    \end{subfigure}
    \\
    \begin{subfigure}[b]{0.45\textwidth}
        \centering
        \includegraphics[width=\textwidth]{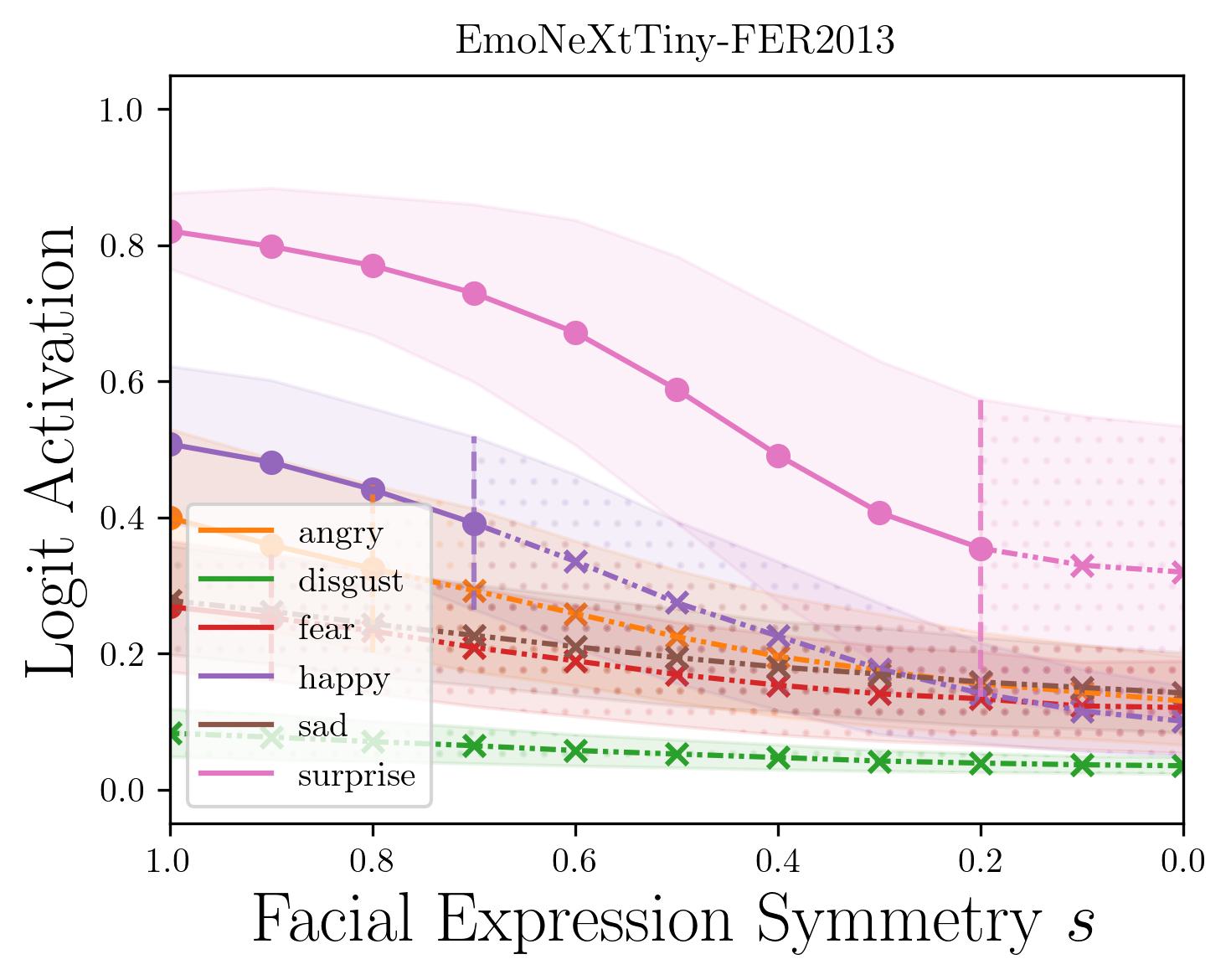}
        \caption{EmoNeXt-Tiny~\cite{boundori2023EmoNext}}
        \label{fig:sup_emonxtT_sym}
    \end{subfigure}
    \hfill
    \begin{subfigure}[b]{0.45\textwidth}
        \centering
        \includegraphics[width=\textwidth]{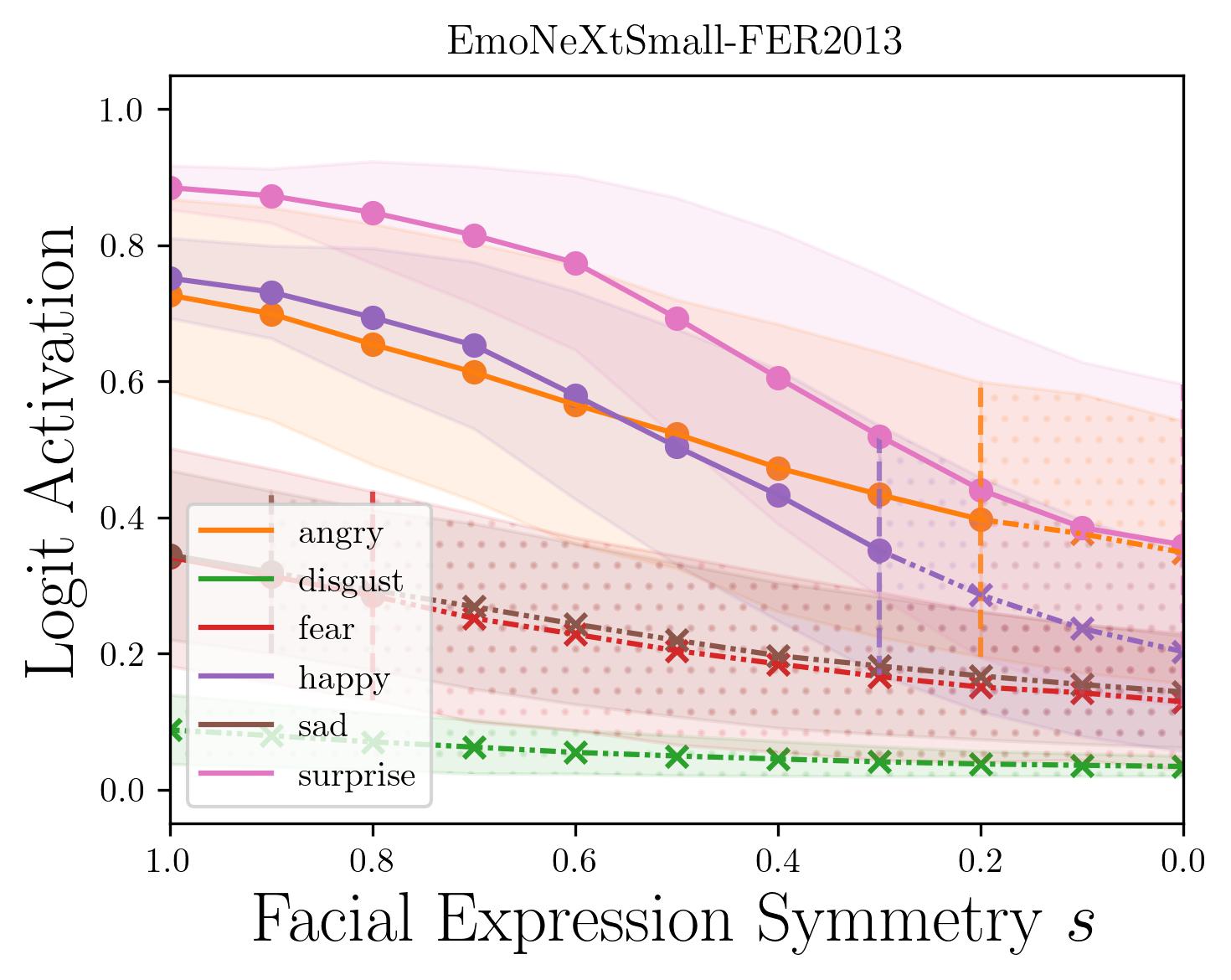}
        \caption{EmoNeXt-Small~\cite{boundori2023EmoNext}}
        \label{fig:sup_emonxtS_sym}
    \end{subfigure}
    \\
    \begin{subfigure}[b]{0.45\textwidth}
        \centering
        \includegraphics[width=\textwidth]{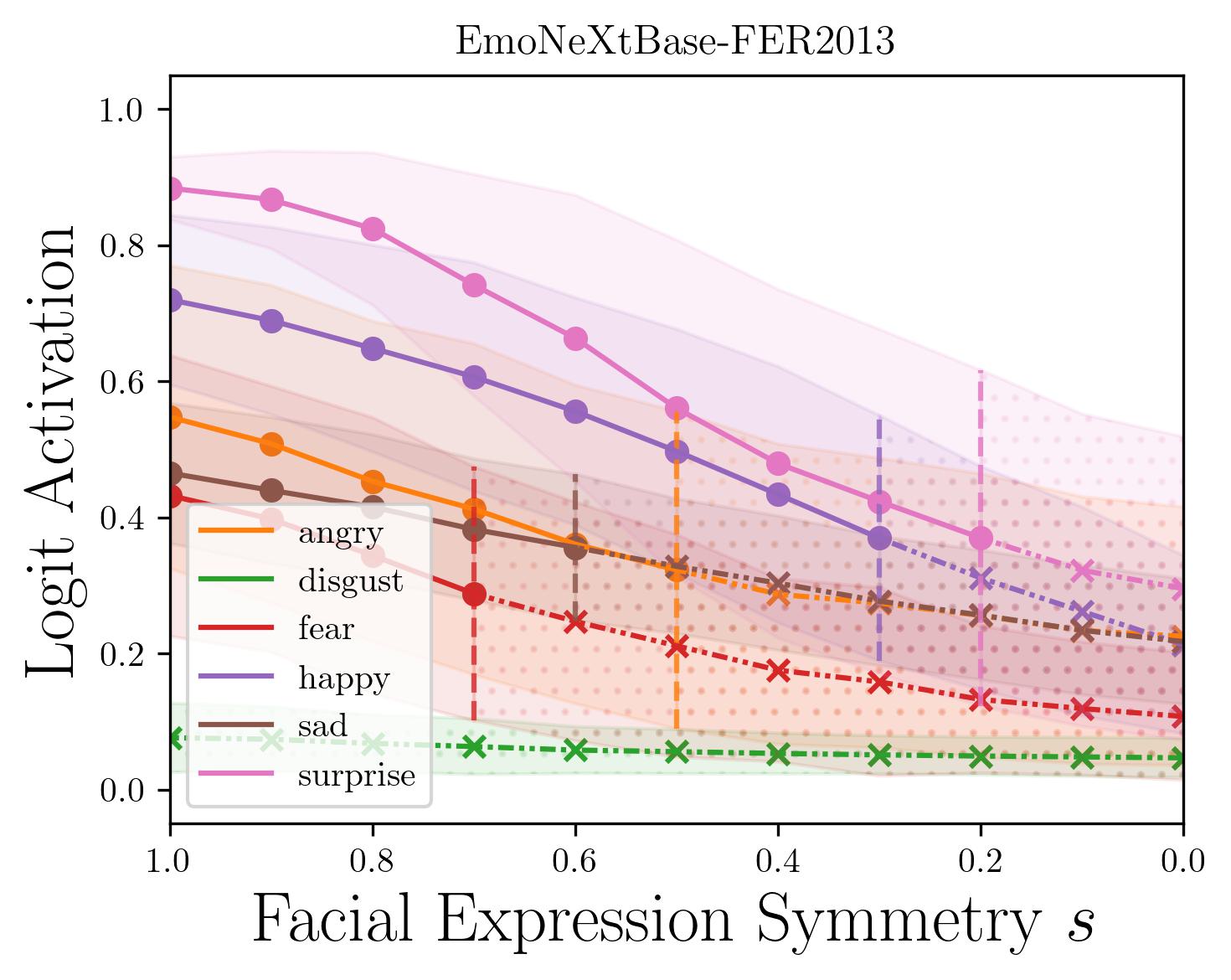}
        \caption{EmoNeXt-Base~\cite{boundori2023EmoNext}}
        \label{fig:sup_emonxtB_sym}
    \end{subfigure}
    \hfill
    \begin{subfigure}[b]{0.45\textwidth}
        \centering
        \includegraphics[width=\textwidth]{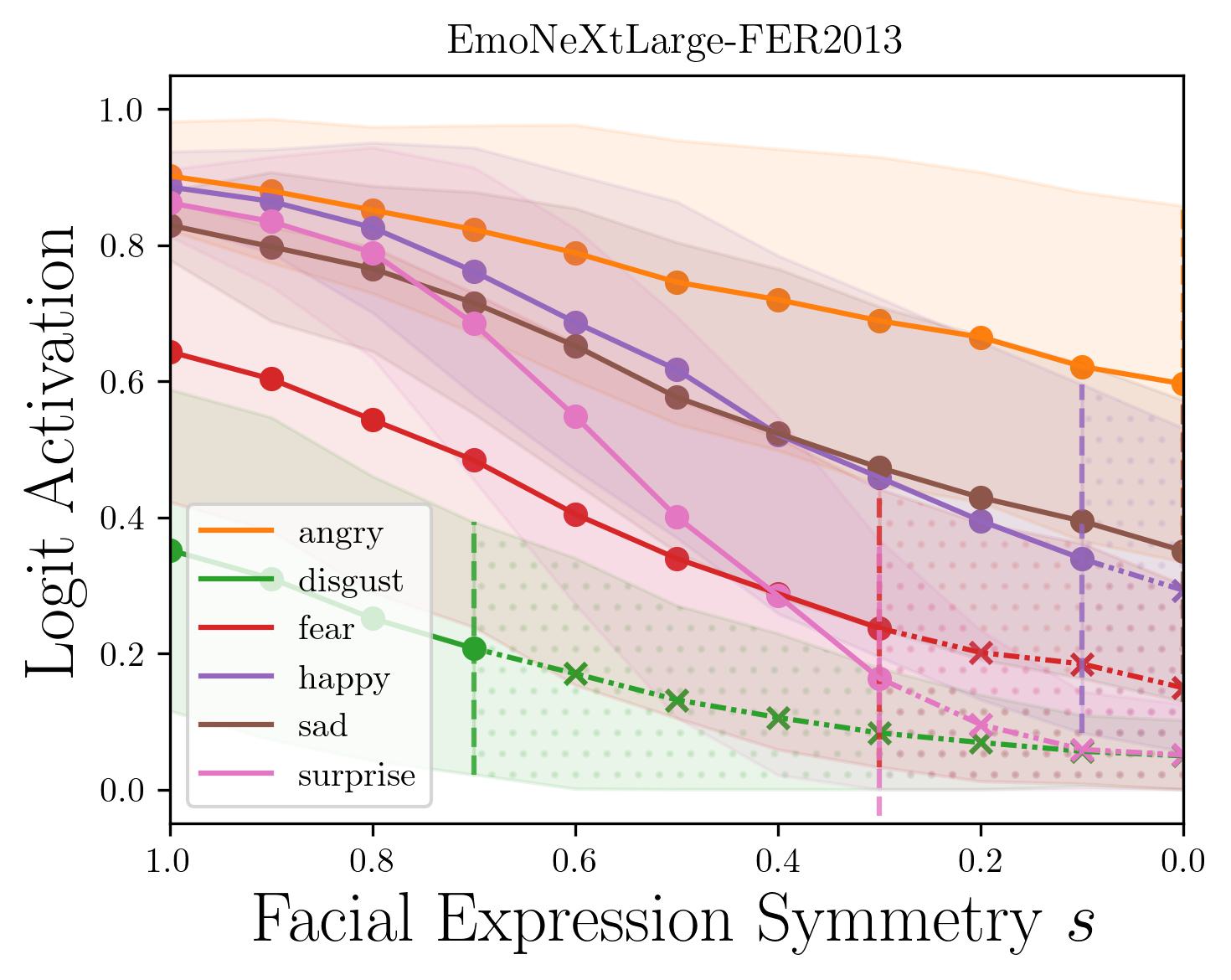}
        \caption{EmoNeXt-Large~\cite{boundori2023EmoNext}}
        \label{fig:sup_emonxtL_sym}
    \end{subfigure}
    \caption{FER2013~\cite{dumitru2013fer}}
    \label{fig:sup5}
\end{figure}

\begin{figure}
    \centering
    \begin{subfigure}[b]{0.45\textwidth}
        \centering
        \includegraphics[width=\textwidth]{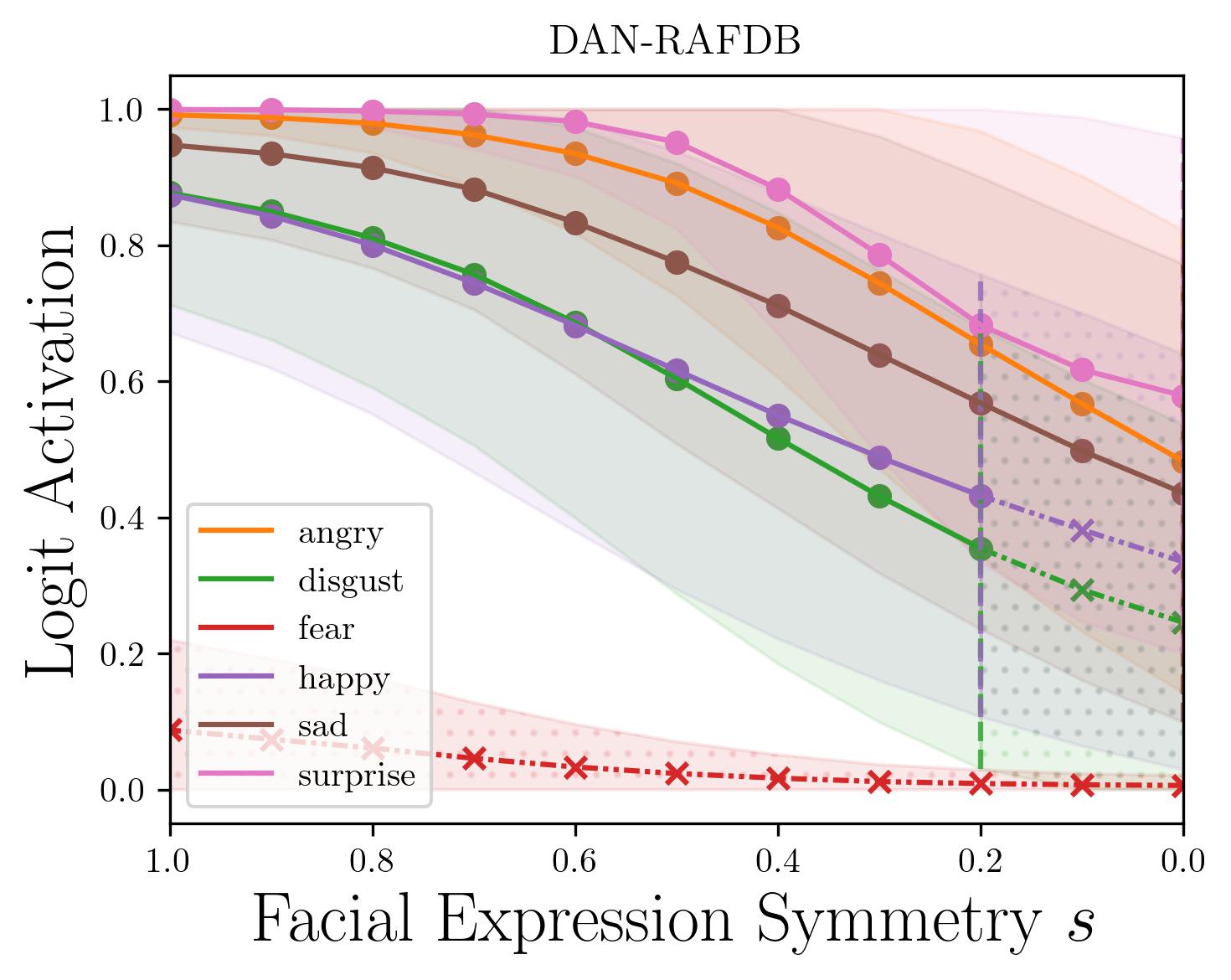}
        \caption{DAN~\cite{wenDistractYourAttention2023}}
        \label{fig:sup_DAN_sym}
    \end{subfigure}
    \hfill
    \begin{subfigure}[b]{0.45\textwidth}
        \centering
        \includegraphics[width=\textwidth]{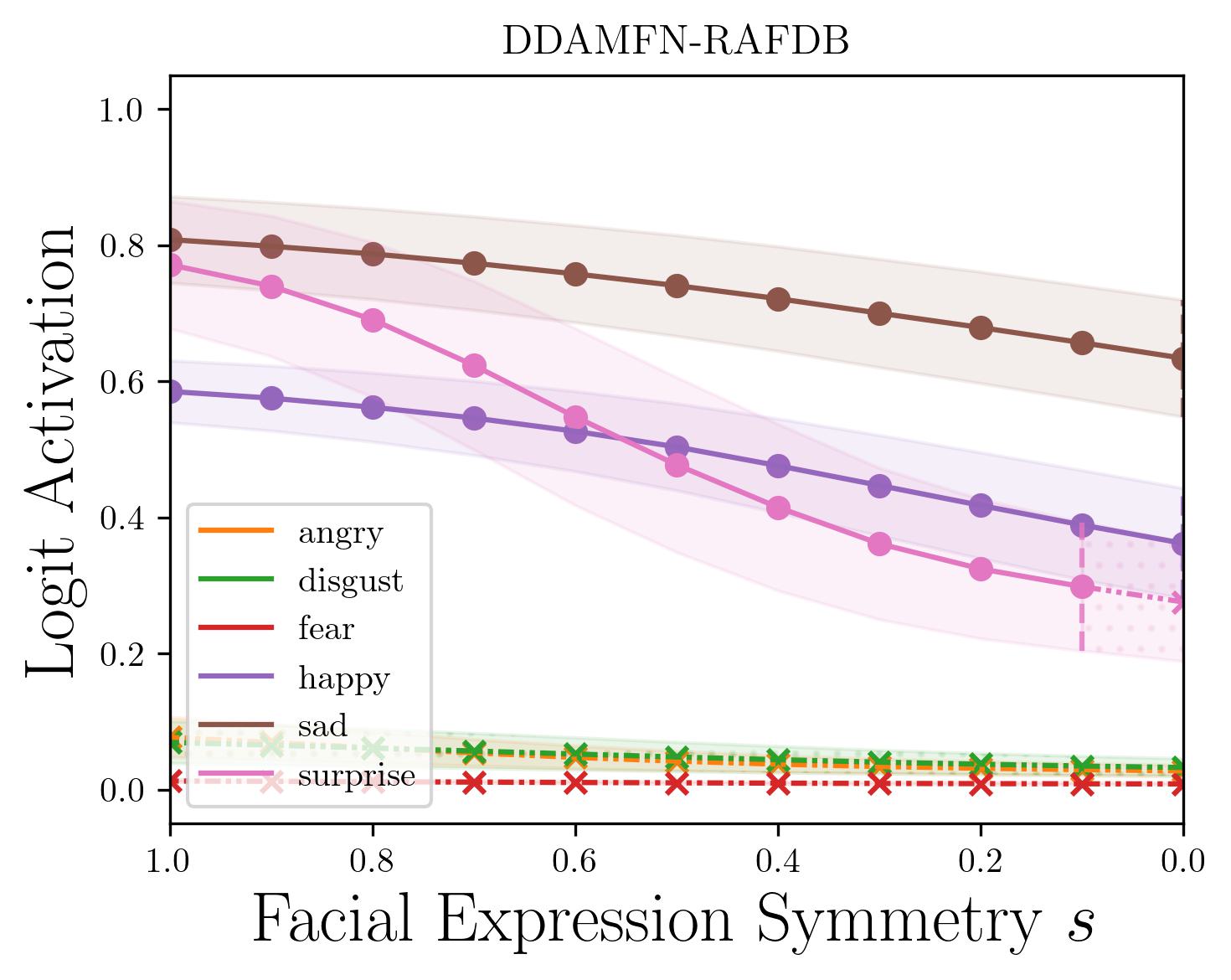}
        \caption{DDAMFN++~\cite{zhangDualDirectionAttentionMixed2023}}
        \label{fig:sup_ddamfn-rafdb_sym}
    \end{subfigure}
    \\
    \begin{subfigure}[b]{0.45\textwidth}
        \centering
        \includegraphics[width=\textwidth]{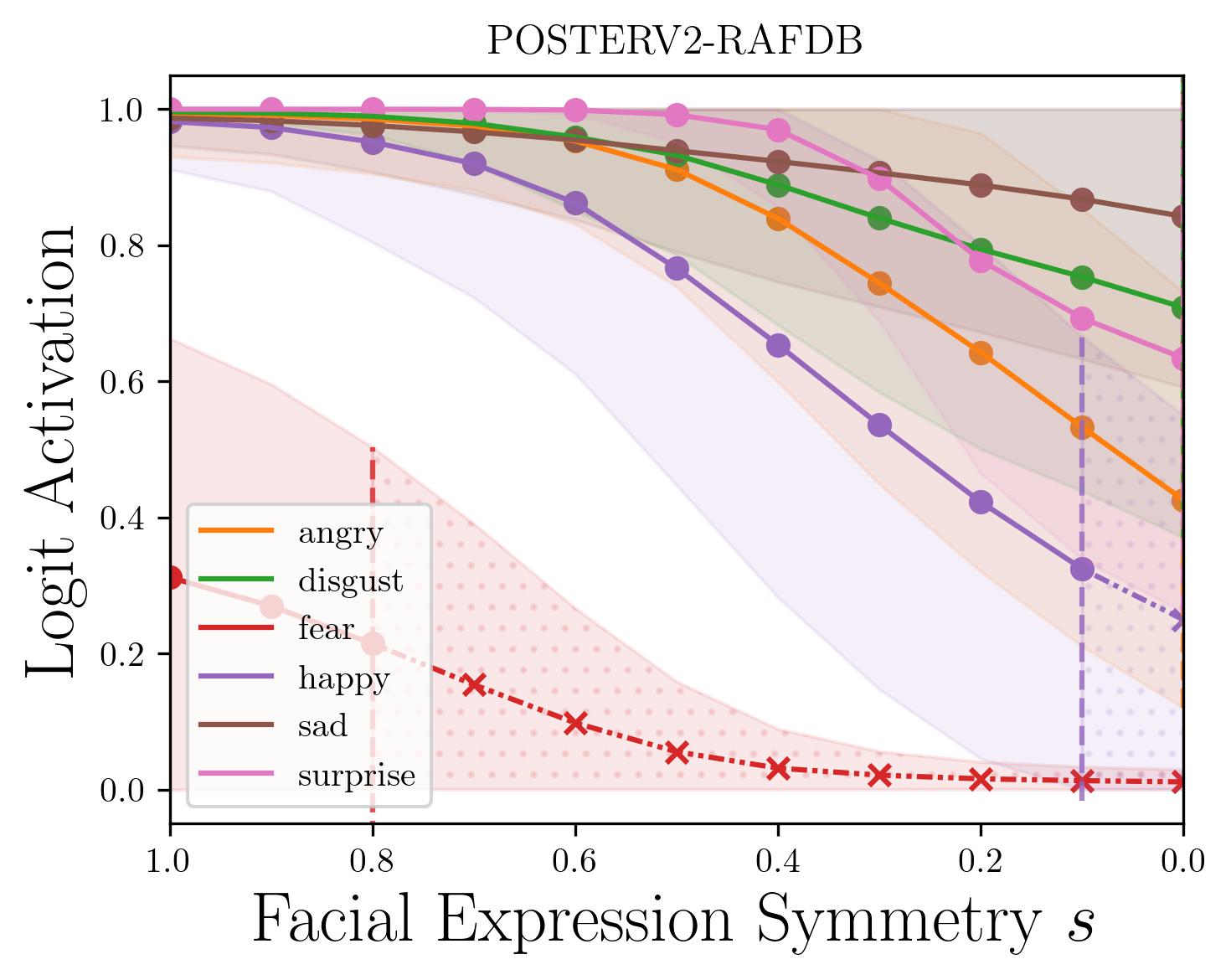}
        \caption{PosterV2~\cite{maoPOSTERSimplerStronger2023}}
        \label{fig:sup_posterv2-rafdb_sym}
    \end{subfigure}
    \caption{RAFDB~\cite{li2017reliable,li2019reliable}}
    \label{fig:sup6}
\end{figure}

\newpage
\subsection{Local Explanations - Saliency Maps}
We aim to understand the impact of facial asymmetry globally; local explanations via saliency maps still offer insights but require human interpretation.
We use an occlusion-based interpretation approach~\cite{zeiler2014visualizing} for the ground truth and predicted label using the average emotion simulated with the default identity.

\subsubsection{Local Explanations - AffectNet7}
The saliency maps indicate that independent of the predicted or the ground truth label, the majority impact is only on one side of the face.
This supports our global observation that facial symmetry has a strong impact on the model behavior.
\begin{figure}[H]
    \centering
    \begin{subfigure}[b]{0.75\textwidth}
        \centering
        \includegraphics[width=\textwidth]{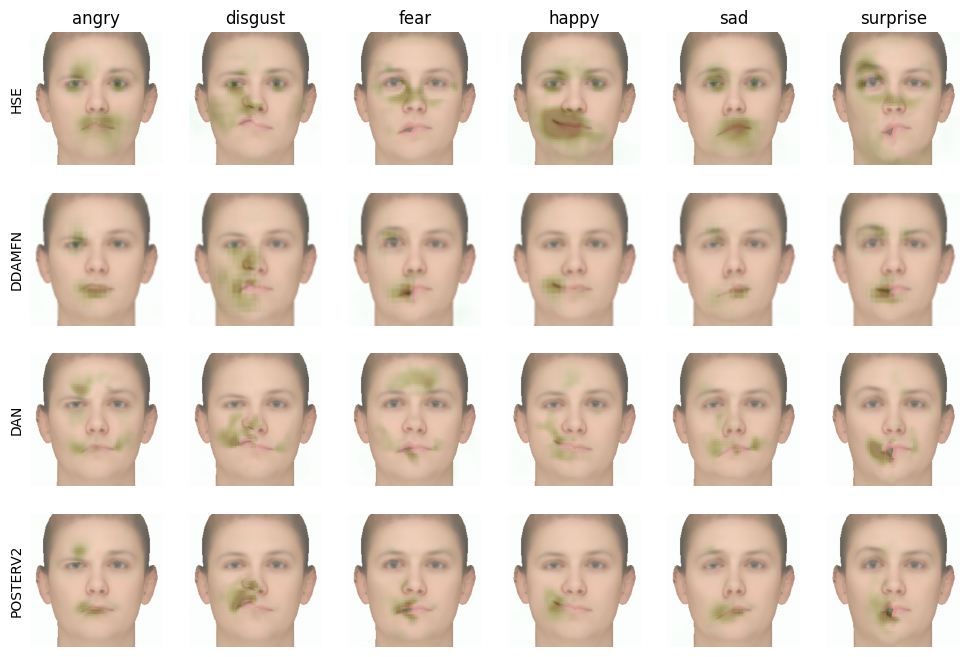}
        \caption{Model focus based on the ground truth label.}
        \label{fig:AffectNet7_occlusion_label}
    \end{subfigure}
    \\
    \begin{subfigure}[b]{0.75\textwidth}
        \centering
        \includegraphics[width=\textwidth]{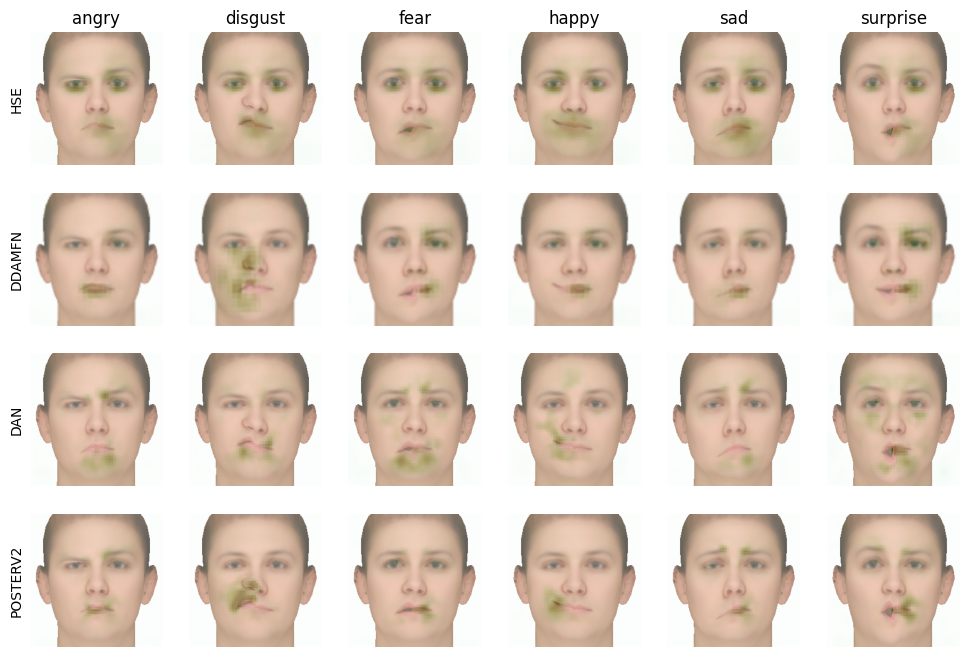}
        \caption{Model focus based on the predicted truth label.}
        \label{fig:AffectNet7_occlusion_predicted}
    \end{subfigure}
    \caption{The occlusion-based saliency maps for models trained on AffectNet7~\cite{Mollahosseini2019affectnet}}
    \label{fig:sal_affectnet7}
\end{figure}

\subsubsection{Local Explanations - AffectNet8}
The saliency maps indicate that independent of the predicted or the ground truth label, the majority impact is only on one side of the face.
This supports our global observation that facial symmetry has a strong impact on the model behavior.
\begin{figure}[H]
    \centering
    \begin{subfigure}[b]{\textwidth}
        \centering
        \includegraphics[width=0.85\textwidth]{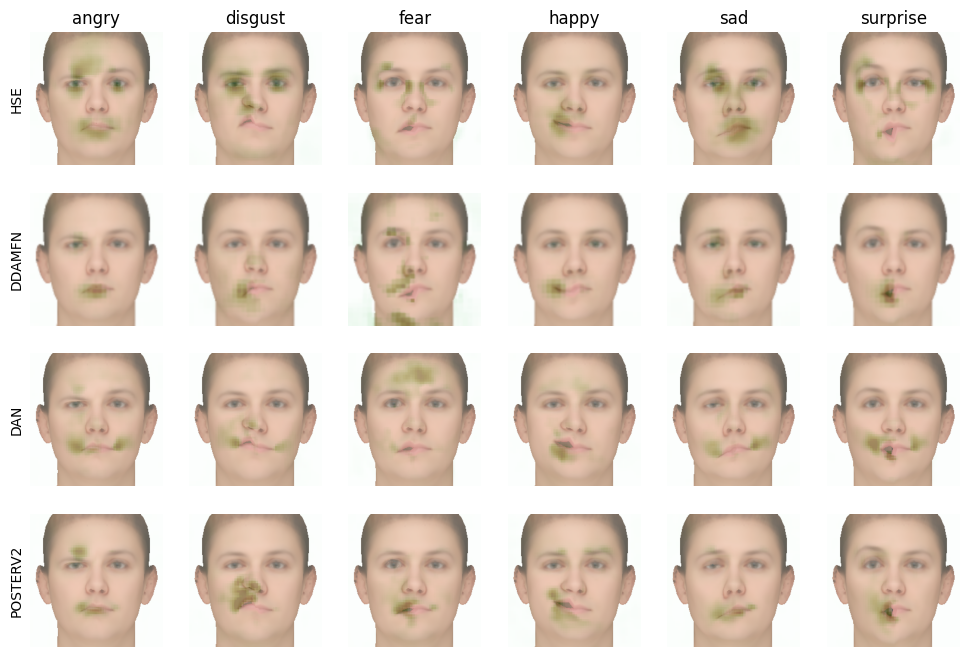}
        \caption{Model focus based on the ground truth label.}
        \label{fig:AffectNet8_occlusion_label}
    \end{subfigure}
    \\
    \begin{subfigure}[b]{\textwidth}
        \centering
        \includegraphics[width=0.85\textwidth]{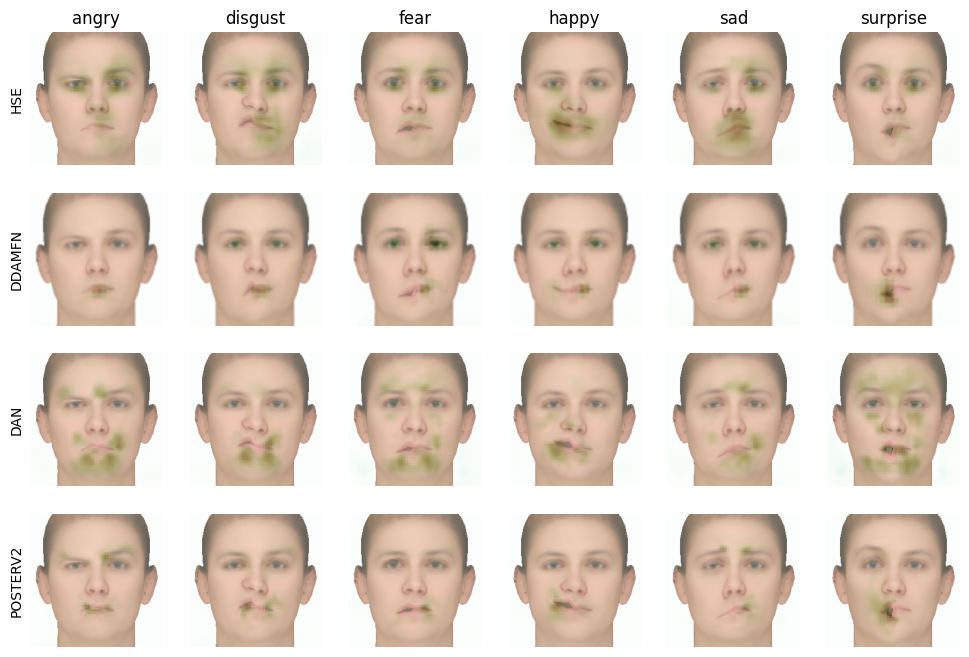}
       \caption{Model focus based on the predicted truth label.}
        \label{fig:AffectNet8_occlusion_predicted}
    \end{subfigure}
    \caption{TThe occlusion-based saliency maps for models trained on AffectNet8~\cite{Mollahosseini2019affectnet}}
    \label{fig:sal_affectnet8}
\end{figure}

\subsubsection{Local Explanations - FER2013}
The saliency maps indicate that independent of the predicted or the ground truth label, the majority impact is only on one side of the face.
This supports our global observation that facial symmetry has a strong impact on the model behavior.
\begin{figure}[H]
    \centering
    \begin{subfigure}[b]{\textwidth}
        \centering
        \includegraphics[width=0.6\textwidth]{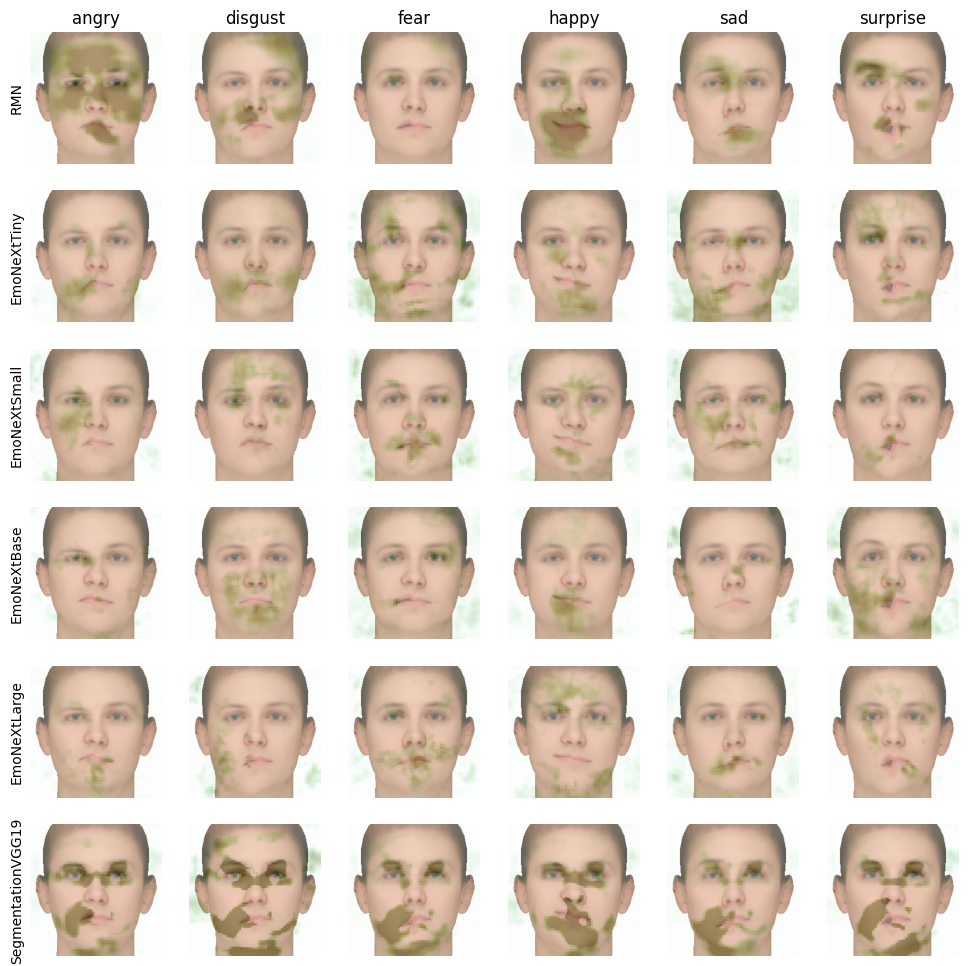}
        \caption{Model focus based on the ground truth label.}
        \label{fig:FER2013_occlusion_label}
    \end{subfigure}
    \\
    \begin{subfigure}[b]{\textwidth}
        \centering
        \includegraphics[width=0.6\textwidth]{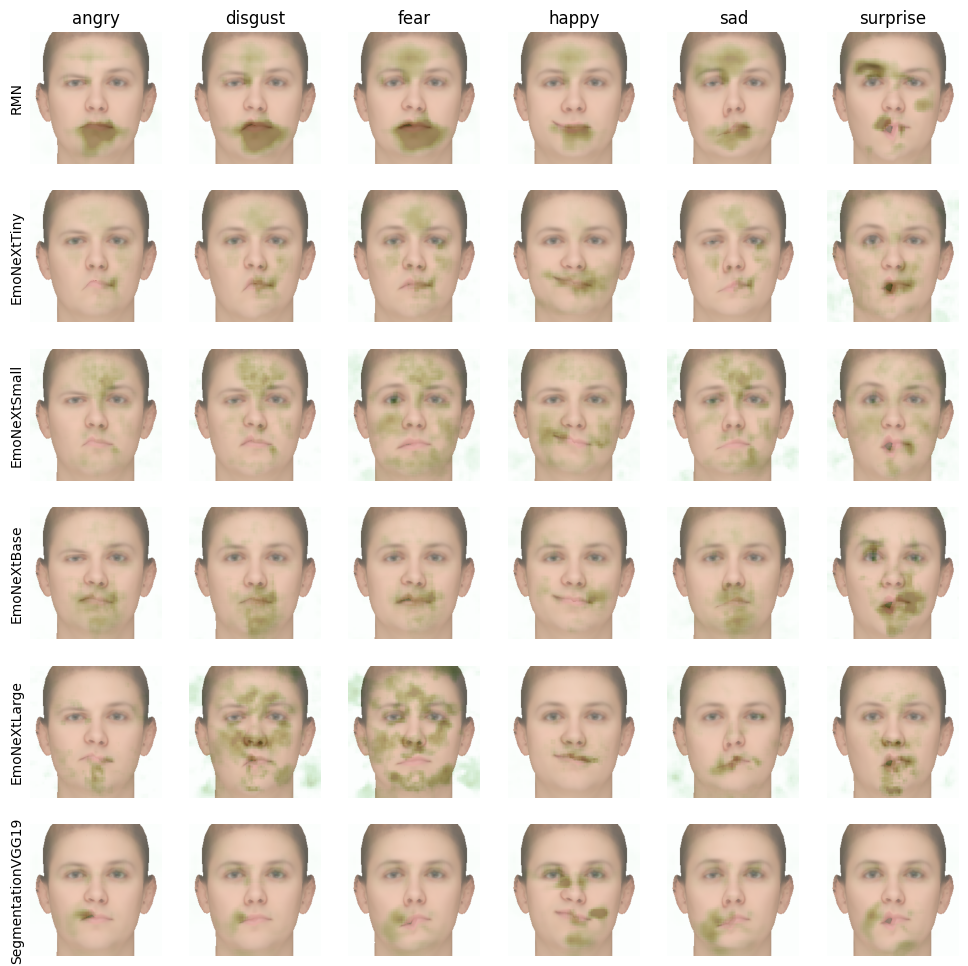}
        \caption{Model focus based on the predicted truth label.}
        \label{fig:FER2013_occlusion_predicted}
    \end{subfigure}
    \caption{The occlusion-based saliency maps for models trained on FER2013~\cite{dumitru2013fer}}
    \label{fig:sal_fer2013}
\end{figure}

\subsubsection{Local Explanations - RAFDB}
The saliency maps indicate that independent of the predicted or the ground truth label, the majority impact is only on one side of the face.
This supports our global observation that facial symmetry has a strong impact on the model behavior.
\begin{figure}[H]
   \centering
    \begin{subfigure}[b]{\textwidth}
        \centering
        \includegraphics[width=0.85\textwidth]{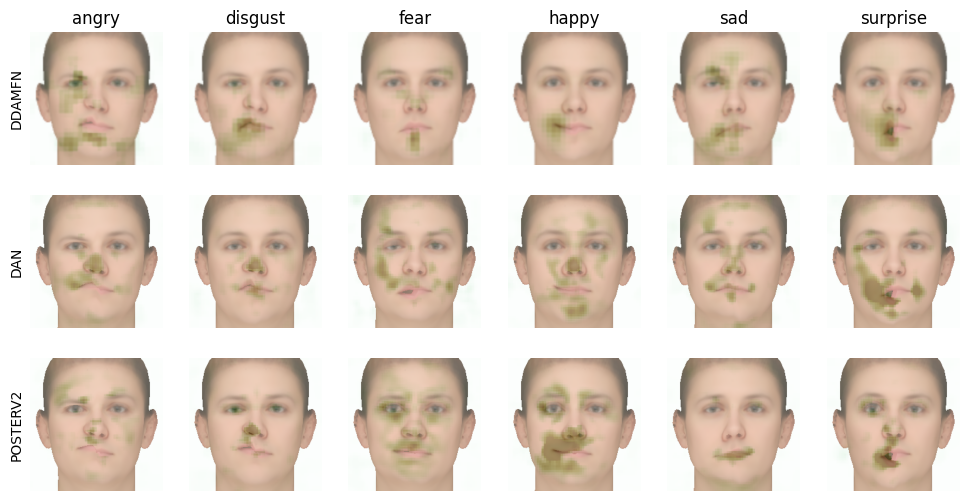}
        \caption{Model focus based on the ground truth label.}
        \label{fig:RAFDB_occlusion_label}
    \end{subfigure}
    \\
    \begin{subfigure}[b]{\textwidth}
        \centering
        \includegraphics[width=0.85\textwidth]{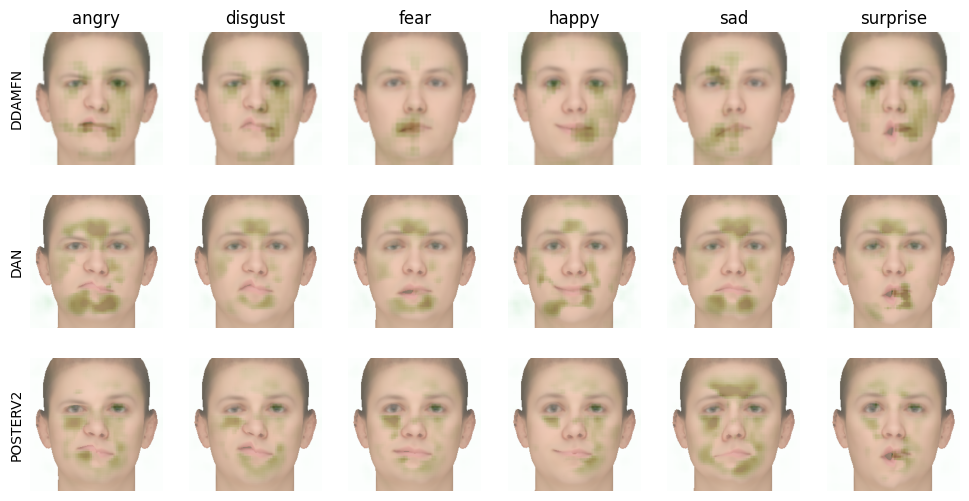}
        \caption{Model focus based on the predicted truth label.}
        \label{fig:RAFDB_occlusion_predicted}
    \end{subfigure}
    \caption{The occlusion-based saliency maps for models trained on RAFDB~\cite{li2017reliable,li2019reliable}}
    \label{fig:sal_rafdb}
\end{figure}
\end{document}